\documentclass{article}
\usepackage{frExamplee}
\usepackage{graphicx}
\usepackage{apalike}
\usepackage{setspace}
\usepackage{amsmath}

\usepackage{xcolor}
\usepackage{float}
\usepackage{subcaption}
\usepackage{multicol}
\usepackage{multirow}
\usepackage{array}
\usepackage{caption}
\usepackage{float}
\usepackage{multicol}
\usepackage{soul}
\usepackage{colortbl}
\usepackage{boldline}
\usepackage{appendix}

\newcolumntype{C}[1]{>{\centering\let\newline\\\arraybackslash\hspace{0pt}}p{#1}}

\definecolor{lightgray}{rgb}{0.83, 0.83, 0.83}
\definecolor{vlgray}{rgb}{0.95, 0.95, 0.95}
\definecolor{tearose}{rgb}{0.96, 0.76, 0.76}
\definecolor{orange}{rgb}{1.0, 0.55, 0.0}
\definecolor{green}{rgb}{0.13, 0.55, 0.13}
\definecolor{lightgreen}{rgb}{0.56,0.76,0.56}

\title{Can A Single Human Supervise A Swarm of 100 Heterogeneous Robots? }

\author{
Julie A. ~Adams \\
Collaborative Robotics and Intelligent Systems Institute\\
Oregon State University\\
Corvallis, OR 97331 \\
\texttt{julie.a.adams@oregonstate.edu} \\
\And
Joshua Hamell\\
SIFT, LLC \\
Minneapolis, MN 55401 \\
\texttt{jhamell@sift.net} \\
\And
Phillip Walker\\
SIFT, LLC  \\
Minneapolis, MN 55401 \\
\texttt{pwalker@sift.net} \\
}

\begin{document}

\maketitle

\begin{abstract}
An open research question has been whether a single human can supervise a true heterogeneous swarm of robots completing tasks in real world environments. A general concern is whether or not the human's workload will be taxed to the breaking point. The Defense Advanced Research Projects Agency's OFFensive Swarm-Enabled Tactics program's field exercises that occurred at U.S. Army urban training sites provided the opportunity to understand the impact of achieving such swarm deployments. The Command and Control of Aggregate Swarm Tactics integrator team's swarm commander uses the heterogeneous robot swarm to conduct relevant missions. During the final OFFSET program field exercise, the team collected objective and subjective metrics related to the swarm commander's human performance. A multi-dimensional workload algorithm that estimates overall workload based on five components of workload was used to analyze the results. While the swarm commander's workload estimates did cross the overload threshold frequently, the swarm commander was able to successfully complete the missions, often under challenging operational conditions. The presented results demonstrate that a single human can deploy a swarm of 100 heterogeneous robots to conduct real-world missions. 

\end{abstract}

\section{Introduction}

Stated simply, the answer to the title's question is: Yes! The Defense Advanced Research Projects Agency's (DARPA) OFFensive Swarm-Enabled Tactics (OFFSET) program \cite{darpa_offset} created a unique opportunity to investigate a long standing open question related to a single human's ability to supervise a true heterogeneous swarm of robots completing a complex mission in a complex urban environment. This manuscript presents the first human performance results for such real-world swarm deployments. Swarms of this nature have broad future application in domains, such as disaster response (e.g.,\  infrastructure safety inspections, wildland fire identification and tracking) and commercial applications (e.g.,\ general logistics, deliveries). 

The Command and Control of Aggregate Swarm Tactics (CCAST) DARPA OFFSET Program integrator team, led by Raytheon BBN and including personnel from Oregon State University and SIFT, LLC, developed a heterogeneous swarm  to advance and accelerate elements of enabling swarm technologies, focusing on the swarm autonomy and human-swarm teaming \cite{ClarketalMobile2021}. A near-the-battle human supervisor,  the Swarm Commander (SC), deployed the heterogeneous robot swarm using mission plans and SC generated tactics to complete the assigned missions. The SC used the CCAST Immersive Interaction Interface, a virtual reality based system, as the only human responsible for deploying the swarm. 

The OFFSET program incorporated six Field Exercises (FXs) conducted in urban environments. CCAST supports approximately 200 hardware ground (UGV) and aerial (UAV) vehicles (summarized in Table \ref{Tab:Robots}) and 250 simulation vehicles that were deployed throughout the program at United States military urban operations training facilities, or Combined Arms Collective Training Facilities (CACTFs). The missions incorporated either hardware only, CCAST's multi-resolution simulation's virtual, or live-virtual (i.e.,\ hardware and virtual vehicles) swarms. The CCAST system supports hardware and virtual vehicles identically, and the SC interactions are agnostic to the vehicles' instantiation. 

The final field exercise, FX-6, occurred at Fort Campbell's Cassidy CACTF in November 2021. A human subjects evaluation collected performance metrics from the team's two SCs during shift deployments. Given the nature of the CCAST swarm, the SCs must be trained with deploying the swarm and using the SC interface. The evaluation's results support the qualitative evidence generated during the prior field exercises, that a single human SC can achieve the mission deployment and associated mission goals. The SCs' overall workload was assessed based on individual contributors to overall workload. Specifically, a multi-dimensional workload algorithm was used to estimate and continuously classify overall workload based on recorded 
measurements of the cognitive, speech, auditory, and physical workload components that were combined with separate 
visual workload model values.
The SC's estimated overall workload was only classified as an overload state for 3.2\% of the 12,181 usable workload estimates, and the algorithm demonstrated sensitivity to workload changes for this challenging human subjects evaluation environment. 

The background, Section 2, provides overviews of the CCAST swarm implementation, the immersive virtual reality  interface, and the multi-dimensional workload algorithm. The experimental methodology, including important context related to field exercises, and specifically FX-6, are provided in Section 3. An analysis of the evaluation results is provided in Section 4, with Section 5 providing conclusions.

\section{Background}
\label{Background}

\subsection{CCAST System Overview}

\begin{table}[t]
\caption{CCAST's Swarm Robot Platforms} 
\label{Tab:Robots}
\center
\begin{tabular}{|m{1.15in}|m{1.15in}|m{1.15in}|m{1.15in}|m{1.15in}|}
  \hline
\center{\textbf{Aion Robotics R1}}  & \center{\textbf{3DR Solo}} & \center{\textbf{UVify IFO-S}} & \center{\textbf{Modal AI VOXEL M500}} &  \textbf{Modal AI Seeker}\\
 \center{ \includegraphics[width=1.15in]{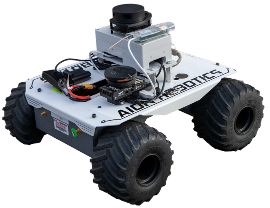}}  & \center{\includegraphics[width=1.15in]{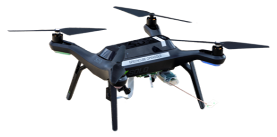}} & \center{\includegraphics[width=1.15in]{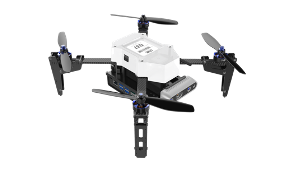}} & \center{\includegraphics[width=1.15in] {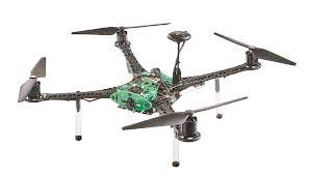}} &  
 \includegraphics[width=1.15in]{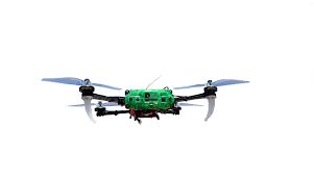} \\
  \hline\hline
\center{ \textasciitilde\$3600 }&   
\center{ \textasciitilde\$750} &   
\center{ \textasciitilde\$3900} &    
\center{ \textasciitilde\$2300} &   \textasciitilde\$2700 \\ \hline
 \center{All FXs} &
 \center{All FXs} &   
\center{FX-4 \& FX-6} &   
\center{FX-6} &    
FX-6   \\ \hline \hline

  Custom expansion board
 & 
 Nvidia Jetson TX2 co-processor
 & 
 Nvidia Jetson Nano co-processor
& 
VOXL companion board
& 
VOXL companion board
\\ 
Raspberry Pi 3B+ co-processor & & & & \\ \hline
 Raspberry Pi Cam
 & 
 Intel RealSense depth camera
 & 
 Intel RealSense depth camera
& 
Stereo camera
& 
Stereo camera
\\ 
 & 
2D spinning lidar
 & 
Downward-facing Raspberry Pi Cam and optical flow 
& 
Forward-/Downward-facing cameras and optical flow 
& 
Downward facing camera and optical flow 
\\ 
 &  &  & & Time-of-flight camera 
\\ \hline
  USB LTE modem
 & 
 USB LTE modem
 & 
 USB LTE modem
& 
Integrated LTE modem
& 
Integrated LTE modem
\\ 
\hline
\end{tabular}
\end{table}

CCAST's heterogeneous autonomous hardware swarm is composed of physically small inexpensive commercial-off-the-shelf vehicles that support large scale swarm operations in small congested areas.  The robots' computational capabilities and payloads differ, as detailed in Table \ref{Tab:Robots}; however, all robots 
can be assigned the majority of the missions tactics. For example, when the mission planner issues a Surveil tactic of the outside of a structure, the assigned UAVs must have the specified number of UAVs with forward facing cameras and a UAV with a downward facing cameras. The UAVs are assigned to the tactic only based on their camera payload position, irrespective of the UAV hardware model. Robots designated for indoor operations (i.e.,\ the Aion UGVs, UVify IFO-S, and Modal AI Seeker) have more expensive and capable payloads that support running computationally complex algorithms (e.g.,\ simultaneous localization and mapping). Otherwise, all UAVs can be assigned the same tasks simultaneously. 

The swarm vehicles maintain a communication link to support vehicle deconfliction and tasking. The individual vehicles communicate, via an LTE network, discovered obstacles and objectives of interest (i.e.,\ artifacts), as well as a telemetry package to a centralized dispatcher that also enables communication with the SC. The LTE communication network requires the vehicles to have line-of-sight to the base station in order to maintain communications, as a result, vehicles are periodically out of communications.

The dispatcher translates the SC's commands, called tactics, into vehicle understandable instructions \cite{ClarketalMobile2021}. If the SC explicitly specifies particular vehicles to execute a tactic, the dispatcher's commands are directed at those vehicles. However, the SC does not have to select specific vehicles, rather the dispatcher can automatically select and assign vehicles with the necessary capabilities that are proximally close to the specified tactic's goal execution location. The dispatcher deconflicts vehicle assignments for some tactics, but other tactics require explicit communication of the assigned vehicles' positions so that the vehicles can deconflict themselves.

The CCAST Swarm Tactics Exchange library incorporates both CCAST developed tactics and tactics developed by external collaborators \cite{PrabhakarRSS2020}. Tactics include surveillance (Surveil) of structures or areas of interest, Cordon, Flocking, agent Following, Exploring the interior of buildings, etc. The swarm robots are assigned tactics, either as individuals, or as a coordinated team. The robots can automatically Swap in order to continue tactic execution when robot (i.e.,\ UAV) battery levels become too low \cite{DiehlDARS2022}. Once a tactic is assigned, the robots conduct real-time navigation planning using extensions to the real-time, rapidly exploring random tree star (RT-RRT*) algorithm \cite{rt_rrt}.

CCAST's Swarm Tactics and Operations Mission Planner is used prior to mission deployments for developing multi-dimensional, multi-phase mission plans. This planner is integrated with CCAST's multi-resolution swarm simulation, which facilitates evaluating and refining the plans. Once the hardware vehicles are staged in the launch area and powered on, the mission plan is instantiated, binding available vehicles on the LTE network to roles or groups. The SC loads the mission plan and either executes the entire mission plan, or portions of (i.e.,\ signals within) a multi-phase mission plan. 

The CCAST team extended Microsoft Research's AirSim \cite{airsim} to provide a multi-fidelity swarm simulator. The simulator facilitates system development, pre-field exercise (e.g.,\ congestion testing) and pre-mission (e.g.,\ mission planning) analysis with larger swarm sizes at a more rapid, cheaper and larger scale.  The simulator capabilities directly support live-virtual swarms composed of hardware and virtual vehicles during field exercise mission deployments. 

CCAST's 3D terrain elevation model, includes obstacles, and is used to generate a spatial database that also includes the swarm vehicles' telemetry information. Telemetry information is considered to be approximate, given the hardware vehicles' known GPS error. 

The DARPA OFFSET program  provided proxies for real world entities using AprilTags~\cite{Olson2011AprilTagAR} that were easy for the vehicle's on-board image analysis tools to sense. The tags are placed on flat vertical and horizontal surfaces around the CACTF, such as outside and inside buildings, or on boxes with AprilTags on the four sides and top. The tags represent artifacts ranging from general navigation hints (e.g.,\ building identifiers, ingress markers), non-combatants, hostiles, coded intelligence, and high-value targets. Some artifacts are \textit{active} and can interact with the vehicles, or vice versa, via Bluetooth. For example, a hostile or an explosive device can neutralize a vehicle before the vehicle neutralizes the hostile or explosive\footnote{Neutralization causes a vehicle to stop its tactic execution. The vehicle cannot execute tactics towards the mission objective until it is revived by a medic, but can be commanded to move about the CACTF.}. The imaging payload is used to recognize the AprilTag identifier that is matched via a look up table to the corresponding artifact, which triggers any necessary vehicle responses.

\begin{figure}[htb]
\centering
\begin{subfigure}{.49\textwidth}
  \centering
   \includegraphics[width=.9\linewidth]{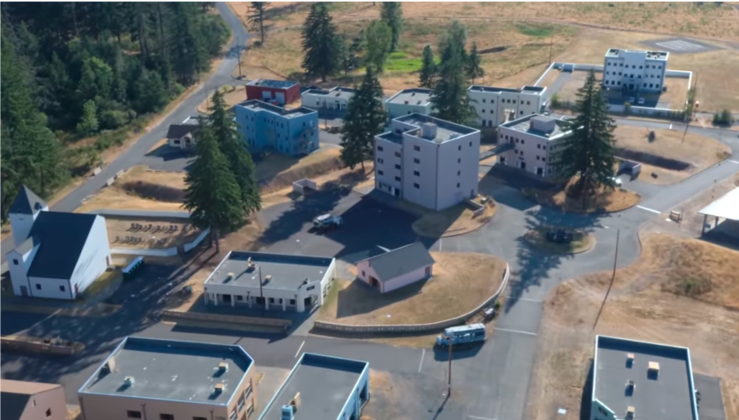}
    \caption{A common FX built environment.}
    \label{Fig:JBML}
\end{subfigure}%
\hfill
\begin{subfigure}{.49\textwidth}
  \centering
   \includegraphics[width=.9\linewidth]{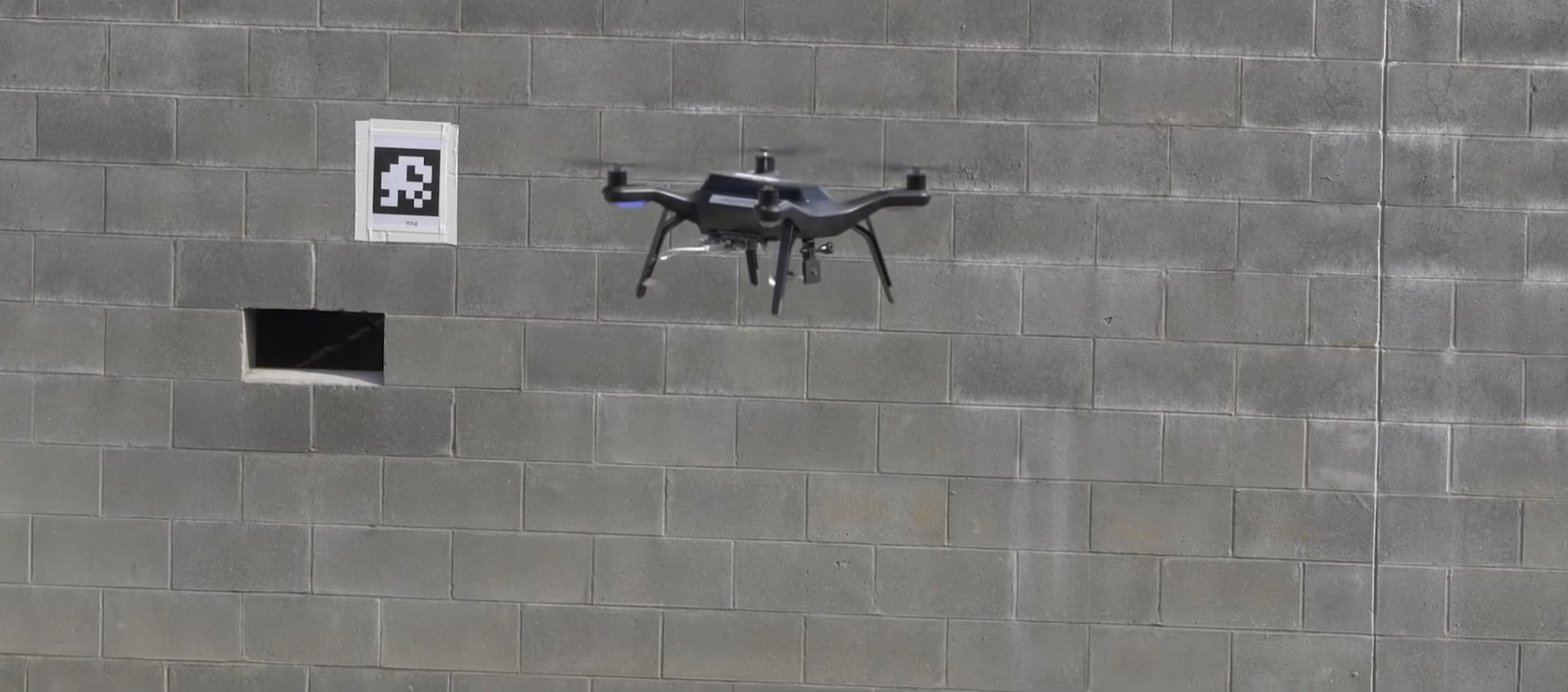}
    \caption{A CCAST UAV reading an AprilTag.} 
    \label{fig:SoloReadingTag}
\end{subfigure}%
\caption{(a) The FX-4 Joint Base Lewis-McChord CACTF. (b) A UAV conducting a building Surveil. Photos courtesy of DARPA.}
\end{figure}

The programs' constraints and the shear expected swarm size, combined with the built urban deployment environments created unique challenges from a robotics perspective. The missions required autonomous robots capable of navigating while avoiding obstacles and power lines, collecting intelligence, and responding to artifacts using Bluetooth that require, in some cases close proximity, which are challenging objectives for more advanced robots, even more so for the CCAST swarm. The CACTFs' presented common built environment challenges, such as curbs, steps, barriers, street signs, and power lines, as shown in Figure \ref{Fig:JBML}. UGVs can leverage the road network, but the CCAST 3D terrain elevation model also provided necessary context regrading obstacles, such as barriers, steps, and drainage ditches. While the UAVs autonomously ascended to a safe flight altitude, above buildings, trees, and other obstacles for autonomous enroute navigation to tactic goal locations, the UAVs frequently performed autonomous tasks at lower altitudes within the built environment, which required avoiding tress, bridges between buildings, and power lines. A common phase I mission plan tactic was to autonomously Surveil the CACTF's buildings' exteriors to collect intelligence. 
The UAVs with forward facing cameras descending into the built environment in order to detect and classify the AprilTags on the sides of buildings, such as in Figure \ref{fig:SoloReadingTag}. The FX-6 mission scenario necessitated the need to use multiple UAVs to descend and interact via their Bluetooth beacons with active artifacts on the ground, an achievable, but also challenging accomplishment. 

More specifically, the autonomous swarm robots are assigned the same tactics per either the mission plan or the SC. A typical Phase I mission plan incorporated multiple Surveil tactics to gather information to inform specific Phase II mission plan actions (e.g.,\ searching a specific location for additional information). All robots gathered the information throughout the CACTF. UAVs with forward facing cameras typically gathered information on the sides of structures, those with downward facing cameras gathered information on the tops of structures or that were flat on the ground, while UGVs gathered some information from structures as well as 3-D artifacts on the ground. UGVs conduct tactics in accessible buildings (e.g.,\ doorways that UGVs can drive to), but the Uvify IFO-S and Modal AI Seeker UAVs were integrated to extend the swarm's access to buildings inaccessible to the UGVs. All robots had either an electronic payload (e.g.,\ disabling improvised explosive devices) or an anti-personnel payload (e.g.,\ secure adversaries). Any UGV or UAV with an electronic payload was able to disable the adversaries' electronic systems, similarly any UGV or UAV with an anti-personnel payload was able to secure an adversary. Some active artifacts required simultaneous interaction by multiple robots with the necessary payload combinations. The Phase II mission plan typically incorporated taking action based on the Phase I intelligence to locate the high valued target, while a Phase III mission plan focused on securing that target. Neutralized UGVs autonomously navigated to a known medic, while neutralized UAVs autonomous returned to the launch zone and landed. Once revived by a medic, the robots either continued their prior tactic or were assigned a new tactic.  

\subsection{Immersive Interaction Swarm Commander Interface Overview}
\label{I3Over}

Real world command and control of heterogeneous swarms requires exploring new interface and control concepts.  The CCAST swarm command necessitates a control system in which a single operator can efficiently task hundreds of robots, while maintaining awareness of the environment.  OFFSET focused on urban operations across multiple urban blocks. The Immersive Interactive Interface (I3) is the CCAST team's solution \cite{Walkeretal2023}.  Traditional command and control stations often rely on a two dimensional top-down map views annotated with entity and tasking symbology, 
which cannot adequately support the types of missions and tactics required to conduct the OFFSET mission. The OFFSET swarm missions challenged CCAST to move away from these traditional control systems in order to accommodate:
\begin{itemize}
    \item Swarm groupings, a fluid concept, potentially representing a collection of mixed capability robots.
    \item Verticality, critical when expressing urban terrain, especially for multi-story structures.
    \item The sides of structures, critically important for tactical tasking in urban environments.
    \item The volume of occupied space, especially along the vertical axis. 
    \item Multiple perspective inspection of scenario elements, 
    in terms of raw viewpoint and level of detail/abstraction.
\end{itemize}
The SC is assumed to be ``near-the-battle'' with reliable, low latency data links to the battle field.  The SC did not have line of sight observability of the swarm or urban environment. 
The swarm's UGVs and UAVs composition necessitates the SC's simultaneous viewing of both robot types. Further, since robots were deployed in the urban environment and entered buildings, the SC needed 
different viewing perspectives, including the swarms' egocentric perspective. The observability of the vehicles and artifacts can be obstructed by three dimensional virtual CACTF structures; however, the perceived benefits to the SC's overall awareness, including spatial awareness and the ability to precisely localize vehicles and artifacts, outweigh the negatives often associated with immersive interfaces. 

I3's virtual reality interface is built within the Unity game engine and leverages SteamVR and the Valve Index hardware system \cite{Walkeretal2023}.  The virtual reality places the SC directly
in the virtual battle space, enabling the SC to inspect and interact with the swarm at varying detail and control levels. The SC is assumed to be in a dedicated command center ``near-the-battle'', not physically in the battle field. The SC's laptop is connected to
the swarm control network, but is positioned in a physically suitable environment that supports safe usage of the virtual reality hardware.  I3 receives live (or low latency) telemetry from all vehicles via the dispatcher, while the SC issues commands in the form of tactics and mission plan engagements. These aspects help to minimize the impacts of virtual reality induced motion sickness. 

The SC's Valve Index head mounted display provides a three dimensional perspective.  The two Valve Index handheld controllers are used to inspect, interact with, and navigate the virtual world.  The Valve Index chest tracker enables separate reference frames for the head and body, which supports virtual side panels.  The system relies on outside-in virtual reality tracking; thus, two tripod-mounted tracking beacons are used during field exercise deployments.  

\begin{figure}[htb]
\centering
\begin{subfigure}{.65\textwidth}
  \centering
   \includegraphics[width=.8\linewidth]{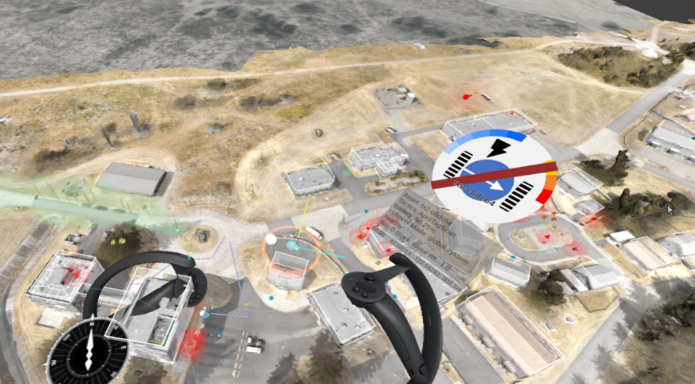}
    \caption{Visualization of the I3 sand table during FX-4.}
    \label{Fig:swarmtable}
\end{subfigure}%
\hfill
\begin{subfigure}{.35\textwidth}
  \centering
   \includegraphics[width=.65\linewidth]{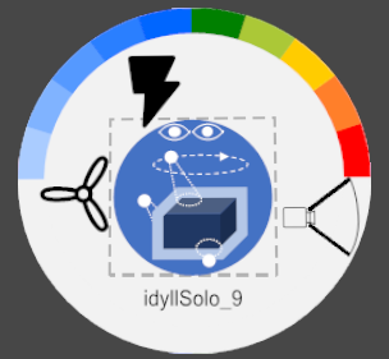}
    \caption{An example UAV glyph.} 
    \label{fig:glyph}
\end{subfigure}%
\caption{(a) A sand table representation, including  a neutralized UGV's glyph. (b) A UAV glyph (indicated by the propeller), where the top bars indicate communication connectivity (blue) and battery level (multi-colored), the lightning bolt indicates an electronic warfare payload, the forward facing camera, the central icon indicates the task currently being executed, and the gray dashed box indicates the UAV is virtual.}
\end{figure}

I3's virtual world began with a sand table concept \cite{Walkeretal2023}, see example in Figure \ref{Fig:swarmtable}.  The sand table can
support rapid perspective transitions, multimodal interaction, and unique visualization options unavailable elsewhere. The SC can manipulate the world space, effectively transforming the world around them, both in terms of navigation and interaction with proxy elements to engage real-world behaviors. The sand table is built upon a hierarchy of transformations, permitting the SC to manipulate rotation, scale, and translation, while still maintaining spatial relationships between modeled elements.  Given the OFFSET program's field exercise
locations' scale, it was sufficient to treat coordinate translation as a mapping between Latitude, Longitude, and Altitude (mean sea level) into an XYZ reference frame defined in meters.  Static world elements defined the operational environment.

The provided world map is composed of several layers that include a digital elevation model, human-defined and named obstacle and building boundaries, and a photogrammetry generated object model \cite{Walkeretal2023}. During FX-6, externally generated building floor plans were integrated as geo-rectified images, enabling the SC to
inspect the idealized interior of buildings while controlling the swarm.

The Valve Index hand controllers are a primary I3 input mechanism, with each hand assigned a controller.  I3 is sensitive
to the controllers' position in the world space, which permits accurate interaction contexts. The controllers'  haptic feedback provides a cue to important events and supplements the corresponding visual context. The head mounted display's position and orientation within the virtual space are used to recognize the central view axis.  Audio cues indicate incoming important information, such as the detection or neutralization of a hazard and neutralization of CCAST vehicles. Text-to-speech provides notifications of tactic failures (e.g.,\ ``Surveil failed'').

I3 shifts the world around the user during virtual world navigation.  The left hand controller, while the trigger is pressed, supports scaling (i.e.,\ thumb tracker slide), rotating (i.e.,\ joystick) and translating (i.e.,\ moving the controller) the world.
I3 also supports scenario-defined sand table transformations, which effectively map into saved viewpoints.  These capabilities facilitate changes in the visual display of the environment, and no very large viewport changes occur without an explicit action taken by the SC. This interaction requirement provides a means of reducing virtual reality induce motion sickness.

Various world model elements and entities are visualized. A priori static entities (e.g.,\ buildings and obstacles) as well as dynamic entities are populated in the virtual space to represent both physical (e.g.,\ vehicles) and synthetic (e.g.,\ tactics) concepts.  These entities are mapped to the sand table and each entity's visualization depends on its internal state, SC interactions, and a distance-based level of detail capability. Some static objects (e.g.,\ buildings) have identifiers the SC can use when issuing tactics, which can simplify explicit tasking and provide important information to the swarm (e.g.,\ ingress points).  

\begin{figure}[htb]
\centering
\begin{subfigure}{.3\textwidth}
  \centering
   \includegraphics[width=.9\linewidth]{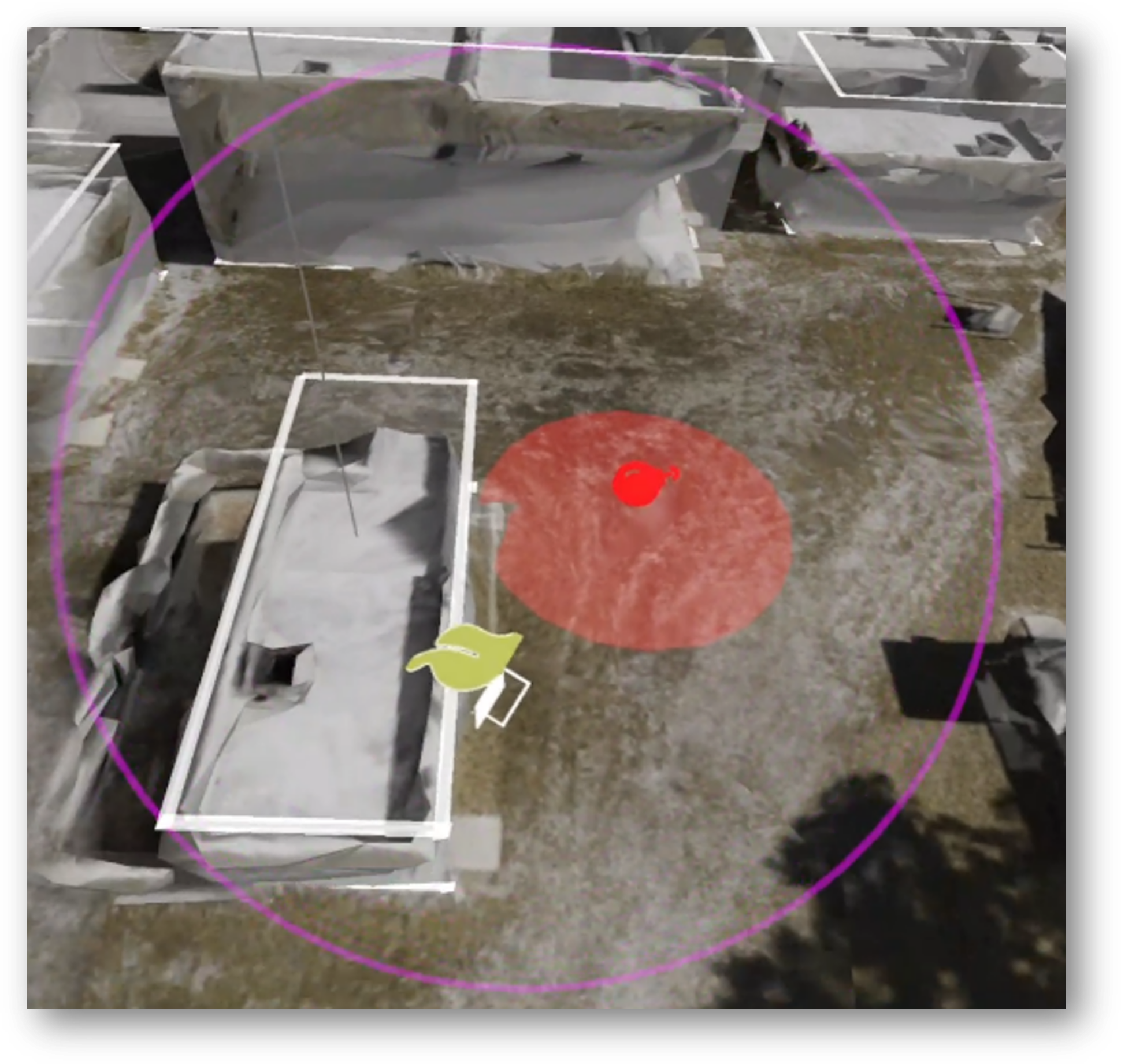}
    \caption{Artifact and threat ring. }
    \label{fig:threat}
\end{subfigure}%
\hfill
\begin{subfigure}{.7\textwidth}
  \centering
    \includegraphics[width=.8\linewidth]{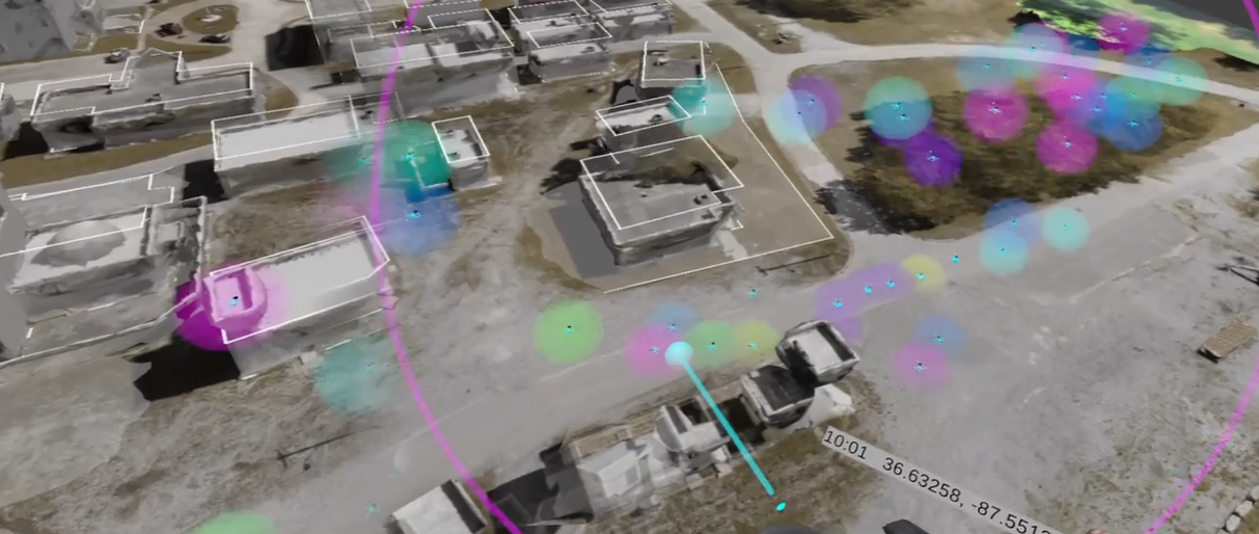}
    \caption{Swarm visualization.}
    \label{fig:swarm-viz}
\end{subfigure}%
\caption{Example of (a) an improvised explosive device artifact and associated threat ring, and (b) swarm visualizations, where six sub-swarms are beginning to execute a mission. Note, the fuchsia ring represents the spatial area the SC sees within the head mounted display.}
\end{figure}

 Representations of the AprilTags (i.e.,\ artifacts) or the entities (e.g.,\ swarm, vehicle, artifact) have customized visualizations in the sand table with which the SC can interact. For instance, when a hostile adversarial artifact is recognized, the tag identifier and pose estimate are communicated to I3, which maps the coordinates to the virtual world space, displays the appropriate icon, adds a type-specific threat ring representing the range at which the hostile can interact with the swarm vehicles, and places it into the table, as shown in Figure \ref{fig:threat}. The very large field exercise scenarios prohibited visualizing all entities, particularly those that were not necessary to support SC's situation awareness and interactions. The capability to enable or disable entity classes via toggling them was implemented.  

The SC typically conducts the mission by interacting at the swarm level \cite{Walkeretal2023}. Individual vehicles are synthesized into swarm groups based on shared tactics, irregardless of whether the assignment derives from the mission plan or SC specified tactics. For example, vehicles assigned to a tactic, see Figure~\ref{fig:swarm-viz}, are designated as a swarm group, with a common shade, and have a reference handle for tactic manipulation. Individual vehicles are represented with a generic object model corresponding to their type (i.e.,\ UAV, UGV) that can represent both hardware (i.e.,\ live) and virtual vehicles. 

\begin{figure}[htb]
\centering
\begin{subfigure}{.49\textwidth}
  \centering
   \includegraphics[width=.9\linewidth]{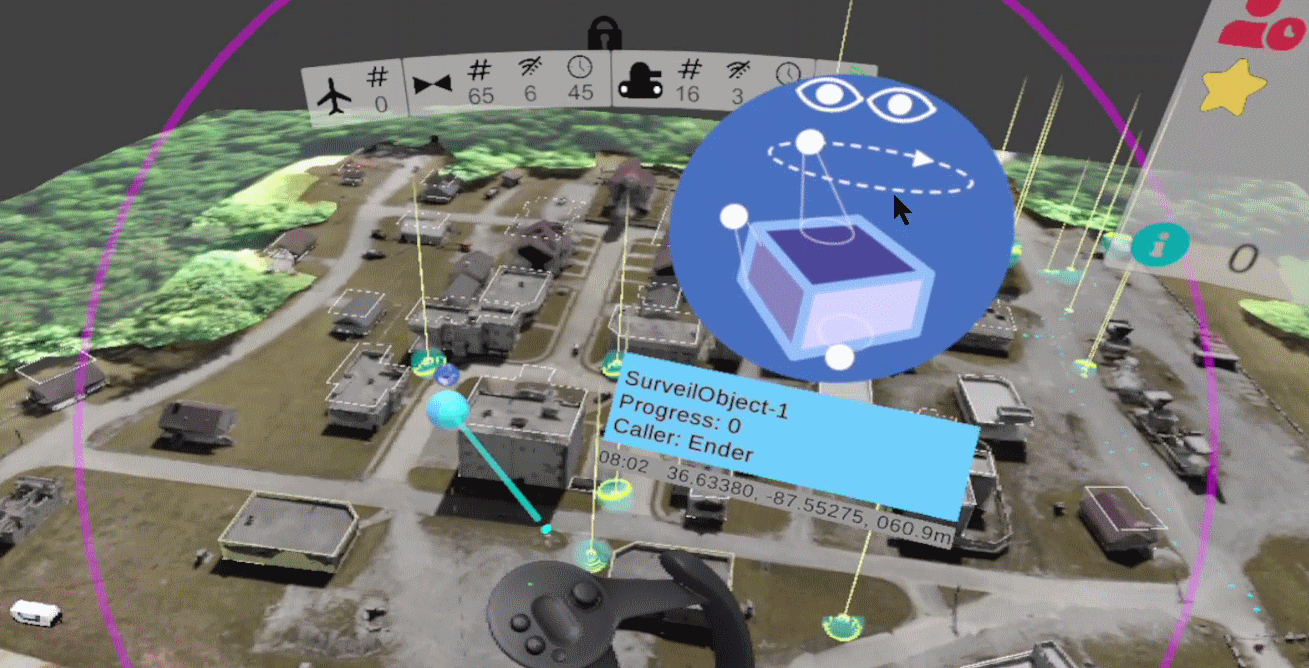}
    \caption{Example tactic visualization.}
    \label{Fig:tactic}
\end{subfigure}%
\hfill
\begin{subfigure}{.49\textwidth}
  \centering
    \includegraphics[width=.8\linewidth]{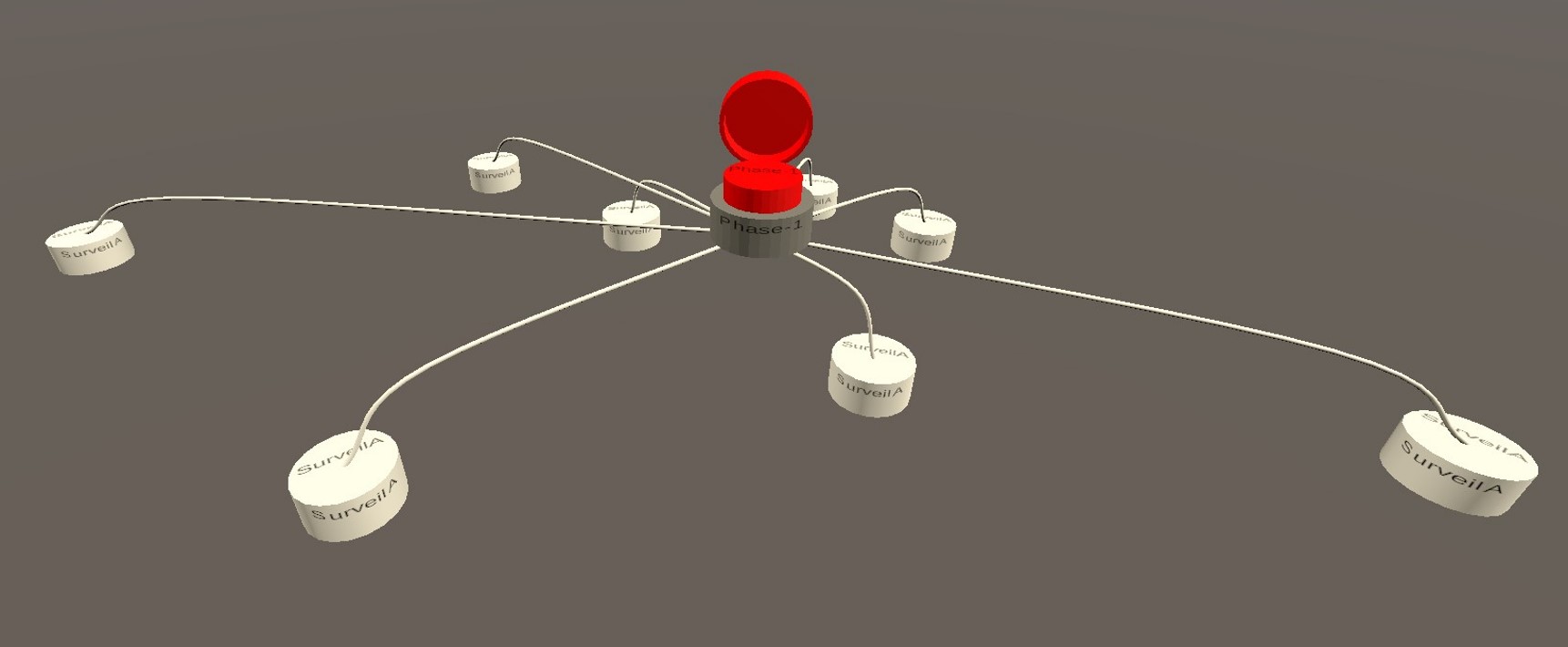}
    \caption{A mission plan visualization.}
    \label{fig:mission-plan}
\end{subfigure}%
\caption{(a) Example tactic and (b) mission plan visualizations, where the mission plan contains several nodes (white disks) gated by a single signal (red button with raised cover).}
\end{figure}

Entities can be inspected using a context-aware system. The right hand controller recognizes when the cursor intersects with an entity and constructs a summary glyph, see the example in Figure \ref{fig:glyph}. The glyph communicates the vehicle's type, payload, remaining battery level, if up-to-date vehicle telemetry is being received, current tactic, and if it is a hardware or virtual vehicle. 
The vehicle's planned navigation route is also displayed. 
The DARPA OFFSET scenario can cause vehicles to be neutralized; thus, the vehicle representation changes to indicate a neutralized status, as shown in Figure \ref{Fig:swarmtable}. The hazard (e.g.,\ hostile) summary glyph is similar, but a line is visualized to the tasked vehicles on hover. Hovering over a tactic summary visualization  highlights the associated vehicles or swarms, as shown in Figure \ref{Fig:tactic}.

The I3 SC can create dynamic geometry by entering an input mode that permits specifying a point, a polyline, a polygon, or an extruded polygon.  The right hand controller is used to specify discrete vertices and, optionally, a depth for extruded geometries. The resulting geometries can be used to explicitly specify swarm or individual vehicle tactics. For example, a polygon defining the area 
to Surveil.  

The SC can display a menu system around the right controller's interaction location that facilitates interactions at the world level (e.g.,\ visualization toggles, tactics menu) and context-sensitive queries or tactics. This menu placement allows the SC to maintain attention on the relevant information during its use. The menus are nested arbitrarily deep, contain custom icons and visualizations, support multiple widget types, and for explicit hand controller buttons, supports both long and short click behaviors. The primary menu is accessed by pressing the right hand controller's `\texttt{A}' button and provides all available I3 actions (i.e.,\ visualization toggles, geometry creation menu, tactics menu, and mission plan controls). The context menu supports query or engage behaviors, based upon what is in proximity of the controller's cursor.  Most entities can be interacted with (i.e.,\ buildings, artifacts, vehicles, swarms, tactic visualization nodes, and mission plan elements). The initial menu row contains references to applicable, possibly multiple entities, that  based on a threshold, are identified relative to the interaction point. This approach permits quick selection within a sparse location and enables interactions in  dense locations. The context menu is typically used for explicit tactic invocation.  For example, interacting with an at altitude UAV to specify that it return to the launch (RTL) area immediately.  A tactic can also be specified by interacting with a building element and requesting an immediate Surveil tactic, which may allow the dispatcher to auto-allocate suitable vehicles, or transition into the tactic calling menu with the pre-specified  building as the tactic's target.  

\begin{figure} [h]
    \centering
    \includegraphics[width=.5\linewidth]{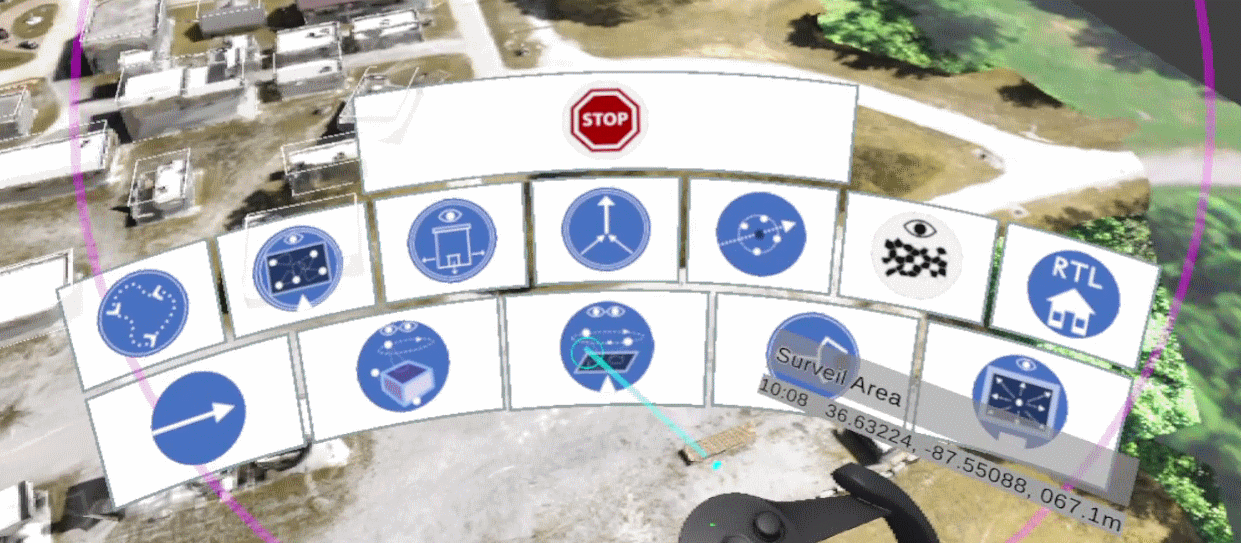}
    \caption{The FX-6 tactics menu.}
    \label{Fig:TacticsMenu}
\end{figure}

The CCAST system facilitates a large number of tactics. A complete description of those tactics is beyond this manuscript's scope; however, a tactic defines a behavior to be performed by the allocated vehicles, along with optional navigation and execution parameters.  Tactics may explicitly reference vehicles, or the dispatcher may autonomously allocate vehicles based on required capabilities and physical proximity.  The tactics menu is customized prior to an FX to provide the most relevant tactics, which are filtered by use case, see the FX-6 tactics menu in Figure \ref{Fig:TacticsMenu}.  Three different agent specification levels exist.  The most granular is the vehicle level, at which explicit vehicle(s) call sign(s) are provided.  The next level uses swarm labels that cause any vehicle with the specified label to accept the tactic.  The final level specifies (or accepts default) ``wildcard'' values that leaves the vehicle selection to the dispatcher. 

Tactics are called by I3 using three different mechanisms \cite{Walkeretal2023}. The SC can \textit{explicitly} select the tactic and vehicle(s), which provides  great control, but costs execution time. The context menu can be used to identify vehicle(s) or a target for which the menu system \textit{pre-seeds} one or more of the tactic fields, which the operator can refine or reject, but at the very least must manually engage the tactic. The context menu can also be used to \textit{instantly} execute a tactic without specifying the vehicles or other details. For example, to execute a simple Stop or RTL command on a vehicle, or an automated execution of a Surveil tactic on a building. Each called tactic has a visualization (see Figure \ref{Fig:tactic}) that indicates the tactic type and any associate geometry (e.g.,\ search area). This visualization changes as the top level tactic moves through its lifecycle, eventually disappearing upon tactic completion.

The CCAST mission plan is critical to achieving the mission objectives. Mission plans are developed a priori and the I3 SC loads the plan from a centralized repository.  Mission plans contain nodes, each containing one or more tactics.  Tactics may begin at mission start, commence upon explicit SC or software issued signals, or execute upon completion conditions asserted by predecessor tactics. The mission plan is visualized above the sand table as a hierarchical tree (see Figure~\ref{fig:mission-plan}). The plan tree contains top level signals, and subsequent levels conforming to the tactic completion dependencies.  The physical  signal and tactic nodes' positions are generated from the centroid of deconflicted associated tactic geometries. The SC can trigger signals relative to the mission plan that gate the execution of one or more mission plan nodes. Typically, a mission has multiple phases, represented by the gated signals, that permit triggering the scenario phases as the SC determines conditions are suitable.  For instance, most FX mission plans involve an initial series of Surveil tactics deconflicted by region to reduce the risk of mid-air UAV collisions during navigation to or from the launch area.  Each region has an independent discrete signal. The SC engages the signals by interacting with the associated nodes.  Hovering the right controller over a specific mission plan node causes the vehicles and geometries associated with the specific sub-tactics to be highlighted within the sand table.

\begin{figure}[htb]
\centering
\begin{subfigure}{.49\textwidth}
  \centering
   \includegraphics[width=.9\linewidth]{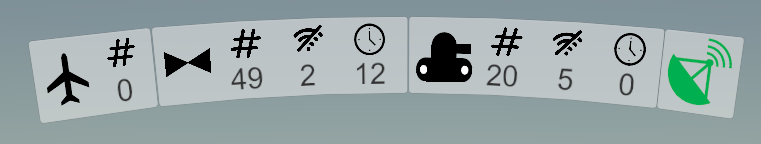}
    \caption{The heads up summary display.}
     \label{fig:vr-hud}
\end{subfigure}%
\hfill
\begin{subfigure}{.49\textwidth}
  \centering
    \includegraphics[width=.9\linewidth]{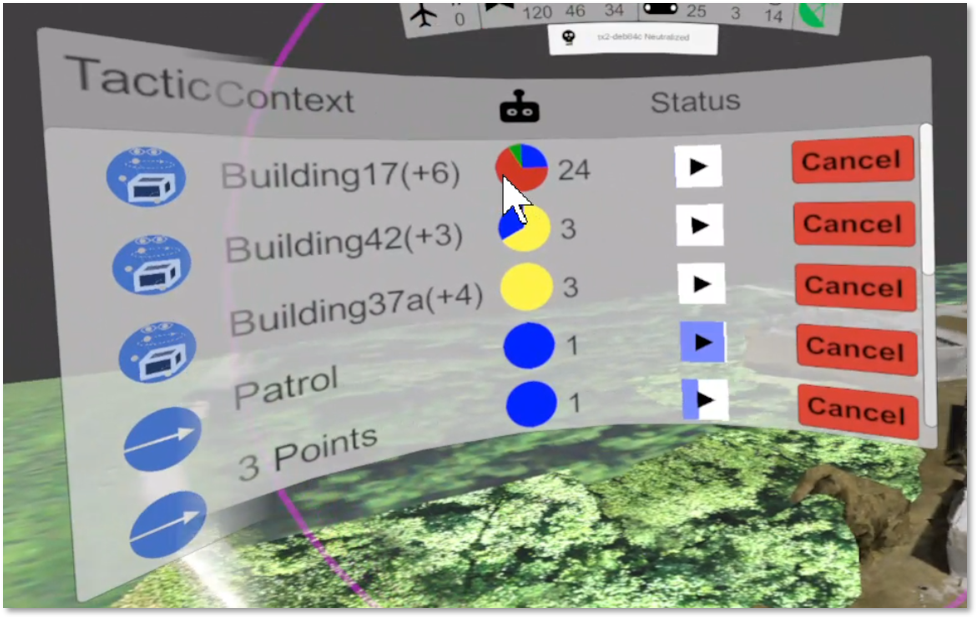}
    \caption{The tactics side panel.}
    \label{fig:TacticsPanel}
\end{subfigure}%
\caption{An example (a) heads-up display indicating 49 UAVs and 20 UGVs, and (b) a tactics panel showing five tactics. Both are displayed relative to the SC's focus to support quick viewing.}
\end{figure}

Always-available information is provided via a heads-up display positioned relative to the SC's viewpoint, as shown in Figure~\ref{fig:vr-hud}, that incorporates current vehicles' telemetry status, which can indicate communication issues, and is constantly updated with available vehicle counts by type.  A notification pane provides critical information, including new scenario intelligence sightings or vehicle neutralizations.

The Valve Index's chest tracker provides an inertial frame for displaying at-a-glance tactic and scenario panels to the SC's sides.  The tactics panel, displayed on the SC's left side, lists tactics being executed, including a notion of tactic type, target, composition, and state.  Selecting a tactic facilitates terminating it. 
The panel on the SC's right side lists hazard and artifacts of interest, emphasizing threat status, including whether or not they have been addressed. 

The deployment of the CCAST swarm using I3 is sufficiently complex as to require a minimal level of training. This training can be used to vet the susceptibility of potential SCs to virtual reality induced motion sickness. Throughout the OFFSET program many untrained or minimally trained individuals used I3 via the virtual swarm capabilities with no ill effects.

\subsubsection{Typical Mission: Swarm Commander's Perspective}

The SC, during a typical shift, executes at least one mission plan and specified tactics. At shift start, the SC loads the mission plan and when all systems are ready, requests either specific mission plan signals, or the entire plan be executed. Once the assigned vehicles clear the launch area, the SC often begins issuing  tactics to the remaining vehicles. The initial phases seek to gather intelligence and identify important information (e.g.,\ locations of high valued targets or the medic) needed to execute the mission. As UGVs encounter adversaries and are neutralized, they autonomously navigate to a medic, if it has been located. Otherwise, the neutralized UGVs RTL. Neutralized UAVs autonomously RTL. UAVs that have completed their tactics will also automatically RTL, while UGVs will wait in place for a new assignment. 

As intelligence is gathered, the SC can customize the swarm's response to act on the information. For example, if the swarm finds information about a high value target's location, the SC may send vehicles to that location to investigate. The mission plan often includes phased signals intended to respond to the gathered intelligence that can facilitate continued intentional mission progress, including mission phases. A typical second mission phase acts on the gathered intelligence to localize a high value target, while the next phase focuses on neutralizing that target.  

During FX-6, the DARPA provided scenario quickly neutralized large numbers of vehicles. Thus, a mobile medic was introduced at the launch area to revive neutralized UAVs. The mobile medic required a human to walk through the launch zone with a Bluetooth-enabled device that revived the vehicles, which, for safety purposes, required waiting until a large number of the UAVs had RTL'ed. The long CACTF shifts require the UAV batteries to be swapped during the mission execution, which was completed by humans. 

\subsection{Multi-Dimensional Workload Estimation}

Human supervisors (e.g.,\ CCAST's SC) may experience erratic workload levels \cite{Wickens2004,Sim2008}, where performance tends to decline when workload is too high (overload) or too low (underload) \cite{Wickens2004}. An increase in overall workload does not necessarily mean task performance will decrease, as performance depends on the human's overall resources and if there are competing resources (e.g.,\ multiple tasks requiring human visual attention). Overall workload is frequently assessed as a discrete measurement of cognitive workload (e.g.,\ \cite{Kaber2004,Schwarz2018}). However, overall workload can be decomposed into workload components (i.e.,\ cognitive, auditory, visual, speech, and physical workload \cite{McCraken1984,Mitchell2000}) in order to provide necessary insight into the factors contributing to the human's current workload state. 

Non-invasive wearable devices can collect objective workload metrics (e.g.,\ heart-rate variability), whose values have been found to correlate with one or more workload components (e.g.,\ \cite{Harriott2013,HarriottBufordZhangAdamsJHRI15,HarriottPhD15}). Recent workload assessment algorithms have combined these objective metrics into a workload component classification using machine-learning techniques (e.g.,\ \cite{Durkee2016,Popovic2015}). These algorithms typically rely on metrics that are not viable for dynamic domains (e.g.,\ eye-tracking, EEG) to classify cognitive workload, and tend to focus on only the normal load and overload classifications. The algorithms do not discern the state of the other workload components and fail to adequately classify the underload condition. The discrete classifications also do not allow for understanding workload trends (i.e.,\ increasing, decreasing, or unchanged). 

Reviews of relevant metrics exist, but do not address all of the problem's aspects \cite{Harriott2013,HarriottPhD15,CharlesNixonTIE2019}. Heard and Adams reviewed relevant metrics and algorithms for assessing overall workload and its components \cite{Heard2017}. None of the workload assessment algorithms estimated each workload component and classified both the underload and overload workload states using workload metrics collected from wearable devices suitable for the DARPA OFFSET domain.

Many workload assessment algorithms rely on metrics collected from EEG headsets \cite{Bian2019,Durkee2016,GUPTA2021103070}, cameras ~\cite{BloosetalAMIA2019,Heard2019EMBC,ParisetalSPIE2019}, motion capture~\cite{KubotaICRA2019}, dedicated interaction systems (i.e.,\ keyboards~\cite{OliveretalMi2002,Popovic2015}, and smartphones~\cite{RonaotetalESA2016}) to infer the human's cognitive workload. Typically, those systems only infer the normal and overload workload states.
A critical aspect of such systems is their inability to adapt to most unstructured, dynamic environments.

A relevant tree classifier assessed overall workload accurately~\cite{Rusnock2015}, but did not include multidimensional workload components. The MBioTracker is a multimodal wearable system designed to detect workload, but only classifies cognitive workload \cite{DellTBCS2021}. A closely related approach used proprietary algorithms to classify cognitive, visual, auditory, speech, and physical workload, but did not 
estimate overall workload~\cite{Popovic2015}.

\subsubsection{Multi-Dimensional Workload Algorithm Overview}
\label{WorkAlg}
Heard and Adams' multi-dimensional workload algorithm estimates a human's workload components and the composite overall workload state \cite{Fortune2020HFES,Heard2019HFES,Heard2019thri,Heard2019jcedm,HeardDiss2019}. This algorithm was developed specifically to support unstructured dynamic domains (e.g.,\ disaster response, military) using primarily wearable, non-vision based sensors that can objectively measure the human's current performance (e.g.,\ overall workload \cite{Heard2019jcedm,Heard2019thri}). The multi-dimensional workload component states (i.e.,\ auditory, cognitive, physical, speech, and visual \cite{McCraken1984}) are estimated and are used to estimate and classify overall workload (i.e.,\ underload, normal load, and overload). The algorithm incorporates objective physiologically-based metrics, available via wearable sensors, and a non-physiological environmental metric that correlate to overall workload and the multidimensional components \cite{Fortune2020HFES,Harriott2013,HarriottBufordZhangAdamsJHRI15,Heard2017,Heard2019jcedm,Heard2019thri,Heard2019HFES}. 

The multi-dimensional workload algorithm  estimates overall workload and its components by extracting time-based features (i.e.,\ mean, variance, average gradient, and slope) from thirty second epochs for each objective workload metric (e.g.,\ heart-rate variability, posture magnitude, noise level). The time based features serve as inputs to a corresponding neural network that estimates each workload component \cite{Fortune2020HFES,Heard2019jcedm,HeardDiss2019}. The means and standard deviations capture the metrics’ response to workload variations, but do not capture a metric’s directional shift, (e.g.,\ the metric is increasing over the time window). The average gradient and slope features capture this directional change. Slope is the linear change over the window, while the gradient is the average change between each second in the window. 

The multi-dimensional workload assessment algorithm was trained and validated using IMPRINT Pro workload models 
\cite{Heard2019thri}. IMPRINT Pro \cite{Archer2005} supports modeling complex task networks that designate start and stop
times for each task and anchors each task to workload component values (i.e.,\ a conversation is anchored to a speech workload component value of 4.0). The task networks and workload component values are used to derive continuous models across seven workload components: auditory, cognitive, visual, speech, gross motor, fine motor, and tactile. The approach in this manuscript combines the gross motor, fine motor, and tactile components into a physical workload component. An overall workload model is generated by uniformly aggregating the workload component models. IMPRINT Pro uses a linear workload model incorporating the workload components to classify a predicted overall workload, where $\geq 60$ as overload. IMRPINT Pro does not provide an underload threshold. Specific IMPRINT Pro models must be developed to represent the underload, normal load, and overload conditions. The resulting IMPRINT Pro workload models represent predicted workload outcomes, are static, and do not adjust in real-time to the current situation. These modeling constraints limit considerably the ability to use IMPRINT Pro in uncertain and dynamic environments; thus, the need for using the developed multi-dimensional workload assessment algorithm that was shown to have generalizability between task domains and environments (Heard et al., 2019b).

Dynamic environments contain time-varying contributions from multiple workload components and contextual features capture these time-varying workload contributions. Contextual features calculated from the IMPRINT Pro workload models are required by the multi-dimensional workload algorithm to produce more accurate workload estimates. Three contextual features exist: cognitive task composition, physical task composition, and auditory task composition, where task composition represents how much the respective workload component contributes to the human’s overall workload. Speech task composition is not included as a contextual feature, due to using voice activity detection to determine if the human is speaking or not. These contextual features can be set to zero for an unfamiliar environment. Given that the multi-dimensional workload algorithm was trained using the supervisory-based IMPRINT Pro's calculated contextual feature values, the respective values are set to zero for the OFFSET FX-6 workload estimates.

The multi-dimensional workload algorithm estimates the cognitive, auditory, and physical workload components every five seconds. The speech workload component is estimated every second and is resampled to a five second frequency before estimating overall workload. A separate neural network exists for each workload component. Visual workload is estimated using a relevant IMPRINT Pro model. The component estimates are uniformly aggregated to estimate overall workload, which was mapped to a state (i.e.,\ underload, normal load, or overload) using thresholds. Heard et al.\ conducted extensive validation of the multi-dimensional algorithm across supervisory and peer-based relationships, tasks, workload conditions and populations \cite{Fortune2020HFES,Heard2019HFES,Heard2019thri,Heard2019jcedm,HeardDiss2019}.  These validations used IMPRINT Pro models, developed prior to  conducting human subjects evaluations, as the comparison to the multi-dimensional workload algorithm's results. Separate IMPRINT Pro models were developed for the underload, normal load, and overload conditions in order to support the validation efforts. 

It is well known that some physiological metrics (e.g.,\ heart rate, respiration rate) are impacted by other human performance factors (e.g.,\ stress). The multi-dimensional workload algorithm mitigates the impacts of other performance factors in two ways. It is common in the literature to equate overall workload and cognitive workload. Rather, the multi-dimensional algorithm estimates overall workload based on the individual workload components, where the individual workload components use different sets of metrics to estimate the corresponding component's workload value. This approach decreases the influence of a particular metric that may be influenced by another human performance factor. The cognitive and auditory workload components also incorporate noise level, as measured with a noise meter, a non-physiological metric. The second factor that contributes to mitigating the impact of potentially confounding factors (e.g.,\ stress) is the incorporation of the time-based directional change features (i.e.,\ average gradient and slope) that ensures the algorithm does not rely solely on a metric's overall magnitude. 

Underload, normal load and overload IMPRINT Pro models were developed for a supervisory-based adaptive human-robot teaming architecture \cite{Heard2020CogSima}. The corresponding human subjects evaluation incorporated a physically expanded version of the NASA Multi-Attribute Task Battery (NASA MATB-II) \cite{Comstock1992}. The physically distributed NASA MATB-II simulated supervising a remotely piloted aircraft, incorporated four tasks: tracking, system monitoring, resource management, and communications. Each task was distributed to different monitors, two of which required physically walking around a table. Workload was manipulated by changing various parameters of each task in order to determine the adaptive teaming system's effectiveness. The adaptive architecture was shown to select an appropriate level of autonomy or system interaction based on real-time workload estimates from the multi-dimensional workload algorithm, and resulted in improved overall task performance \cite{Heard2020CogSima}. 

The nature of the FX-6 deployments do not support developing a priori IMPRINT Pro models to support an analysis similar to the prior analyses. The goal for the DARPA OFFSET program was to leverage an existing model to generate estimates for the SC during mission deployments. Therefore, the neural network models developed for the supervisory-based adaptive human-robot teaming architecture validation \cite{Heard2020CogSima} were used to provide the multi-dimension workload component estimates and the overall workload estimates for FX-6. Heard et al.\ calculated overload and underload thresholds previously using prior multi-dimensional workload algorithm results and their underload, normal load, and overload models. The overload threshold was found to be 60, which matches IMPRINT Pro's threshold, and the underload threshold was determined to be 25. These thresholds are used in this manuscript to classify workload states.

\section{Method}
The human subjects evaluation's purpose was to understand a single SC's ability to conduct missions using I3 and the swarm. Unlike controlled laboratory evaluations, the OFFSET FXs include uncontrollable variances, such as extreme weather conditions impacting hardware functionality that causes autonomous UAVs and UGVs to perceive the environment and conduct their tactics differently across shifts. 

\subsection{SCs}
The presented results are for two SCs, both of whom are core CCAST team members and system developers. The SCs are 31-40 years old, have at least a Bachelor's degree, and are highly proficient computer users, using such devices eight or more hours a week. The SCs play video games on average 3-8 hours a week and consider themselves proficient players. Finally, both SCs spend on average 3-8 hours a week using I3, with the virtual reality equipment, and consider themselves to be very to highly proficient system users.   

The nature of the DARPA OFFSET field exercises, including the swarm's size, the developmental nature of the technology as well as the associated costs and safety concerns, implies that a team member acts as the SC. Both SCs became project team members in October 2017, when the program began. Swarm Commander 1 ($SC_1$) completed shifts at all field exercises, while Swarm Commander 2 ($SC_2$) only attended FX-3 and FX-6. During FX-3 and FX-6, the SCs traded off shifts, generally serving as SC for as close to an equivalent number of shifts as possible. $SC_1$ was the sole SC at all other field exercises. 


\subsection{Field Exercises}

FX-6 was conducted Nov 3-19, 2021 at Fort Campbell.  
The FXs always include shifts for integrating the CCAST system with the government systems and dry runs. Exercise shifts, during which the CCAST team attempted to achieve the mission, account for the remaining shifts. It is noted that some early exercise shifts are effectively dry runs, as system modifications are the focus. 
Human subjects data collection commenced once the team transitioned to addressing the mission objectives. Even after this transition, some shifts encountered unavoidable technical difficulties (e.g.,\ LTE communication failures). The CCAST team completed twenty shifts during FX-6, and human factors data collection occurred during twelve shifts.


\subsubsection{FX Operational Conditions}

The FXs are physically and mentally draining, with sifts occurring seven days a week, with an 
average of 13.5 hours at the CACTF daily,
often with additional work conducted in the evening. At FX-6, the team worked 
in a large tent, without climate control, potable running water, etc. Teams must supply their meals and beverages, as external sources are not easily accessible. 

The shift preparation required distributing and preparing all hardware vehicles in the launch area and setting up the command center (C2) systems. The off-shift SC often contributed to the hardware vehicle distribution and set up, while the on-shift SC set up I3 in the C2. The CCAST system dispatcher was set up in another C2 area, sufficiently distant from the SC to prohibit direct communication. During shift preparation, the SC verified communications between I3 and the dispatcher system. The CCAST team member setting up the dispatcher system verified communications between it and the LTE basestation. During a shift, dedicated CCAST team members were responsible for acting as in field safety spotters, managing the vehicle hardware (e.g.,\ swapping UAV batteries), etc. Communication between the distributed human team members occurred via walkie-talkie. The human subjects experimenter was responsible for relaying communications that required SC response, or originated with the SC (e.g.,\ ``launching UAVs''). 

\begin{figure} [!htb]
    \centering
    \includegraphics[height=2.0in]{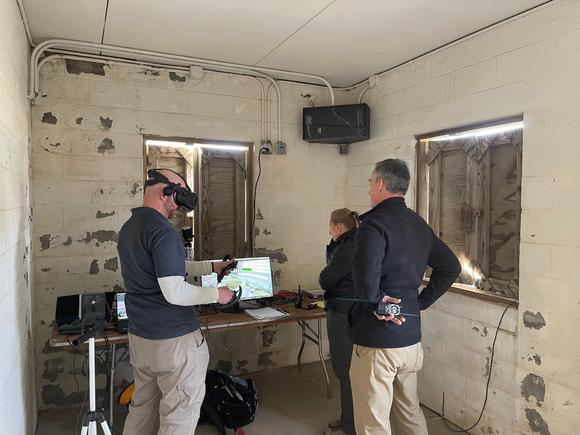}
    \caption{The FX-6 I3 C2 demonstration environment. Note, the larger display, right side, supports explanations to external observers, and is not used to control the swarm. Photo courtesy of DARPA.}
    \label{fig:C2}
\end{figure}

The FX-6 C2 was in a cinder block building (see Figure \ref{fig:C2}). 
The SC's I3 station was set up in a single room on the second floor that minimized light pollution. The I3 virtual reality headset and the C2's room location
resulted in all swarm operations being beyond the SC's visual line of sight; however, the SC was able to hear to the UAVs take off, depart and RTL, but is unable to hear the UGVs. 

Prior to mission start, a mission brief provided the mission's objectives. Upon shift completion, 
a shift debrief was conducted, usually followed by a brief break. After the break, the SCs frequently completed system development tasks, or provided demonstrations for visitors. 

During FX-3, it was determined that the virtual reality hand controllers' functionality was impacted negatively by cold temperatures. The FX-6 cinder block C2 room was frequently many degrees colder than the outside ambient temperature. The hand controllers were placed inside the SC's clothing or hand warmers were used to maintain the controllers' responsiveness. During the shift, the hand controllers did not exhibit issues. A similar concern arose for the virtual reality chest tracker, which was also kept under the SC's clothing until the shift began; thus, avoiding temperature induced issues. 
Appendix \ref{Appx:Conditions}'s Table \ref{Tab:FX6Info} provides detailed weather conditions for each CACTF data collection day.

\subsection{FX Variances} 
\begin{figure}[htb]
\centering
  \includegraphics[width=0.5\linewidth]{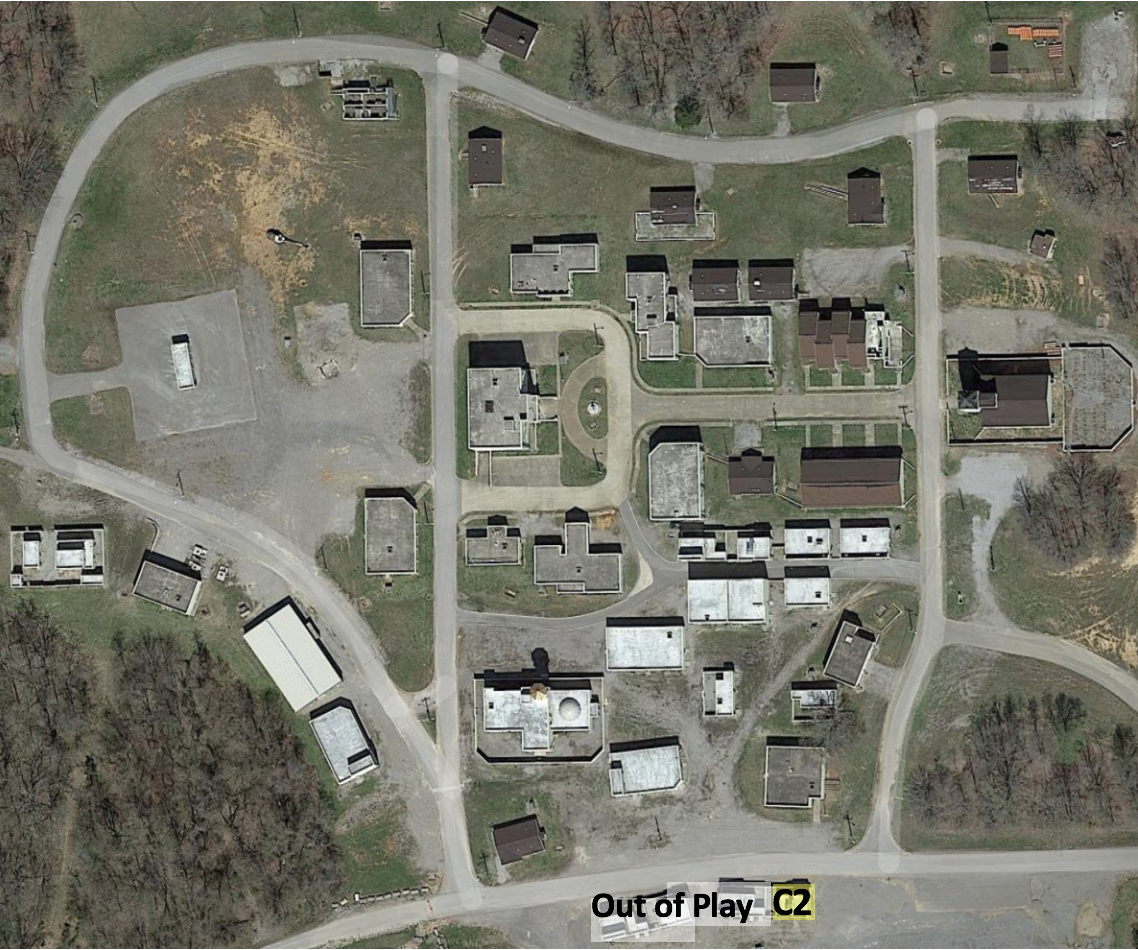}

\caption{The Fort Campbell Cassidy CACTF, the site of FX-6. The yellow area is the C2 building.}
 \label{fig:CACTFs}
\end{figure}

Field exercises provide ecologically valid opportunities to assess SC performance with actual hardware systems in representative environments while conducting representative missions; however, they also create numerous challenges. 
Each OFFSET field exercise increasingly scaled the mission and swarm complexity. Both the CCAST's swarm's number of vehicles and heterogeneity increased with each FX. Table \ref{Tab:Robots} indicates which hardware vehicles were used at each FX. The FX-3 swarm shifts included up to 55 UAVs and 30 UGVs, FX-4's swarms had up to 50 UAVs and 60 UGVs, while the FX-6 swarms incorporated up to 139 UAVs, 44 UGVs and up to 100 virtual vehicles.  Further, each field exercise was conducted at a different CACTF, where each CACTF's built environment varied substantially. 
The FX-6's Fort Campbell Cassidy CACTF is more compact than FX-4's Joint Base Lewis-McChord's Leschi Town, but presented a denser urban environment, see Figure \ref{fig:CACTFs}. The FX-6 C2 building 
is identified on the figure and FX-6's launch area 
was located on the road in front and in the parking lot to the east of C2. 

Each field exercise increased the mission complexity by significantly increasing the number and types of artifacts to be detected and responded to appropriately. The number of artifacts that neutralized vehicles increased with each field exercise, as did the complexity of responses that vehicles were to perform upon detecting an active artifact. As such, the mission objectives and the associated mission plans varied across the field exercises. Mission plans also varied across shifts within a FX as new information became available, artifacts were modified by the DARPA team, etc. FX-6's increased mission complexity resulted in a higher neutralization of vehicles before they were able to venture very far into the CACTF. 

FX-6 shift durations 
varied from 1 
to 3.5 hours. Longer shifts occurred later in the field exercise. The UAVs tend to have short battery lives (i.e.,\ 10-20 minutes), which makes
it is necessary 
that the UAVs autonomously RTL for battery swaps. Longer shifts often also result in more neutralized vehicles that need to go to the medic (UGVs) or RTL (UAVs). The swarm's mission progression varied significantly during the longer shifts, which may include additional mission plan phrases, or the number and type of SC specified tactics. Each of these factors can dramatically change the SC's actions across all shifts. 

DARPA's invited distinguished visitor day was Nov $16^{th}$. 
The visitors congregated in designated safe observation areas. CCAST's mission objectives for this day were to (a) place every operational UGV and UAV in the launch zone, and (b) deploy all of those vehicles immediately upon shift start and maintain a high vehicle activity deployment tempo for the entire observation period, the first thirty minutes of the shift. Additional relevant facts are provided in a more detailed analysis of this shift in the results section.


An FX-6 ``surprise'', announced on Nov $15^{th}$, was the notion of 
both integrator teams' swarms\footnote{The second integrator team was lead by Northrup Grumman.} performing the mission objectives during 
\textit{Joint Shifts}. During these shifts, both DARPA OFFSET integrator teams deployed vehicles simultaneously. The CACTF was spatially divided, such that the CCAST team conducted their mission activities on the half of the CACTF closest to C2. Both Nov $18^{th}$ shifts were conducted in a similar manner; however, during the 1330-1500 shift, the CCAST SCs jointly deployed the swarm. 

\subsection{Data Collection}

\subsubsection{Physical Data Collection Configuration}
\begin{figure}[htb]
\centering
\begin{subfigure}{.49\textwidth}
  \centering
  \includegraphics[width=\linewidth]{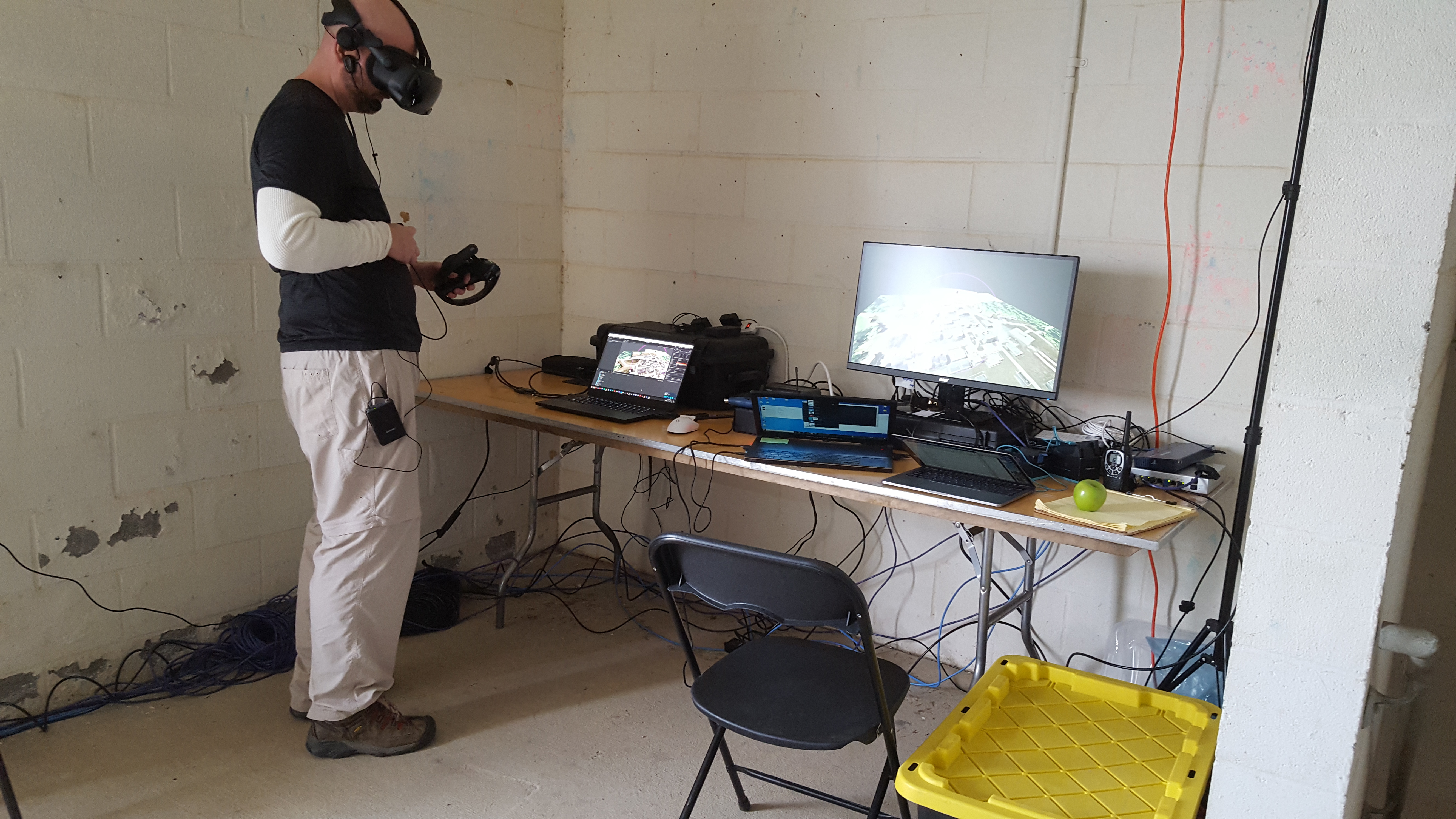}
  \caption{I3 SC area (right) and the human subjects data collection area (left).}
  \label{fig:I3-HS}
\end{subfigure}%
\hfill
\begin{subfigure}{.49\textwidth}
  \centering
  \includegraphics[width=\linewidth]{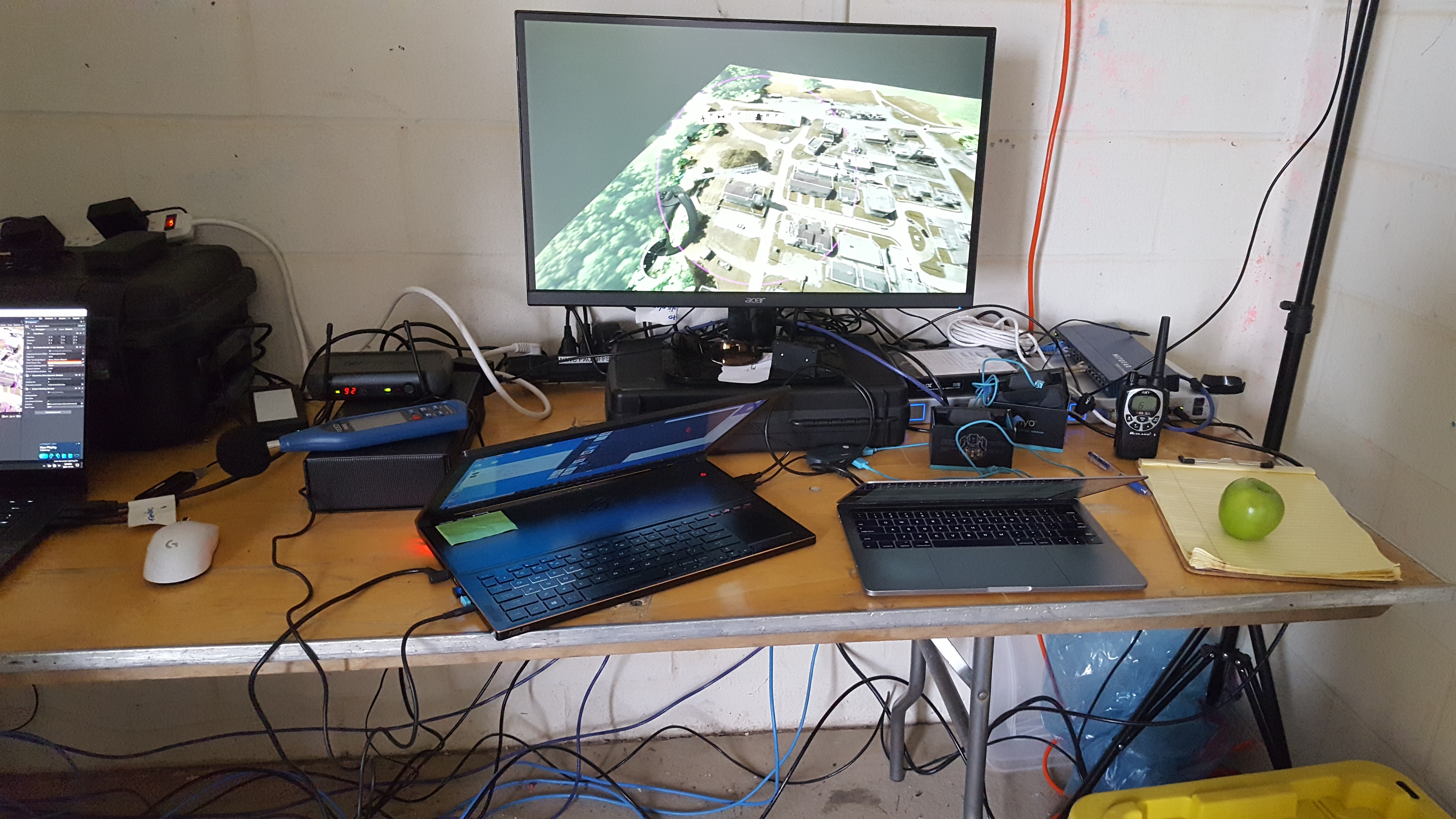}
    \caption{The table top human subjects data collection equipment.}
  \label{fig:HFTableSetup}
\end{subfigure}%

\caption{The I3 SC operational and the human subjects data collection area.}
 \label{fig:I3Area}
\end{figure}

The I3 SC and the human subjects evaluator shared a table in C2, as shown in Figure \ref{fig:I3-HS}. The I3 SC requires the virtual reality equipment with associated charging cables, and the laptop that runs the I3 software. The evaluator's equipment is positioned to the right on the same table. 

The evaluator's monitor, shown in Figure \ref{fig:I3Area}, is directly connected to the I3 laptop and displays the virtual environment and I3 interaction components in real-time. The evaluator's viewable area is larger than the SC's in the virtual reality headset; thus, an indicator assists the evaluator in understanding what the SC can currently view. The evaluator's tools include the laptop on which the data collection software runs, a second laptop for recording notes, events and in situ responses, as well as all necessary sensors and their associated components, shown in Figure \ref{fig:HFTableSetup}.  

\subsubsection{Objective Data Collection Sensors}
\label{sensors}

The multi-dimensional workload algorithm can estimate the cognitive, speech, auditory, visual and physical workload components, which are used to estimate overall workload. The visual workload estimate 
requires an eye tracker. The Valve Index headset does not incorporate an eye tracker, and the evaluator's eye tracker  cannot be worn with the headset. Thus, visual workload was not objectively measured, but was estimated using existing an IMPRINT Pro model.

The multi-dimensional workload algorithm incorporates the physiological-based metrics: heart-rate, heart rate variability (HRV), respiration rate, posture magnitude, speech rate, voice pitch, intensity, and activity, as well as  
noise level (decibels:\ DB). These metrics are used to estimate the component workload levels that are combined into the overall workload estimate.  Cognitive workload is estimated using heart rate, heart rate variability, and noise level variability. Physical workload relies on heart rate, respiration rate, and posture magnitude. 
The auditory workload is estimated using noise level variability, while speech workload is estimated using voice intensity, pitch and activity, as well as speech rate. 

\begin{table}[htb]
\centering
\caption{The correspondence between the sensors and the multi-dimensional workload components.}
\label{tab:WLMetrics}
\begin{tabular}{|l|l|l|l|l|l|}
\hline
\multicolumn{1}{|c|}{\textbf{Sensor}} & \multicolumn{1}{c|}{\textbf{Metric}} & \multicolumn{1}{c|}{\textbf{Cognitive}}& \multicolumn{1}{c|}{\textbf{Auditory}} & \multicolumn{1}{c|}{\textbf{Speech}} & \multicolumn{1}{c|}{\textbf{Physical}}   \\ \hlineB{3}

\multirow{4}{*}{BioHarness} & Heart rate & \cellcolor[HTML]{656565} &  &  & \cellcolor[HTML]{656565} \\ \cline{2-6} 
 & HRV & \cellcolor[HTML]{656565}  &  &  &  \\ \cline{2-6} 
 & Respiration rate &  &  &  & \cellcolor[HTML]{656565}  \\ \cline{2-6} 
 & Postural magnitude &  &  &  & \cellcolor[HTML]{656565}  \\ \hlineB{3}

\multirow{3}{*}{Microphone}  & Speech rate &  &  & \cellcolor[HTML]{656565} &   \\ \cline{2-6} 
 & Voice intensity &  &  &  \cellcolor[HTML]{656565} &  \\ \cline{2-6} 
 & Voice activity &  &  &  \cellcolor[HTML]{656565} &  \\ \cline{2-6} 
 & Pitch &  &  &  \cellcolor[HTML]{656565} &  \\ \hlineB{3}
Reed decibel meter & Noise level & \cellcolor[HTML]{656565} & \cellcolor[HTML]{656565} &  &   \\ \hlineB{3}
\end{tabular}%
\end{table}

The heart rate, heart rate variability, respiration rate and posture magnitude are measured using a BioPac Bioharness\texttrademark~ sensor attached to a chest strap. 
A Reed R8080 decibel meter provides the noise level data. 
The 44100 Khz dual-channel audio signal captured by a Shure PGX1 microphone is transformed into a mono-channel signal prior to calculating the speech rate, as well as voice intensity, activity and pitch metrics. Table \ref{tab:WLMetrics} correlates the sensors to the respective workload component. 

The SC wears the virtual reality headset and chest tracker, as well as the Bioharness BioPac chest strap and sensor, and the Shure microphone headset with the transmitter attached to the SC's pocket, both of which are visible in Figure \ref{fig:I3-HS}.
The Bioharness chest strap is worn underneath the SC's clothing. The noise meter is positioned on the table, left side of Figure \ref{fig:HFTableSetup}.

All of the sensors are designed for indoor, controlled environments and are not hardened for use in extreme conditions. FX-6 was the first time the sensors were used outside of controlled laboratory conditions. Generally, the sensors performed as expected. The Bioharness 
transmits measurements in real-time to the data collection laptop via Bluetooth. 

Prior to shift start, the SC donned the sensors. The Bioharness sensor must be placed on the side of the upper torso. If the sensor is improperly placed, the heart rate readings are very low (e.g.,\ $\leq 50$). A correct heart rate reading is around 80, but varies by individual. The experimenter conducted data collection to verify that the Bioharness 
data was accurate. The experimenter learned the expected heart rate values for each SC. 
After the SC donned and positioned the microphone, the experimenter asked the SC to speak a sentence recorded using the Audacity software. If the speech was not adequately captured, the microphone was adjusted and the test repeated. 


The speech data was not collected on Nov $11^{th}$ due to a missing component, or for November $12^{th}$ due to experimenter error. It is unclear why only two minutes of speech data was recorded during the Nov $18^{th}$ 1330-1530 shift. The noise meter malfunctioned on each data collection shift. Often the noise meter functioned properly for a period of time, and then malfunctioned. There were a small number of instances where the sensor data collection was interrupted and then restarted, which are classified as ``No data''.

The sensor streams for cognitive, speech, auditory and physical were processed using the multi-dimensional workload algorithm's neural networks for the supervisory-based adaptive human-robot teaming architecture, see Section~\ref{WorkAlg}. 
The FX's variable nature makes developing OFFSET specific training data sets or corresponding IMPRINT Pro models difficult. 
Using the reduced set of workload components to estimate overall workload actually underestimates overall workload, as it does not incorporate the missing visual workload and for some shifts, the missing speech workload. 
The IMPRINT Pro models developed for the supervisory-based adaptive human-robot teaming architecture (Section \ref{WorkAlg}) can be leveraged to provide reasonable estimates of overall workload for the missing workload components (i.e.,\ all visual, and some speech). 

The impact of the missing workload components on the overall workload estimate is a percent change relative to the current components' workload value. The resulting overall workload estimate needs to be normalized. The standard normalization equation (i.e.,\ $(value - min)/(max - min)$) can be reduced to Equation 1, where the \textit{min} and \textit{max} values map to the supervisor-based adaptive human-robot teaming architecture's IMPRINT Pro model's values, 0 and 70.4, respectively. This reduction results in a normalization equation, $value/MaxOverallWorkloadVal$, where \textit{MaxOverallWorkloadVal} is the maximum raw value from the IMPRINT Pro model. The \textit{value} component usually is the estimated overall workload from the multi-dimensional workload algorithm; however, some component values (i.e.,\ visual, sometimes speech) are missing. The missing components reduce the maximum overall workload value the algorithm can estimate, due to the uniform aggregation of the workload components. Thus, the algorithm's estimated overall workload value must be adjusted by $value = RawVal + MissingComponentVals$, where \textit{RawVal} is the multi-dimensional workload algorithm's estimated overall workload value without the missing components, and the \textit{MissingComponentsVal} is the respective average values from the missing components. Lastly, the result is multiplied by 100 to ensure the values are in the range 0 – 100, resulting in Equation 1's \textit{ScaledNormalizedVal} producing the estimated overall workload.
\begin{equation}
\label{Eq:1}
    ScaledNormalizedVal = 100*(RawVal + MissingComponentsVals)/ MaxOverallWorkloadVal.
\end{equation}
The lower the estimated workload using the FX data, the larger the impact of incorporating the missing components' contributions. It is important to recognize that the missing components are very unlikely to be at their maximum value, especially if other components are overloaded; thus, estimating the overall workload in this manner provides a more accurate overall workload estimate for the FX-6 results. 

The overall workload estimates were classified into the workload levels (i.e.,\ underload, normal load, overload) using the same thresholds as the prior work, see Section \ref{WorkAlg}. The resulting overall workload estimate was classified as underload if the value was $\leq 25$, overload if the value was $\geq 60$, and normal load otherwise.

\subsubsection{Subjective Data Collection}

In situ probes \cite{Harriott2013} focused on the workload components, stress and fatigue. 
The experimenter asked the SC, approximately every ten minutes to respond to each of the in situ probes with their subjective rating. The meaning of the in situ probe terms were defined for the SCs, who verified their understanding of the terms' meanings prior to data collection. Overtime, the experimenter simply stated, for example ``[SC name]: cognitive'', in order to minimize the disruption to the SC's current tasks. 


The SCs rated their perceived workload components' (i.e.,\ cognitive, auditory, speech, visual and physical), stress and fatigue levels on a scale from 1 (very low) to 7 (very high). All in situ probe responses 
were normalized to a value between 1 and 100. The SCs provided subjective workload component weightings post-FX. The SCs were instructed to weight each component relative to how much they felt each component impacted their overall workload. The total of the components weights was required to equal 100. 
Each individual normalized workload component in situ response was averaged. The respective component weighting, using the subjective weightings, was applied to create a weighted mean for each component. The weighted component means were summed to generate the overall subjective workload values. 

\subsubsection{Dependent Variables}
\label{DepVars}

The objective dependent variables are the workload component estimates as well as the overall workload estimate. 
Speech workload metrics were not collected on 
Nov $11^{th}$ (all shifts), Nov $12^{th}$, and Nov $18^{th}$ 1330-1530.  Table \ref{Tab:FX6Channels} indicates, by shift, which workload component estimates were determined using the collected objective metrics (green), and which used the IMPRINT Pro-based model estimates (orange).

Due to an out of the box default programming parameter, the noise meter stopped recording data during the data collection. Resetting the noise meter during a shift and debugging the issue did not resolve the issue. 35,562 good readings were recorded across five shifts during the FX (i.e.,\ Nov 13, 14, 16, 17 1200-1400, 18 1000-1130) or 21.9\% of those shifts' total data points. The weighted minimum raw noise meter reading value across these shifts was 50.78 dB (i.e.,\ moderate rainfall~\cite{DecibelPro2022}) and the weighted maximum of 81.89 dB (i.e.,\ an alarm clock). The weighted mean noise level across these data points was 60.75 dB (weighted standard deviation = 6.00, i.e.,\ normal conversation). The analysis used all good recorded raw noise meter values. The bad readings were replaced with a point sample from a Gaussian distribution with the weighted mean and weighted standard deviations as the $\mu$ and $\sigma$ (i.e.,\ distribution parameters), respectively. The sampled point was clipped to be within the weighted minimum and maximum. This approach is more representative of the actual auditory workload in the FX environment. As such, the light green in Table \ref{Tab:FX6Channels} for the Auditory component represents the use of actual and mean dB values for the shifts with valid recorded values, or the substituted dB values. 


\begin{table}[htb]
\caption{Objectively assessed workload components (Green) vs.\ components estimated using IMPRINT Pro-based model results (Orange), by shift. No data was collected for the two shifts (red).} \label{Tab:FX6Channels}
\begin{center}
\begin{tabular}{|c|c||c|c|c|c|c|}
  \hline

\multicolumn{2}{|c||}{ \textbf{Shift}} & \multirow{2}{*}{\textbf{Cognitive}} &  \multirow{2}{*}{\textbf{Physical}} & \multirow{2}{*}{ \textbf{Speech}} & \multirow{2}{*}{ \textbf{Auditory}} & \multirow{2}{*}{ \textbf{Visual}} \\ \cline{1-2}
 \textbf{Date} & \textbf{Time} & &  & &  & \\
  \hline\hline
\multirow{4}{*}{ 11-Nov } & 1100-1200 & \cellcolor{tearose} & \cellcolor{tearose} & \cellcolor{tearose} & \cellcolor{tearose} & \cellcolor{tearose}\\ \cline{2-7}
 & 1300-1400& \cellcolor{green} & \cellcolor{green} & \cellcolor{orange} & \cellcolor{lightgreen} & \cellcolor{orange}  \\ \cline{3-7}
  & 1500-1600 & \cellcolor{green} & \cellcolor{green} & \cellcolor{orange} & \cellcolor{lightgreen} & \cellcolor{orange}  \\ \cline{3-7}
  & 1630-1730 & \cellcolor{green} & \cellcolor{green} & \cellcolor{orange} & \cellcolor{lightgreen} & \cellcolor{orange}  \\ \cline{3-7}
  & 1800-1900 & \cellcolor{green} & \cellcolor{green} & \cellcolor{orange} & \cellcolor{lightgreen} & \cellcolor{orange}  \\ \hline
12-Nov & 0830-1130 & \cellcolor{green} & \cellcolor{green} & \cellcolor{orange} & \cellcolor{lightgreen} & \cellcolor{orange} \\ \hline
13-Nov & 1430-1630 & \cellcolor{green} & \cellcolor{green} & \cellcolor{green} & \cellcolor{lightgreen} & \cellcolor{orange} \\ \hline
14-Nov & 0800-1130& \cellcolor{green} & \cellcolor{green} & \cellcolor{green} & \cellcolor{lightgreen} & \cellcolor{orange}\\ \hline
15-Nov & 1300-1630 & \cellcolor{tearose} & \cellcolor{tearose} & \cellcolor{tearose} & \cellcolor{tearose} & \cellcolor{tearose}\\ \hline
16-Nov & 1000-1200 & \cellcolor{green} & \cellcolor{green} & \cellcolor{green} & \cellcolor{lightgreen} & \cellcolor{orange}\\ \hline
\multirow{2}{*}{17-Nov} & 1200-1400 & \cellcolor{green} & \cellcolor{green} & \cellcolor{green} & \cellcolor{lightgreen} & \cellcolor{orange}\\ \cline{2-7}
 & 1400-1630 & \cellcolor{green} & \cellcolor{green} & \cellcolor{green} & \cellcolor{lightgreen} & \cellcolor{orange}\\ \hline
\multirow{2}{*}{ 18-Nov} & 1000-1130 & \cellcolor{green} & \cellcolor{green} & \cellcolor{green} & \cellcolor{lightgreen} & \cellcolor{orange}\\ \cline{2-7}
 & 1330-1530 & \cellcolor{green} & \cellcolor{green} & \cellcolor{orange} & \cellcolor{lightgreen} & \cellcolor{orange}  \\ \hline 
\end{tabular}
\end{center}
\end{table}

The respective missing workload component estimates (i.e.,\ orange in the table) estimated overall workload using the IMPRINT Pro model's average, or mid-point, values. The estimated missing components are combined with the workload estimates to obtain the overall workload estimation. This estimation approach is justified given that it is highly unlikely that all workload components will be overload simultaneously, which was confirmed via observation of the OFFSET SCs. Specifically, the IMPRINT Pro model averages were used for the visual and some speech workload estimates.

\section{FX-6 Human Subjects Evaluation Results}
\label{Sec:Results}

Human subjects data was collected over eight days and twelve shifts, with swarms that differed in the numbers and combinations of hardware and virtual vehicles. 
The SCs generally selected amongst themselves who served as a shift's commander; however, the experimenter did discuss with them balancing the number of shifts and hours serving as the commander.

Rain caused the Nov $11^{th}$'s five shifts to occur in the hotel conference room using only virtual vehicles. 
No objective data was collected for the 1100-1200 shift. 
Six dedicated CCAST shifts at the CACTF occurred Nov $12^{th}$ through the Nov $17^{th}$ 1200-1400 shift. The remaining three CACTF shifts were  Joint Shifts. During the final joint shift, both CCAST SCs jointly deployed the swarm.  The primary data collection days between Nov $14^{th}$ and the Nov $17^{th}$ 1200-1400 shift had a range of 81 (Nov $17^{th}$ 1200-1400: 10 UGVs and 71 UAVs) to 93 hardware platforms (Nov $14^{th}$: 8 UGVs and 78 UAVs). During these same dates, the number of virtual vehicles ranged from 30 (three shifts: 10 UGVs, 20 UAVs) to 125 virtual vehicles (Nov $17^{th}$ 1200-1400: 20 UGVs and 105 UAVs). The largest number of vehicles were used for the Nov $18^{th}$ 1330-1530 joint shift (Hardware: 30 UGVs, 110 UAVs; Virtual: 10 UGVs, 50 UAVs). Additional swarm vehicle composition details are provided in Appendix \ref{Appx:Conditions} Table \ref{Tab:FX6Vehicles}.


Overall, $SC_{1}$ completed eight shifts, totaling 15 hours, and $SC_{2}$ had seven shifts totaling 12.5 hours. An I3 hardware failure caused $SC_{1}$ to assume the SC role a few minutes into the Nov $15^{th}$ shift. Due to these unexpected changes, no data was collected. 
No data was recorded during the 1100-1200 Nov $11^{th}$ shift, or the last Nov $18^{th}$ joint SC shift. As a result, 12.5 hours of data was collected for $SC_{1}$ and 10 hours for $SC_{2}$.  

\subsection{Subjective Results}
The in situ probes provide insight into the SC's state during a shift, as compared to post-shift (e.g.,\ post-trail) tools, such as NASA Task Load Index \cite{Hart1988}. However, several known issues are associated with subjective metrics, including workload \cite{MatthewsetalTIE2020}. The normalized subjective in situ overall workload results are presented in Table \ref{Tab:WLSubj}. 
The gray rows represent $SC_{2}$'s shifts.

\begin{table}[htb]
\caption{The subjective in situ normalized overall workload, stress and fatigue descriptive statistics, mean (SD), by shift and SC. 
Gray cells represent $SC_{2}$'s results.} \label{Tab:WLSubj}
\begin{center}

\begin{tabular}{|c|c||l|l|l|}
\hline
\multicolumn{2}{|c||}{ \textbf{Shift}} & \textbf{Overall Subjective} & \multirow{2}{*}{ \textbf{Stress}} & \multirow{2}{*}{ \textbf{Fatigue}}\\ \cline{1-2}
 \textbf{Date} & \textbf{Time} &  \textbf{Workload} & & \\ 
  \hline\hline
\multirow{4}{*}{ 11-Nov }&  1300 & 28.14 (5.27) & 47.2 (13.81) & 34 (0) \\ \cline{2-5}
& 1500 & 22.06 (16.15) & 18 (0) & 18 (0) \\ \cline{2-5} 
& \cellcolor{lightgray}1630  & \cellcolor{lightgray}28.1 (10.43) & \cellcolor{lightgray}14.6 (7.6) & \cellcolor{lightgray}24.4 (8.76)\\ \cline{2-5} 
& 1800   & 31.88 (6.71) & 31.13 (15.54) & 50.5 (0)\\ \hline 
12-Nov & \cellcolor{lightgray}0830  & \cellcolor{lightgray}36.57 (16.92) & \cellcolor{lightgray}26.88 (15.3) & \cellcolor{lightgray}51.92 (19.13) \\ \hline 
13-Nov & 1430  & 35.21 (9.44) & 42.3 (11.55) & 42.25 (8.7) \\ \hline 
14-Nov & 0800 & 37.24 (9.56) & 32.39 (15.68) & 52.33 (9.62) \\ \hline 
 16-Nov & \cellcolor{lightgray}1000 & \cellcolor{lightgray}43.93 (7.53)  & \cellcolor{lightgray}63.46 (18.51)  & \cellcolor{lightgray}44.71 (17.77) \\ \hline 
\multirow{2}{*}{17-Nov} & \cellcolor{lightgray}1200  & \cellcolor{lightgray}32.66 (8.81) & \cellcolor{lightgray}26.73 (8.36) & \cellcolor{lightgray}43.09 (15.25) \\ \cline{2-5} 
 & 1400  & 27.38 (15.77) & 22.36 (18.32) & 31.71 (6.05) \\ \hline 
\multirow{2}{*}{ 18-Nov} & \cellcolor{lightgray}1000 &  \cellcolor{lightgray}28.65 (10.08) &  \cellcolor{lightgray}18 (0) & \cellcolor{lightgray}38.95 (7.97) \\ \cline{2-5} 
 & 1330 &  37.51 (11.02) &  41.73 (22.83) & 35.35 (15.65) \\ \hline 
\end{tabular}
\end{center}
\end{table}

The SCs' subjective weightings differed across the workload components. $SC_{1}$'s weights were, from highest to lowest: Visual: 35\%, Cognitive: 25\%, Speech: 20\%, Auditory: 15\%, Physical: 5\%, while $SC_{2}$'s responses were: Cognitive: 40\%, Visual: 20\%, Speech and Physical: 15\%, Auditory: 10\%. 
The mean overall subjective workload across all shifts calculated using the subjective weightings was 33 (Standard Deviation, SD = 5.83). Nov $16^{th}$ resulted in the highest perceived overall workload, as shown in Table \ref{Tab:WLSubj}. 

The normalized in situ component responses are provided in Appendix~\ref{Appx:SubResults}'s Table~\ref{Tab:WLCompSubj}. The highest subjective component CACTF shift responses for the Cognitive, Speech and Auditory components occurred on Nov $16^{th}$, the distinguished visitors day.
The Visual workload responses that day were effectively tied for the highest ratings on Nov $13^{th}$, which can also be said for the Physical component on Nov $16^{th}$ and Nov $12^{th}$. 

The in situ subjective stress and fatigue values were normalized, see the descriptive statistics in Table \ref{Tab:WLSubj}. The overall mean stress level across all shifts was 28.94 (SD = 10.8). Stress varied substantially across shifts, with the highest level reported on Nov $16^{th}$. 
This high stress level led to $SC_{2}$'s mean CACTF shifts stress being recorded as 44.67 (SD = 5.41). $SC_{1}$'s reported CACTF shifts stress level was 40.41 (SD = 9.07). 


Generally, fatigue was higher during the CACTF shifts, the exception was the last Nov $11^{th}$ virtual shift, shown in Table \ref{Tab:WLSubj}. The virtual shifts were short (1 hour), with short breaks (30 minutes) between shifts and additional shifts added late in the day. 
The mean subjective fatigue level across all shifts was 32.81 (SD = 14.32).  $SC_{2}$ reported higher mean fatigue, 36.27 (SD = 19.7), than $SC_{1}$, 40.41 (SD = 9.1). 

\subsection{Estimated Workload Results}
The overall workload estimates were classified as normal workload if ($25 < X < 60$), where $X$ represents the overall workload estimate. The estimates were classified as underload, if $X \leq 25$ and overload if $X \geq 60$, with these thresholds set as described in Section \ref{WorkAlg}. 

\subsubsection{Estimated Overall Workload Descriptive Statistics}
\label{Sec:DescStats}

The mean, standard deviation (SD), minimum (min), and maximum (min) overall workload estimates for each shift are presented in Table \ref{Tab:WLMean}. A total of 12,181 usable data points were recorded for all shifts, see Table \ref{Tab:WLState}. 
The mean estimated overall workload weighted by the number of estimates per shift
was 46.58 (SD = 6.4). The CACTF shifts' estimated weighted average overall workload was slightly lower, 46.27 (SD = 6.44). The weighted means for 
Nov $17^{th}$ and $18^{th}$ dropped marginally to 46.18 (SD = 6.24). The difference between the SCs' overall weighted means was 5.23. 
$SC_{2}$ had the higher overall weighted mean workload estimate, 49.56 (SD = 6.23), as compared to $SC_{1}$'s, 44.23 (SD = 6.53). A larger difference, 6.61, existed when comparing the SCs using only the CACTF shifts' results. $SC_{1}$'s weighted mean estimates over four CACTF shifts was 42.98 (SD = 6.54), but was 49.60 (SD = 6.34) across $SC_{2}$'s four CACTF shifts. The minimum estimated overall workload across all shifts was 28.71, a normal load classification. The maximum estimated  overall workload was classified as overload for eight, or two-thirds, of all shifts. 

\begin{table}[htb]
\caption{The overall workload estimates descriptive statistics by shift and SC.
} 
\label{Tab:WLMean}
\begin{center}
\begin{tabular}{|c|c||l|l|}
  \hline
\multicolumn{2}{|c||}{ \textbf{Shift}} & \multicolumn{2}{|c|}{\textbf{Overall Workload}} \\\cline{1-4}
 \textbf{Date} & \textbf{Time} & \textbf{Mean (SD)} & \textbf{Min-Max} \\
  \hline\hline
\multirow{4}{*}{ 11-Nov }& 1300 & 
45.25 (6.76) & 30.45-61.56 \\ 
\cline{2-4}
& 1500 & 
51.88 (6.68) & 32.34-67.19 
\\ \cline{2-4} 
& \cellcolor{lightgray}1630 &  
\cellcolor{lightgray}47.66 (4.90) & \cellcolor{lightgray}34.62-58.55  
\\ \cline{2-4} 
& 1800  & 
44.22 (5.28) & 32.60-54.08

\\ \hline
12-Nov & \cellcolor{lightgray}0830 & 
\cellcolor{lightgray}48.68 (6.33) & \cellcolor{lightgray}32.91-64.61 
\\ \hline
13-Nov & 1430 & 
44.40 (6.52) & 31.01-64.19 
\\ \hline
14-Nov & 0800 &
43.07 (6.96) & 28.71-63.61 
\\ \hline
 16-Nov & \cellcolor{lightgray}1000 & 
 \cellcolor{lightgray}50.48 (6.25) & \cellcolor{lightgray}32.11-70.23 
 \\ \hline
\multirow{2}{*}{17-Nov} & \cellcolor{lightgray}1200 & 
\cellcolor{lightgray}50.25 (6.52) & \cellcolor{lightgray}32.76-69.59 
\\ \cline{2-4} 
 & 1400 & 
 41.73 (4.90) & 29.87-54.63 
 \\ \hline
\multirow{2}{*}{ 18-Nov} & \cellcolor{lightgray}1000 & 
\cellcolor{lightgray}48.42 (6.27) & \cellcolor{lightgray}34.06-63.57 
\\ \cline{2-4} 
 & 1330 & 
 42.20 (6.88) & 30.76-58.36

 \\ \hline 
\end{tabular}
\end{center}
\end{table}

The joint shifts on Nov $17^{th}$ 1400-1630 and Nov $18^{th}$ resulted in $SC_{1}$'s lowest FX overall workload estimates over all shifts. 
$SC_{2}$'s estimated overall workload during the joint shift on Nov $18^{th}$ was lower than the two prior shifts, but was not this SC's lowest. A review of the CACTF shifts in Table \ref{Tab:WLMean} reveals that two days stand out, as far as the highest estimated workloads. Both occurred for $SC_{2}$ on Nov $16^{th}$, and the Nov $17^{th}$ 1200-1400 shift. These two shifts will be the focus of additional analysis in Section \ref{Sec:Graphs}.  

The descriptive statistics provide a high level perspective of the overall workload estimates, but also obscure more important and impactful results. These descriptive statistics results do not communicate the extent to which the SCs experienced overload or underload states, or the sensitivity of the multi-dimensional workload algorithm to changes in overall workload.  

\subsubsection{Estimated Workload State Frequencies}
\label{Sec:Freq}

 12,242 estimates were generated across all data collection shifts; however, 61 estimates were invalid, resulting in 12,181 usable estimates. Each usable overall workload estimate 
was classified using the defined thresholds, as normal load, overload, or underload. No underload instances existed. The frequency counts by shift, classification, and SC 
are summarized in Table \ref{Tab:WLState}. $SC_{1}$ completed seven shifts with 6,712 ($55.1\%$) usable overall workload estimates. $SC_{2}$'s five shifts resulted in 5,469 ($44.9\%$) usable estimates. The 61 ``No Data'' estimates represent instances where the data recording software failed, but was restarted during the shift. 

\begin{table}[htb]
\caption{The overall workload estimate instances' state classifications 
by shift, and SC. 
} \label{Tab:WLState}
\begin{center}
\begin{tabular}{|c|c||c|c|c||c|}
  \hline
\multicolumn{2}{|c||}{ \textbf{Shift}} & 
\multicolumn{2}{|c|}{\textbf{Overall Workload}} 
&\multirow{2}{*}{\textbf{No Data}} & \multirow{2}{*}{\textbf{Total}}\\ \cline{1-4} 
 \textbf{Date} & \textbf{Time} & 
 \textbf{Normal Load} & \textbf{Overload} 
 & &\\
  \hline\hline
\multirow{4}{*}{ 11-Nov }& 1300-1400 & 
 569 & 5 
& 0 & 574 \\ \cline{2-6} 
  & 1500-1600 & 
 689 & 72 
  & 0 & 761 \\ \cline{2-6} 
  & \cellcolor{lightgray}1630-1730 & 
 \cellcolor{lightgray}401 & \cellcolor{lightgray}0 
  &\cellcolor{lightgray}0 & \cellcolor{lightgray}401  \\ \cline{2-6} 
  & 1800-1900 & 
234
& 0 
  & 0 & 234  \\ \hline
12-Nov & \cellcolor{lightgray}0830-1130 & 
 \cellcolor{lightgray}1,041 & \cellcolor{lightgray}35
& \cellcolor{lightgray}48 & \cellcolor{lightgray}1,124\\ \hline
13-Nov & 1430-1630 & 
 1,098 & 25 
&0 & 1,123 \\ \hline
14-Nov & 0800-1130 & 
2,205 & 12 & 
13 & 2,230
\\ \hline
16-Nov & \cellcolor{lightgray}1000-1200 & 
\cellcolor{lightgray}1,521 & \cellcolor{lightgray}127 
& \cellcolor{lightgray}0 & \cellcolor{lightgray}1,648 \\ \hline
\multirow{2}{*}{17-Nov} & \cellcolor{lightgray}1200-1400 & 
 \cellcolor{lightgray}1,174 & \cellcolor{lightgray}76 
&\cellcolor{lightgray}0 & \cellcolor{lightgray}1,250 \\ \cline{2-6} 
 & 1400-1630 & 
757 &0 
 & 0 & 757 \\ \hline
\multirow{2}{*}{ 18-Nov} & \cellcolor{lightgray}1000-1130 & 
\cellcolor{lightgray}1,069 & \cellcolor{lightgray}25 
& \cellcolor{lightgray}0 & \cellcolor{lightgray}1,094 \\ \cline{2-6} 
 & 1330-1530 & 
 1,046 & 0 
 & 0  & 1,046 \\ \hline \hline

\multicolumn{2}{|c||}{\textbf{Total}} &
11,804 & 377
& 61& 12,242\\ \hline
\end{tabular}
\end{center}
\end{table}


A total of 377  overload instances (3.19\% of all usable estimates) occurred. $SC_{2}$ encountered the majority of overload instances, 263 (2.22\% of usable estimates), across this SC's four CACTF shifts. $SC_{2}$ had no overload instances during the virtual shift. The $SC_{2}$'s highest overload instance frequency, 203 (53.85\% of all overload state instances), occurred on two days. The highest frequency, 127 (33.69\%), occurred on Nov $16^{th}$, while 76 (20.02\%) instances occurred the next day. $SC_{1}$'s highest overload frequency, 72 ($19.10\%$), occurred during a  Nov $11^{th}$ virtual shift, with the second highest frequency, 25 ($6.63\%$), being the Nov $13^{th}$ CACTF shift. $SC_{1}$'s overload frequencies occurred across four shifts, two virtual shifts on Nov $11^{th}$ and two CACTF shifts, with no overload classifications during three of $SC_{1}$'s seven shifts. 
During $SC_{1}$'s two joint shifts, the last shifts on both Nov $17^{th}$ and $18^{th}$, all estimated overall workload instances were classified as normal load; however, $SC_{2}$ experienced 25 overload state instances during the Nov $18^{th}$ 1000-1130 joint shift. 


\subsubsection{Individual Shift Overall Workload Estimate Analyses}
\label{Sec:Graphs}

The estimated overall workload state classification frequencies hint at differences within shifts. Those results also demonstrate 
that the SCs' estimated overall workload generally remains in the normal load range across the shifts, number of hardware and virtual vehicles, and mission plans. However, those results may also lead to the incorrect conclusion that a SC experienced the overload states consecutively during a particular shift. Plotting the individual overall workload estimates across a shift provides a better continuous representation. The plots of the individual overall workload estimate instances, estimated every five seconds per Section \ref{WorkAlg}, were generated and analyzed for each shift, but due to space limitations only three are presented. Additional analysis for each shift relates the in situ subjective results by their recorded times to the associated overall workload estimates (i.e.,\ twelve estimates based on the five seconds between estimates). As well, the number of tactics issued, the number of tasked and active vehicles, and the number of vehicles blocked due to congestion are presented, as each are indicative of the SC's task demands that impact workload. 

\paragraph{Nov $16^{th}$ Shift:}
This shift was the FX-6 distinguished visitor day, which 
is generally the most stressful for the entire CCAST team, including the SC. 
The FX-6 distinguished visitor day occurred 14 days into the FX and was the largest contingent of any FX. The distinguished visitors observed the shift's mission deployment from 1000 to about 1035. The mission plan was designed to deploy all 91 hardware vehicles (10 UGVs and 81 UAVs) immediately upon commencing the mission and to continue maximizing the number of deployed vehicles throughout the observation period. 

\begin{figure} [htb]
    \centering
    \includegraphics[height=3in]{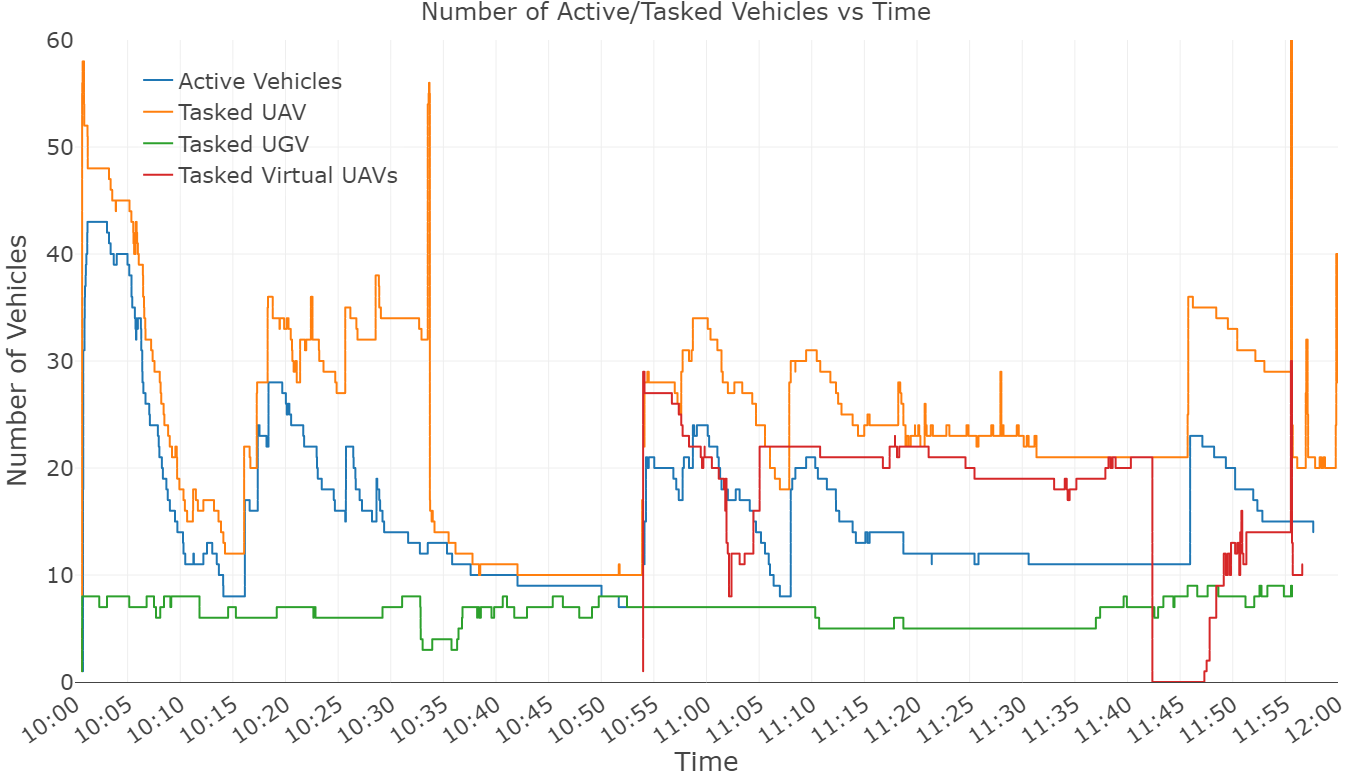}
    \caption{The Nov $16^{th}$ 1000-1200 shift's tasked (by vehicle type) and active vehicles.}
    \label{Fig:16-NovVehicles}
\end{figure}

$SC_{2}$ was the shift's commander. 
The ambient temperature and the mean wind speeds were relatively reasonable, see Appendix~\ref{Appx:Conditions}'s Table \ref{Tab:FX6Info} for details. 
During the first 35 minutes, 74 unique hardware vehicles were deployed, many multiple times. The number of tasked and active vehicles is shown in Figure ~\ref{Fig:16-NovVehicles}. After the observation period completed, ten virtual UGVs and twenty virtual UAVs were added, and the SC deployed 103 unique vehicles.  Note that active agents in the figure only represents active hardware vehicles, as tasked hardware agents may fail to execute their tactics. Therefore, both metrics are plotted for hardware vehicles. The tasked simulated vehicles automatically execute the tactics.

\begin{figure} [htb]
    \centering
    \includegraphics[height=3.5in]{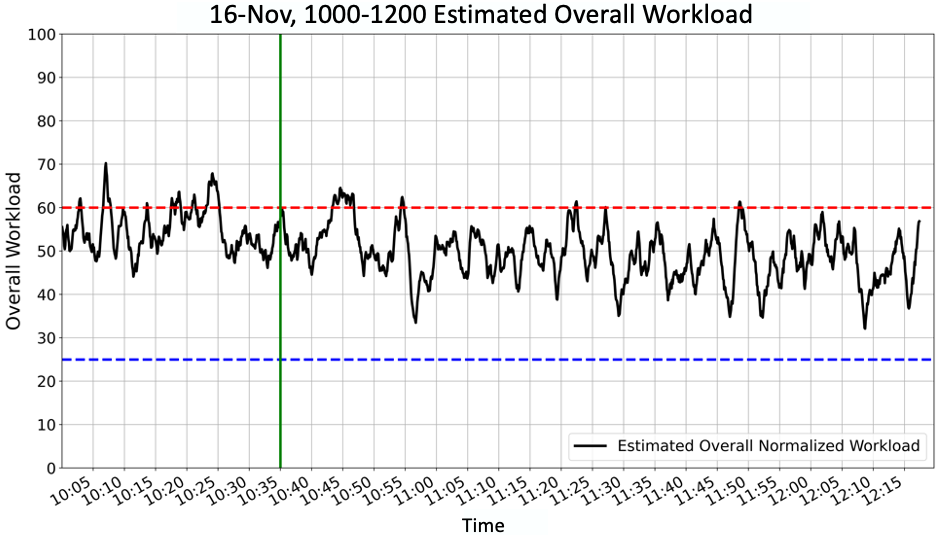}
    \caption{Overall workload estimates for the Nov $16^{th}$ 1000-1200 shift. The majority of distinguished visitors moved on to other activities at 1035 (green line). Note, the blue line represents the underload threshold and the red line represents the overload threshold.}
    \label{Fig:16-Nov}
\end{figure}

$SC_{2}$'s estimated overall workload throughout the shift is plotted in Figure \ref{Fig:16-Nov}. The dashed red line represents the overload threshold (i.e.,\ 60), the dashed blue line represents the underload threshold (i.e.,\ 25), and the green line represents when the majority of the distinguished visitors left the observation area. The estimated workload components were cognitive, speech, auditory, and physical, per Table ~\ref{Tab:FX6Channels}. 

The first takeaway is that the majority, 70, of the SC's overload classifications occurred during the distinguished visitor observation period, between 1000 and 1035, or 55\% of all overload classifications for the entire shift. The highest overload estimate, 70.23, occurred at 1007. The longest sustained overload classification during the observation period was two minutes and ten seconds between 1023 and 1025. 
While the estimated overall workload values oscillated between normal load and overload during the distinguished visitors observation period, the estimates were generally classified as normal load, 95\% of all estimates, after 1035. The longest sustained overload period occurred between 1043 and 1046, lasting 3 minutes and 35 seconds, with a range of 60.36 to 64.53.

\begin{figure}[!htb]
\centering
\begin{subfigure}{.9\textwidth}
  \centering
   \includegraphics[width=.9\linewidth]{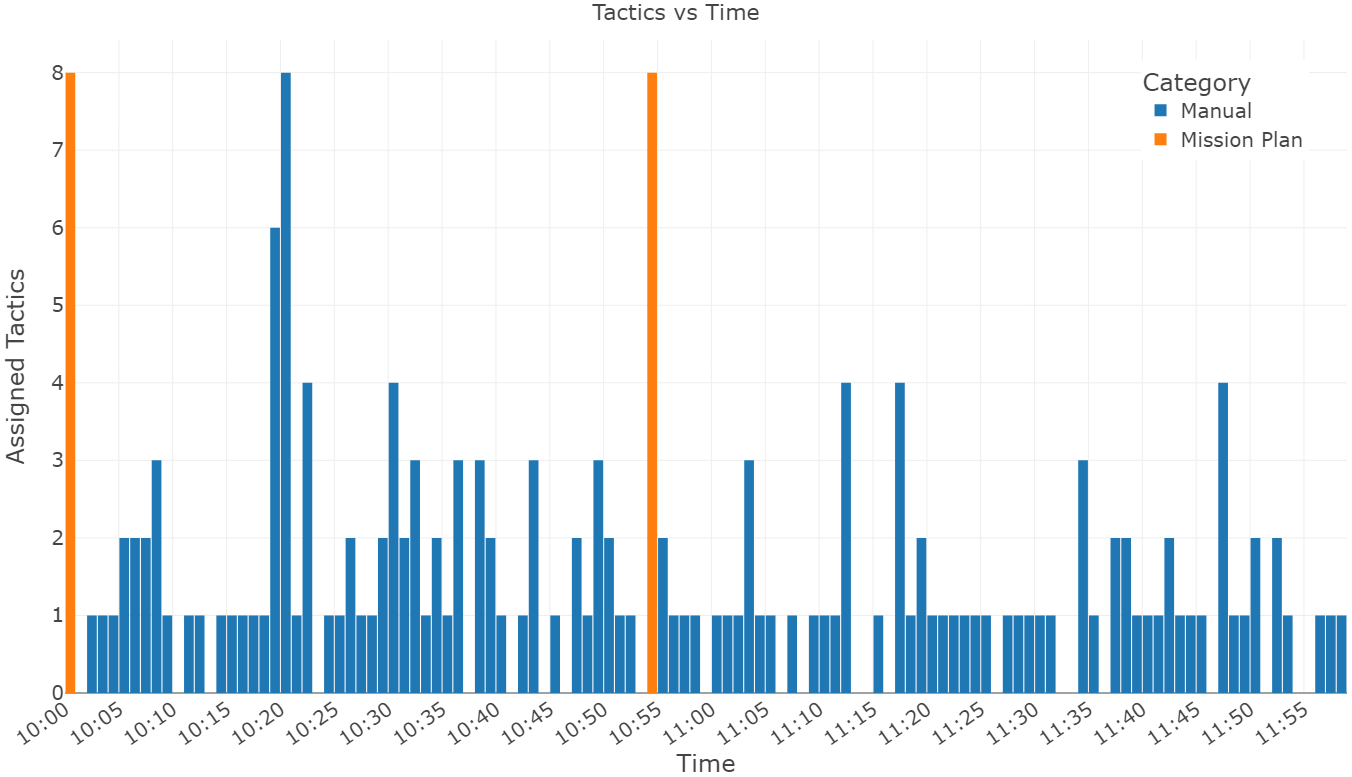}
    \caption{Issued tactics.}
    \label{Fig:16-NovTactics}
\end{subfigure}%
\hfill
\begin{subfigure}{.9\textwidth}
  \centering
   \includegraphics[width=.9\linewidth]{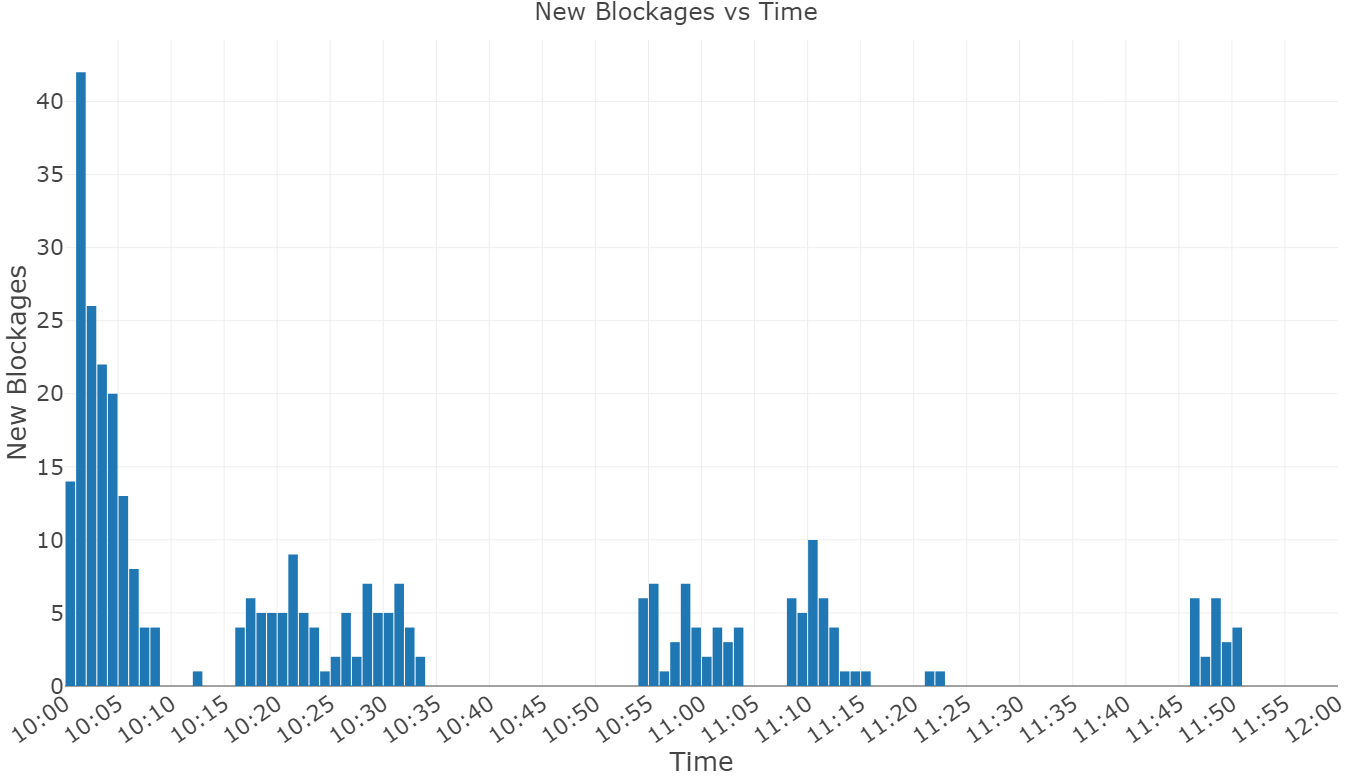}
    \caption{Blockages.}
    \label{Fig:16-NovBlocks}
\end{subfigure}%
\caption{The Nov $16^{th}$ 1000-1200 shift's (a) issued tactics and (b) vehicle blockages by the minute.}
\end{figure}

The mission plan involved deploying all vehicles at the start of the shift (1000) to conduct various tactics around the CACTF. $SC_{2}$ loaded the mission plan at exactly 1000 and launched a volley of vehicles within seconds. The mission plan contained eight tactics, as shown by the orange bar in Figure \ref{Fig:16-NovTactics}, where each tactic tasked multiple vehicles, as shown in Figure~\ref{Fig:16-NovVehicles}. 

The assigned vehicles each plan a navigation path and autonomously begin executing the assigned tactic. 
Typically deploying a large number of vehicles results in a large number of UGVs and UAVs becoming blocked, as shown in Figure~\ref{Fig:16-NovBlocks}. When a block occurs, the impacted vehicle attempts to autonomously resolve the issue. If a block continues for a prolonged period, the SC can issue a Nudge tactic, which causes the vehicle to move a predefined amount before attempting to plan a new navigation path. For example, nudged UAVs will increase in altitude in order to support generating a clear navigation path. The SC may also issue a Stop tactic followed by new tactic with a new goal location, or a RTL tactic. 

The SC uses predefined tactics with modifiable default parameters to specify tactics.
The tactics assign vehicles to complex tactics, such as Surveil a building, or simpler tactics, such as Goto a specific location, Stop, and RTL. The blue tactics shown in Figure ~\ref{Fig:16-NovTactics} between 1001 and 1035 represent SC generated tactics. Tactic generation often leads to higher SC workload, especially if the SC specifies particular vehicles for a tactic, rather than allowing the CCAST system to automatically allocate vehicles to tactics. This increased workload is reflected in Figure \ref{Fig:16-Nov}, where the overall workload estimates increase before tactics are issued. 

The overload instances in the first five minutes are due to tactic generation to resolve the vehicle blockages and deploy more vehicles. Typically the SC waits to allow the vehicles to launch, plan navigation paths and resolve any blockages autonomously. Since this shift was observed by the distinguished visitors, $SC_{2}$ began generating tactics to resolve blockages earlier in order to move the swarm out over the CACTF. Figure \ref{Fig:16-Nov} shows spikes in workload around 1003 related to generating the new tactics that were issued at approximately the same time, as shown in Figure \ref{Fig:16-NovTactics}. 

The second overload instance, and the highest (70.23), is related to $SC_{2}$ attempting to determine two things. The status of the remaining blocked vehicles over the launch area, and if the returning vehicles had completed their tactic or were neutralized. A medic's location, near the launch area, was identified 
at the start of the mission. After locating the medic, neutralized UGVs autonomously navigated to it in order to be revived, otherwise they RTL.  Neutralized UAVs autonomously RTL. UAVs are not revived until it is safe for a human mobile medic to walk through the launch area. As soon as a vehicle is neutralized, it is no longer tasked. A large number of UGVs and UAVs were neutralized during the initial deployment, which is evident in the steep decline of active vehicles in Figure ~\ref{Fig:16-NovVehicles}. The mobile medic was deployed around 1015, at which time the UAV batteries were also swapped. As that process was completing, $SC_{2}$ queued tactics to be issued; thus, the spike in workload between 1017 and 1019. 



$SC_{2}$'s estimated workload is in the overload range between 1020 and 1025 as the SC attempts to issue additional tactics with larger numbers of assigned vehicles, which also results in additional blockages. However, the number of active vehicles has again declined due to nerutalizations. Just before 1035 $SC_{2}$ specifies and issues a tactic intended to assign close to sixty vehicles.

Overall, the estimated overall workload mean during the distinguished visitors observation period (1000-1035) was 54.85 (SD = 4.98), with a minimum of 44.1 and a maximum (overload) estimate of 70.23. During this period, 432 workload estimates were generated, with 70 (16.2\%) classified as overload. 

The estimated workload was generally lower after the distinguished visitors departed at 1035. 1,216 workload estimates were generated during this period, of which 57 (4.69\%) were classified as overload. The mean estimated overall workload was 48.92 (SD = 5.91), with a minimum of 32.11 and a maximum of 64.53. 

The longest sustained period three minutes and thirty-five seconds, occurred between 1043 and 1046. During this time, $SC_{2}$ was attempting to assess how many vehicles were still active or were neutralized, and how many UAVs needed battery changes. The mobile medic was deployed and the UAV batteries changed around 1050. The Phase II mission plan was loaded at 1053 and launched just before 1054. Throughout the remainder of the shift, $SC_{2}$ generated vehicle tactics. During the 1054 and 1055 time frame, $SC_{2}$ generated tactics to launch UAV sorties (7 UAVs each) resulting in a thirty second overload with a maximum value of 62.46. $SC_{2}$ tasked UGVs to do various tactics, generally one to four UGVs per tactic, between 1119 and 1122. Forty-eight predominately normal load estimates were generated during that time frame, with a mean of 52.24 (SD = 6.76), of which four were classified as overload. 

$SC_{2}$'s in situ subjective fatigue level was low, 18, just prior to shift start through 1015, with a mean of 24.4 (SD = 8.76), as shown in Table~\ref{Tab:16-NovFatStress}. About 15 minutes prior to shift start, $SC_{2}$ rated the in situ subjective stress level as 7 (i.e.,\ 100 on the normalized scale), the maximum value. At the start of the shift, the in situ subjective stress was rated as a 6, normalized to 83.5 in the table. $SC_{2}$'s reported stress was 83.5 (SD = 11.67) during most of the observation period, but dropped to 52.29 (SD = 9.92) after 1035. The in situ subjective fatigue ratings gradually continued to increase over the remainder of the shift, resulting in a mean of 56 (SD = 8.25).

\begin{table}[htb]
\caption{Nov $16^{th}$ 1000-1200 sift's subjective (Subj.)\ in situ fatigue, stress and overall workload as well as the estimated (Est.)\ overall workload descriptive statistics recorded throughout the shift.} \label{Tab:16-NovFatStress}
\begin{center}
\begin{tabular}{|c||l|l|l|l|}
\hline
\multirow{2}{*}{\textbf{Time}} & \textbf{Subj.\ Overall} & \textbf{Est.\ Overall} & \multirow{2}{*}{\textbf{Stress}} &  \multirow{2}{*}{\textbf{Fatigue}} \\ 
 & \textbf{Workload} & \textbf{Workload} & & \\ \hline\hline
1000 & 31.60 & 53.73 (1.8) & 83.5 & 18 \\ \hline 
1012 & 44.73 & 50.51 (1.96) &  83.5 &	18  \\ \hline
1020 & 44.73 & 58.55 (1.30) & 83.5 &	34  \\ \hline
1030 & 46.38 & 51.95 (1.12) & 67 &	34  \\ \hline
1043 & 39.0 & 60.59 (2.15) & 50.5 &	50.5  \\ \hline
1051 & 48.03 & 45.64 (0.73) & 50.5 &	50.5  \\ \hline
1101 & 42.25 & 50.73 (1.11) & 67	& 50.5  \\ \hline
1111 & 51.15 & 49.63 (0.68) & 50.5 &	50.5  \\ \hline
1120 & 46.38 & 49.91 (2.37) & 50.5 &	50.5  \\ \hline
1130 & 45.55 & 43.74 (1.74) & 50.5 &	50.5  \\ \hline
1140 & 51.33 & 48.25 (2.18) &  34	& 67  \\ \hline
1150 & 43.98 & 50.46 (1.39) & 50.5 &	67  \\ \hline
1155 & 52.98 & 45.04 (2.86) & 67 &	67  \\ \hline
\end{tabular}
\end{center}
\end{table}

The mean overall workload estimates for the minutes at which the in situ subjective assessments were calculated are presented in Table \ref{Tab:16-NovFatStress}. Note that the 1000 in situ ratings were collected seconds before launching the mission plan. The estimated workload is substantially higher than the in situ workload ratings. Eight of $SC_{2}$'s thirteen reported in situ overall workload values were below the estimated overall workload value. All instances where $SC_{2}$'s reported in situ overall workload was above the estimated overall workload occurred after the observation period. The SC reported a subjective high stress level during the distinguish visitor observation period. During this time, the reported in situ overall workload was quite low, even though the SC was doing a large amount of work. 


Stress is a known confound with some physiological metrics (e.g.,\ heart rate). The multi-dimensional workload algorithm is designed to mitigate the effects of other human performance factors (e.g.,\ stress). The individual workload component results (see Appendix \ref{Appx:16-Nov-Results} in Table \ref{Tab:16-NovFatStress}) highlight that the cognitive workload component's metrics appear to be less susceptible to stress, but the physical workload component is influenced by stress and possibly fatigue. Throughout the shift the cognitive workload estimates loosely track the in situ cognitive workload. A limitation is the in situ query's 7 point Likert scale. However, the physical workload estimates are high relative to the corresponding in situ physical workload. This apparent over estimation appears to be due to the heart rate and respiration rate metrics, which are known to be impacted by other human performance factors and represent two of the three metrics for estimating physical workload. Two very high physical workload estimates, at 1000 (61.34) and 1020 (51.73) appear to be due to $SC_{2}$'s stress level (83.5). Two additional instances occurred at 1043 (62.8) and 1120 (51.73). At these times, $SC_{2}$ reported moderate stress, but the fatigue level increased to a moderate level. It is known that $SC_{2}$ was not physically active enough to obtain an overload physical workload estimates, which indicates a clear influence from stress early in the shift and the combination of stress and fatigue later in the shift.  

During a post-FX debrief, $SC_{2}$ commented that this shift resulted in the highest subjective stress level, and at the end of the shift, $SC_{2}$ was very fatigued. After shift completion, $SC_{2}$ indicated a lower stress level, as the major goal had been completed and the CCAST swarm had performed well. 
 
 \paragraph{Nov $17^{th}$ 1200-1400 Shift:} 
 A particularly challenging shift occurred on Nov $17^{th}$, during which the wind gusts were the highest, 28 MPH, CCAST had experienced while on shift (see Appendix \ref{Appx:Conditions} Table \ref{Tab:FX6Info}). 
The wind created a number of issues. The pre-mission brief indicated that 118 hardware vehicles (10 UGVs and 108 UAVs) were to be deployed during this shift. The estimated workload components were cognitive, speech, auditory and physical. 

\begin{figure} [!htb]
    \centering
    \includegraphics[height=3in]{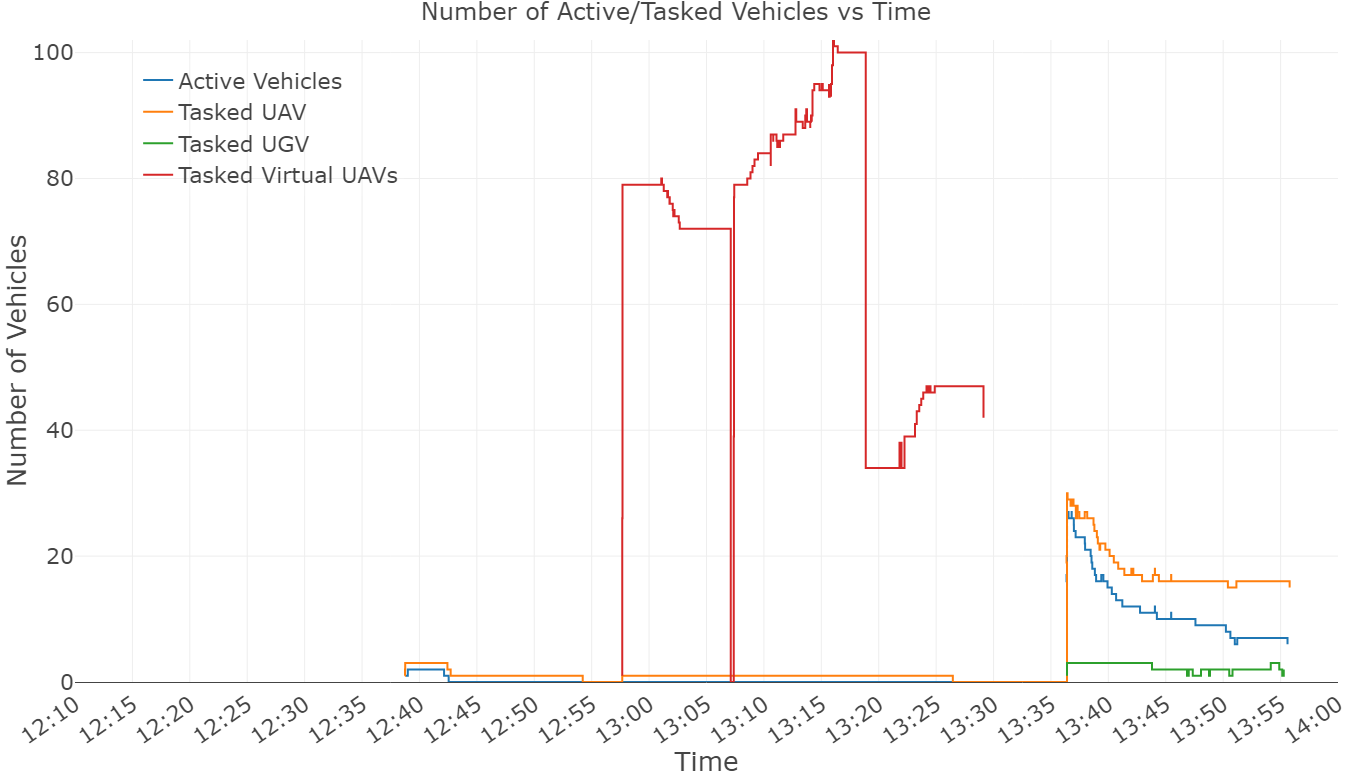}
    \caption{The Nov $17^{th}$ 1200-1400 shift's tasked (by vehicle type) and active vehicles.}
    \label{Fig:17-NovVehicles}
\end{figure} 

The intention at shift start was to test fly one 3DR Solo and one VOXEL M500, as the CCAST team had never flown the UAVs in such high winds. However, the LTE system became a continual problem for the first hour and a half, requiring multiple restarts. Each time the LTE restarts, all vehicles and the dispatcher must be restarted. An I3 restart is not required, but I3 is usually restarted. The LTE issues resulted in no vehicles being deployed early in the shift, as shown in Figure \ref{Fig:17-NovVehicles}. It is also important to note that if the vehicles have intermittent, or no communication with the dispatcher and I3, then the telemetry is not logged, and cannot be represented in the figures related to tasked vehicles or blockages.

\begin{figure}[!htb]
\centering
\begin{subfigure}{.9\textwidth}
  \centering
   \includegraphics[width=.9\linewidth]{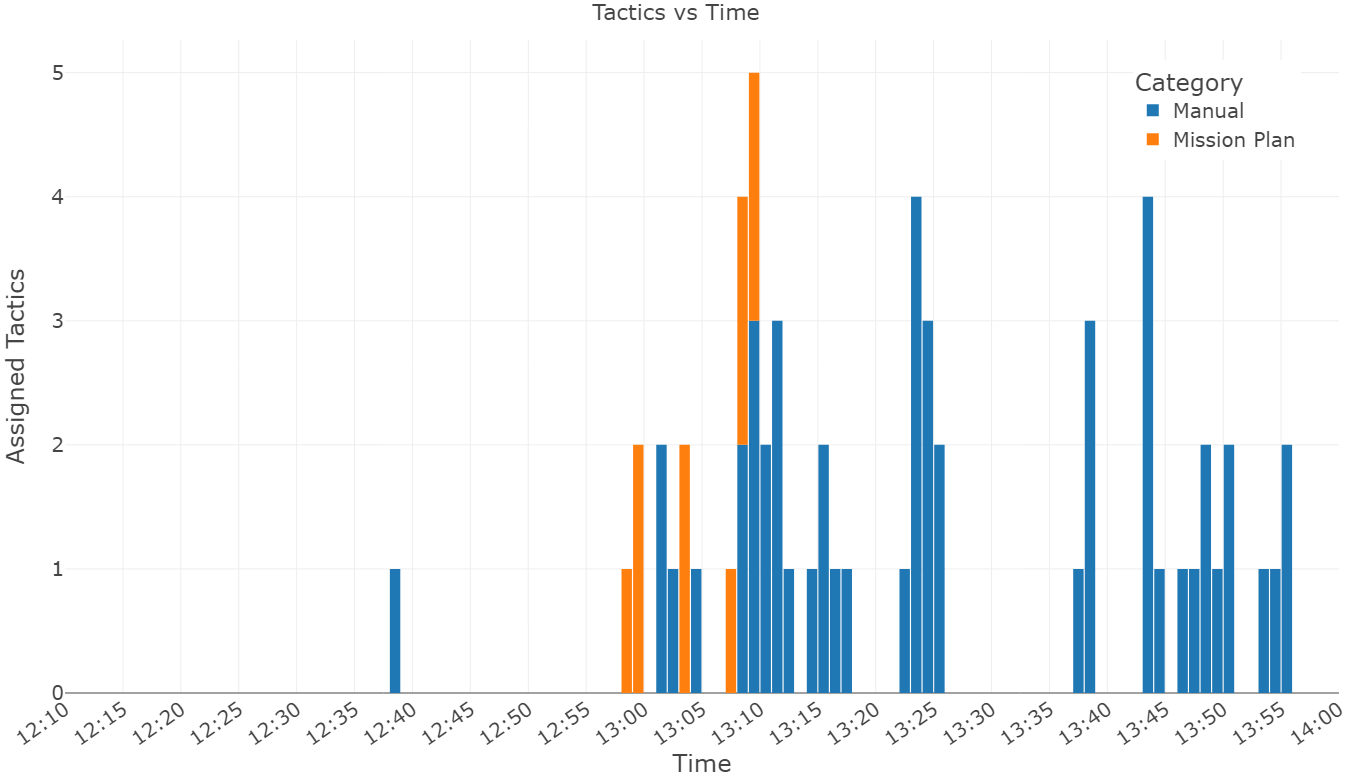}
    \caption{Issued tactics. }
    \label{Fig:17-NovTactics}
\end{subfigure}%
\hfill
\begin{subfigure}{.9\textwidth}
  \centering
   \includegraphics[width=.9\linewidth]{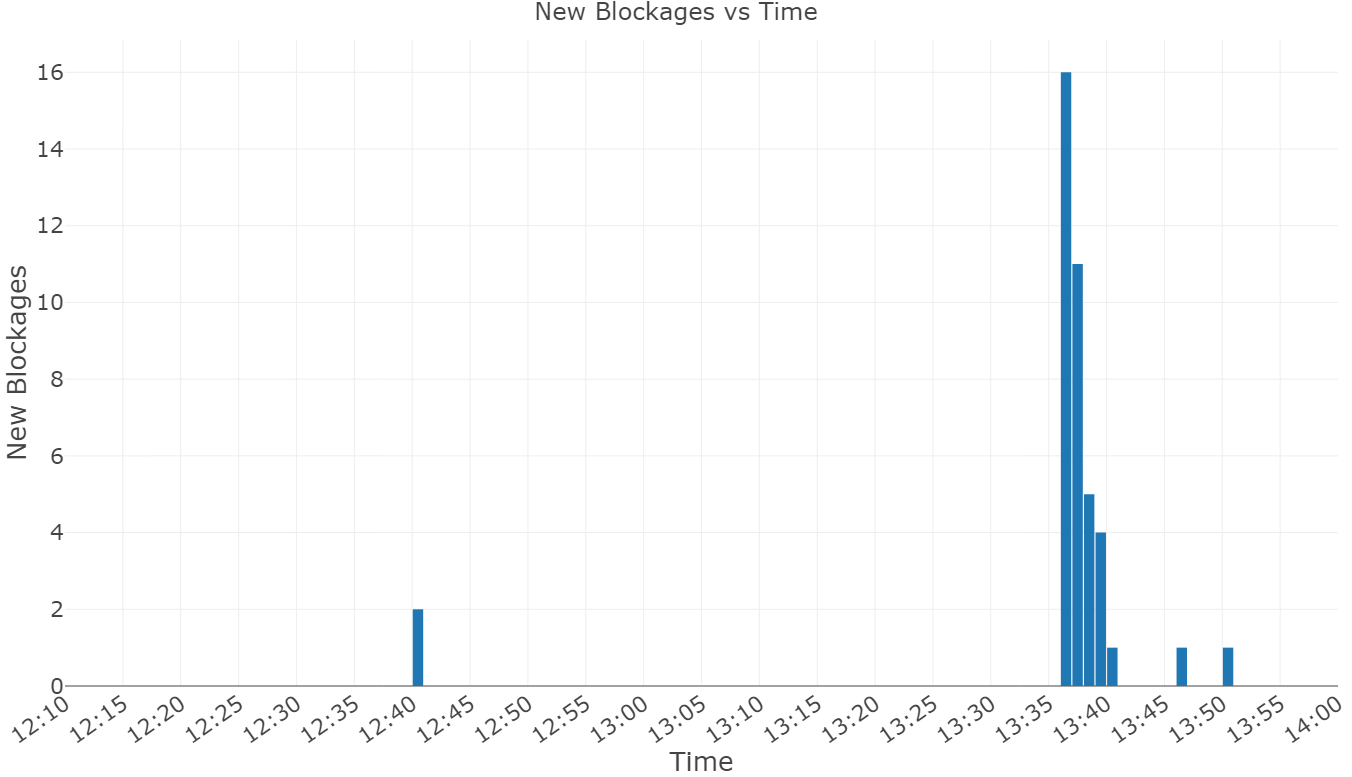}
    \caption{Blockages.}
    \label{Fig:17-NovBlocks}
\end{subfigure}%
\caption{The Nov $17^{th}$ 1200-1400 shift's (a) issued tactics and (b) vehicle blockages by the minute.}
\end{figure}

\begin{figure}[!hbt]
    \centering
    \includegraphics[height=3.5in]{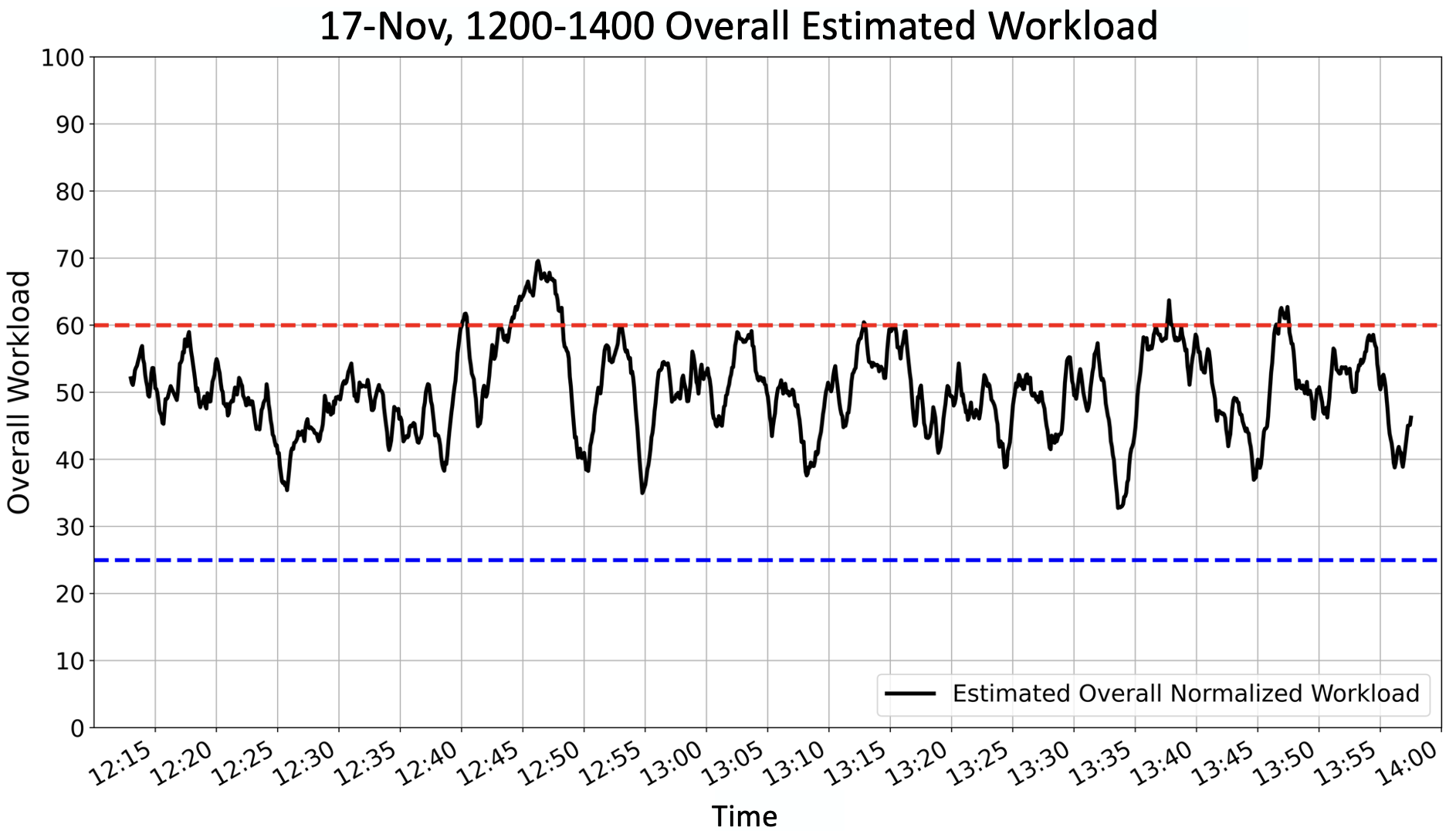}
    \caption{Overall workload estimates for Nov $17^{th}$ 1200-1400 shift. }
    \label{Fig:17-Nov}
\end{figure}

At approximately 1230, it was believed that the LTE issues were resolved and the objective of test flying the UAVs proceeded. $SC_{2}$ generated the tactic at 1237, but the Unity engine required for I3 crashed and had to be restarted. $SC_{2}$ issued the tactic at 1238, as shown in Figure~\ref{Fig:17-NovTactics}. The two vehicles were tasked and active shortly thereafter, as shown in Figure~\ref{Fig:17-NovVehicles}. I3 did not show the tactic visualization, which the SC fixed on the fly and resulted in an estimated overload state at 1240, as shown in Figure~\ref{Fig:17-Nov}.

The LTE issues persisted, at 1245 the team restarted the dispatcher and I3 using virtual vehicles, 10 UGVs and 20 UAVs. $SC_{2}$ created explicit tactics and attempted to issue them, but system issues persisted. The longest sustained estimated overload state duration across all the shifts occurred between 1244 and 1248, a duration of four minutes and fifteen seconds. The mean estimated overall workload during this time frame was 65.16 (SD = 2.35, min = 60.04). The shift's overall maximum estimate, 69.59, occurred during this time period as well. During this period, the cognitive workload estimate was consistently overloaded and the speech workload estimate was frequently overloaded, which is aligned with the SC's activities.

Given the persistent LTE issues, at 1256 the number of virtual vehicles increased to 20 UGVs and 105 UAVs. $SC_{2}$ loaded the mission plan at 1257, but before issuing the plan, wanted to verify that the mission plan was not going to task hardware vehicles in the launch area, given that the LTE was connected. After receiving such verification, the first mission plan signal sent vehicles to the West side of the CACTF at 1300. A second mission plan signal was issued at 1301 sending vehicles to the East side of the CACTF, and the final signal within the same minute, sent vehicles to the center of the CACTF. This activity is shown in Figure \ref{Fig:17-NovTactics}, recall that simulated vehicles cannot be blocked, so no vehicle blockages occurred per Figure \ref{Fig:17-NovBlocks}. The simulated vehicles were not providing artifacts, so the entire system was again shut down and restarted, which is shown as the drop in tasked simulated UAVs in Figure \ref{Fig:17-NovVehicles}. 

$SC_{2}$ reloaded the mission plan at 1307 and began issuing the mission plan signals at 1308. $SC_{2}$ generated and issued a number of tactics between 1310 and 1318 (see Figure \ref{Fig:17-NovTactics}) that increased the number of tasked simulated vehicles to above 100. This activity increased the overall workload estimates, but they generally remained within the normal range. 

At 1328 all tasking of the simulated vehicles stopped and at 1330 the CCAST team restarted with hardware (10 UGVs, 71 UAVs) and virtual (10 UGVs, 20 UAVs) vehicles. The mission plan was executed at 1331, but I3 was not updating with the vehicle telemetry and $SC_{2}$ changed the communication port at 1135, which provided telemetry. Due to not receiving the telemetry, it incorrectly appears that no vehicles launch until 1336 in Figure \ref{Fig:17-NovVehicles}. Almost immediately after the mission plan launch, 3DR Solos' began dropping from the sky\footnote{As noted in Table \ref{Tab:Robots}, the 3DR Solos are an older technology. Two hypotheses exist as to why they failed. The primary hypothesis is that the wind caused the UAV to exceed its maximum configured pitch/roll, causing it to stop making adjustments. The alternative hypotheses are that the barometer configuration was a problem or a hardware failure occurred.}. Once the telemetry was restored, $SC_{2}$ began attempting to command all 3DR solos to RTL at 1337. 

This overall period resulted in elevated overall workload estimates compared to other portions of the shift. During this period, the estimated cognitive, physical and speech workload values all increased. This period of time was stressful given that $SC_{2}$ was unable to issue tactics until telemetry was restored and team members were asking $SC_{2}$ to get the tactics issued to RTL the vehicles. The increases in the cognitive and speech components appear to be similar in magnitude to other high workload periods representative of $SC_{2}$'s increased work. However, the high physical workload estimates are possibly due to $SC_{2}$'s increased stress. Due to the ten minute timing between in situ ratings, no such ratings were recorded during this period, and it is not possible to clearly align $SC_{2}$'s perceived stress with the high physical workload estimates. 

The remainder of the shift $SC_{2}$ was attempting to move UGVs from the launch zone to a building. Multiple UGVs were assigned tactics, and the UGVs were not responding as expected. $SC_{2}$ was having conversation with the team leader about this situation. $SC_{2}$ was also verifying that tasked UGVs had tactics, and whether or not the vehicles were doing their tasks, while verifying information for and receiving instructions from the team leader.  The overload estimates between 1346 and 1347 are a result of these efforts.

$SC_{2}$'s reported relatively low (18-34) in situ stress values throughout the shift, as shown in Table~\ref{Tab:17-NovFatStress}, with a mean of 27.6 (SD = 8.26). Similar to $SC_{2}$'s Nov $16^{th}$ shift, the in situ fatigue level was low, 18, at shift start, and gradually increased over the shift, resulting in a mean of 45.6 (SD = 13.47).

\begin{table}[htb]
\caption{The Nov $17^{th}$ 1200-1400 shift's subjective in situ fatigue, stress and overall workload as well as the estimated overall workload descriptive statistics recorded throughout the shift.} \label{Tab:17-NovFatStress}
\begin{center}
\begin{tabular}{|c||l|l|l|l|}
\hline
 \multirow{2}{*}{\textbf{Time}} & \textbf{Subj. Overall} & \textbf{Est. Overall} & \multirow{2}{*}{\textbf{Stress}} &  \multirow{2}{*}{\textbf{Fatigue}} \\ 
 & \textbf{Workload} & \textbf{Workload} & & \\ \hline\hline
1220 & 24.4 & 50.58	(2.85) & 18 &	18  \\ \hline
1234 & 18 & 45.19 (2.51) & 18 &	34  \\ \hline
1250 & 30.05 & 44.18 (4.57) & 18 &	34  \\ \hline
1300 & 31.6 & 48.76	(3.42) & 34&	50.5 \\ \hline
1310 & 33.3 & 50.93	(1.87) & 18 &	50.5  \\ \hline
1320 & 46.38 & 50.37 (1.96) & 34	& 50.5  \\ \hline
1330 & 30.98 & 50.62 (1.62) & 34 &	50.5  \\ \hline
1340 & 41.5 & 54.90	(1.81) & 34 &	50.5 \\ \hline
1350 & 40.6 & 48.52	(1.62) & 34 &	50.5  \\ \hline
1355 & 39.85 & 48.38 (3.80) & 34	& 67  \\ \hline
\end{tabular}
\end{center}
\end{table}

The in situ subjective overall workload and the corresponding mean overall workload estimates are presented in Table \ref{Tab:17-NovFatStress}. The estimated workload is generally higher than the in situ ratings. The only planned in situ data collection that corresponded with a high estimated overall workload at 1240 was missed, due to distraction.  

The shift's mean estimated overall workload was 50.25 (SD = 6.52, min  = 32.76, max = 69.59), see Table \ref{Tab:WLMean}. 6.1\% of the shift's overall workload estimates were classified as overload, per Table \ref{Tab:WLState}. The in situ and associated workload component estimates are provided in Appendix \ref{Appx:ShiftResults} Table \ref{Tab:17-NovWLSubjObj}, respectively.

\paragraph{Joint Integrator and SCs Shift, Nov $18^{th}$ 1330-1530:}
The joint integrator shifts were the first instances of both integrator teams operating on the CACTF simultaneously. DARPA's objective was to deploy the largest swarm ever.
There was no direct communication between the two teams', rather the CACTF was spatially divided, with the CCAST team being responsible for the South half closest to C2. The only information CCAST received was the other team's vehicle telemetry via I3 glyphs similar to Figure~\ref{fig:glyph} that did not have tactics or a tactic icon, with either an empty (0\%) or full (100\%) battery, no vehicle capabilities (e.g.,\ electronic warfare), and a vehicle identifier that differed (e.g.,\ atx10) from the CCAST identifiers. 

\begin{figure} [hbt]
    \centering
    \includegraphics[height=1.5in]{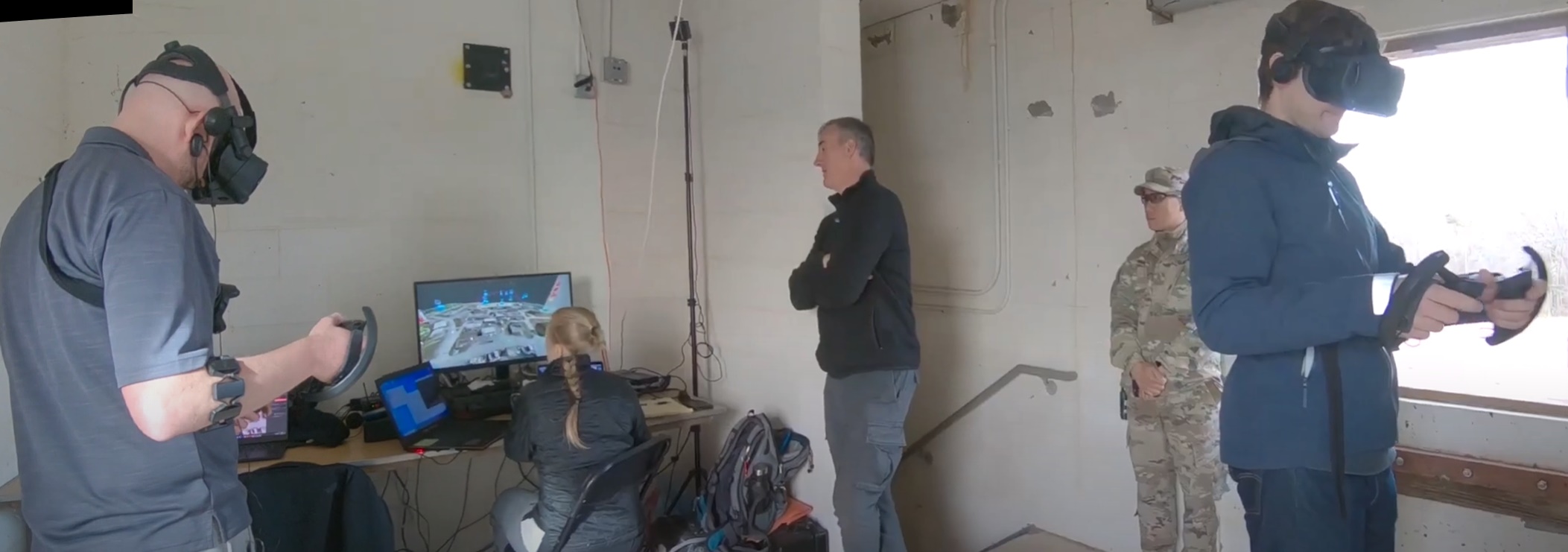}
    \caption{The C2 configuration accommodating two swarm commanders. Photo courtesy of DARPA. }
    \label{Fig:2i3}
\end{figure}

During this shift both SCs simultaneously commanded the swarm. Each SC had their own I3 station, as the CCAST system communicates tactics and vehicle telemetry resulting from each SC. The I3 stations were set up in the C2 SC room, one on each side, as shown in Figure \ref{Fig:2i3}. The SCs' spilt CCAST's assigned CACTF area at the C2 building, with $SC_{1}$ being responsible for the West side and $SC_{2}$ having responsibility for the East.  The two SCs were able to directly speak to one another, but were unable to see what tactics the other was creating until the tactics were issued and assigned to vehicles. 

Per the mission brief, the CCAST team placed 140 vehicles, 30 UGVs and 110 UAVs in the launch area, while the other integrator team had 90 UAVs, 90 UGVs and one vertical takeoff and landing fixed wing aerial vehicle. CCAST added 40 virtual UAVs and 10 virtual UGVs later in the shift, totaling 190 unique CCAST vehicles. The CCAST SCs deployed 110 unique vehicles.

It was predetermined that $SC_{2}$ was responsible for the mission plan and any associated signals. During this shift, workload and performance data was collected for $SC_{1}$ only. The recorded data captured the cognitive and physical components. The auditory workload was estimated using the procedure described in Section \ref{DepVars}. The microphone malfunctioned; thus, speech and visual workload were estimated using the respective IMPRINT Pro models' values. CCAST mission plan and telemetry data logging issues occurred during this shift, the source of which is not clear. Since dual SCs was not specifically a design consideration, the tactics log did not identify which SC explicitly issued tactics. As such, the analysis cannot distinguish who issued which tactics. This mission plan tactics were also not clearly logged. 

\begin{figure} [htb]
    \centering
    \includegraphics[height=3in]{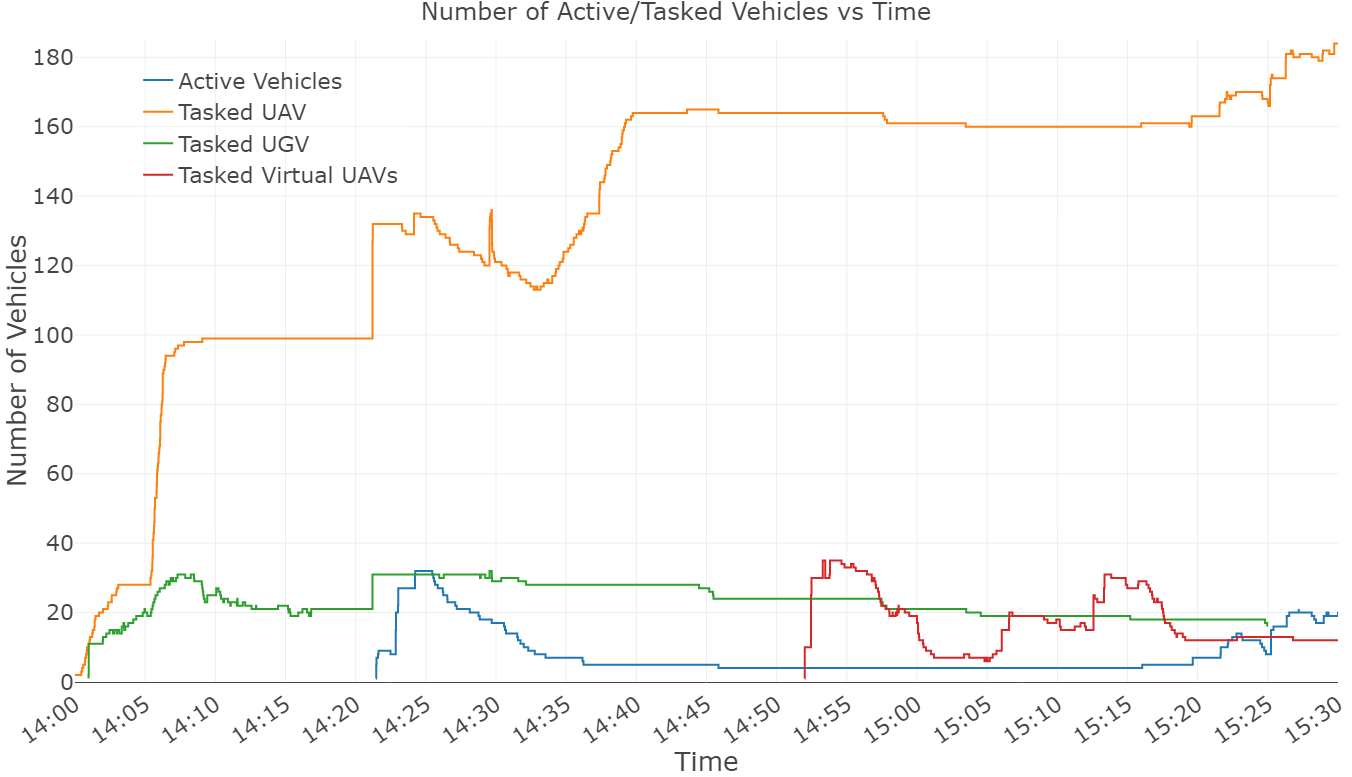}
    \caption{The Nov $18^{th}$ 1330-1530 shift's tasked (by vehicle type) and active vehicles.}
    \label{Fig:18-NovVehicles}
\end{figure} 

\begin{figure}[!htb]
\centering
\begin{subfigure}{.9\textwidth}
  \centering
   \includegraphics[width=.9\linewidth]{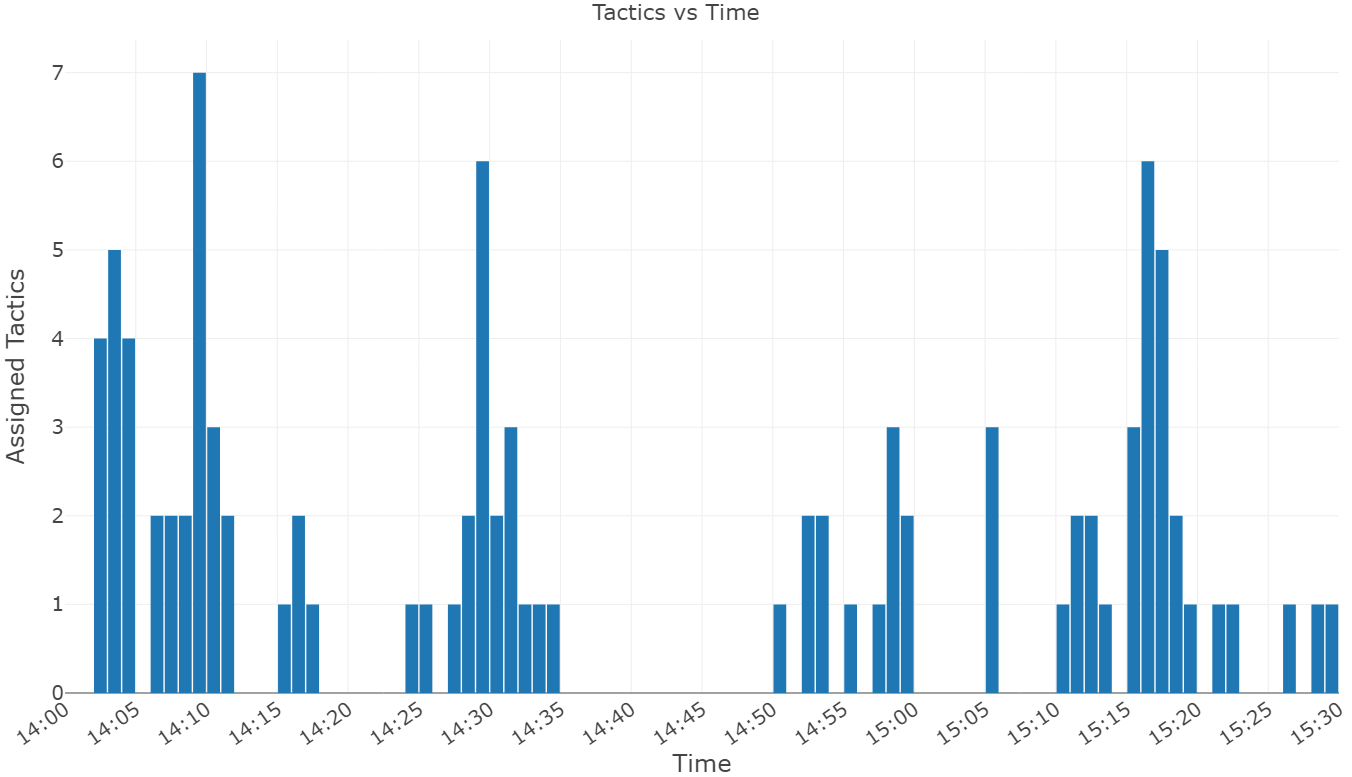}
    \caption{Issued tactics. }
    \label{Fig:18-NovTactics}
\end{subfigure}%
\hfill
\begin{subfigure}{.9\textwidth}
  \centering
   \includegraphics[width=.9\linewidth]{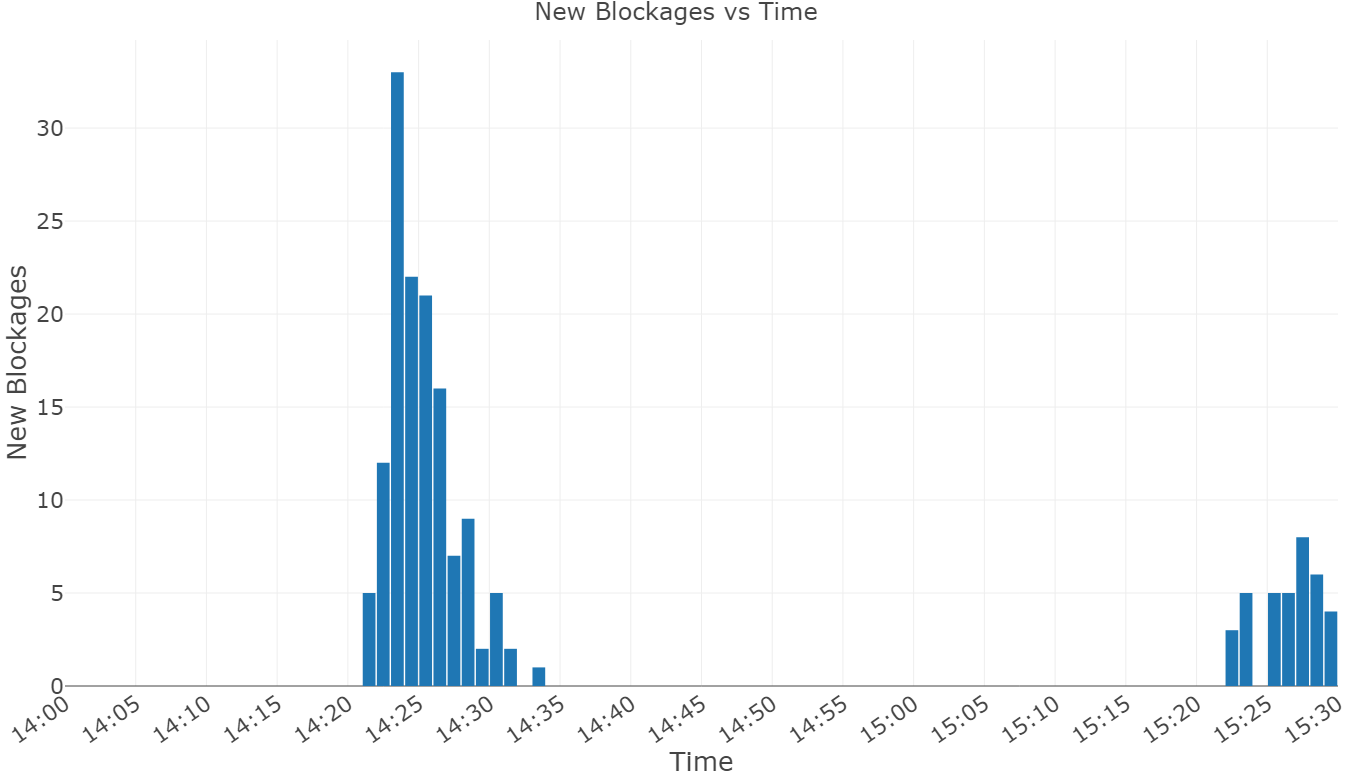}
    \caption{Blockages.}
    \label{Fig:18-NovBlocks}
\end{subfigure}%
\caption{The Nov $18^{th}$ 13300-1530 shift's (a) issued tactics and (b) vehicle blockages by the minute.}
\end{figure}

The shift start was delayed until 1400, at which time $SC_{2}$ loaded the mission plan and executed the first signal, intended to deploy all rovers around CCAST's assigned portion of the CACTF. The experimenter notes, video, and the tasked UGVs (green line in Figure \ref{Fig:18-NovVehicles}) show that $SC_{2}$ fetched and launched the mission plan at 1400 and 1401, respectively. However, the mission plan tactics do not appear in the log file, hence they are not in Figure \ref{Fig:18-NovTactics} (shown as orange for the prior shifts' results). The SCs immediately began stopping the UGV's Surveil tactics and explicitly issued the tactics, which is shown in the tasked agents figure between 1403 and 1410. A number of tactics were issued between 1406 and 1411, including UAVs tactics; however, not all UAVs actually launched. The active agents (blue line in the figure) during this time period were not accurately logged, perhaps due to the shear volume of information from both teams creating logging issues. It is also likely that the LTE was beginning to demonstrate problems, that became more evident later in the shift. It is also important to note that for an unknown reason the number of tasked vehicles only increased with each new tactic, and did not decrease, as shown in Figure \ref{Fig:18-NovVehicles}. 

\begin{figure} [bht]
    \centering
    \includegraphics[height=3.5in]{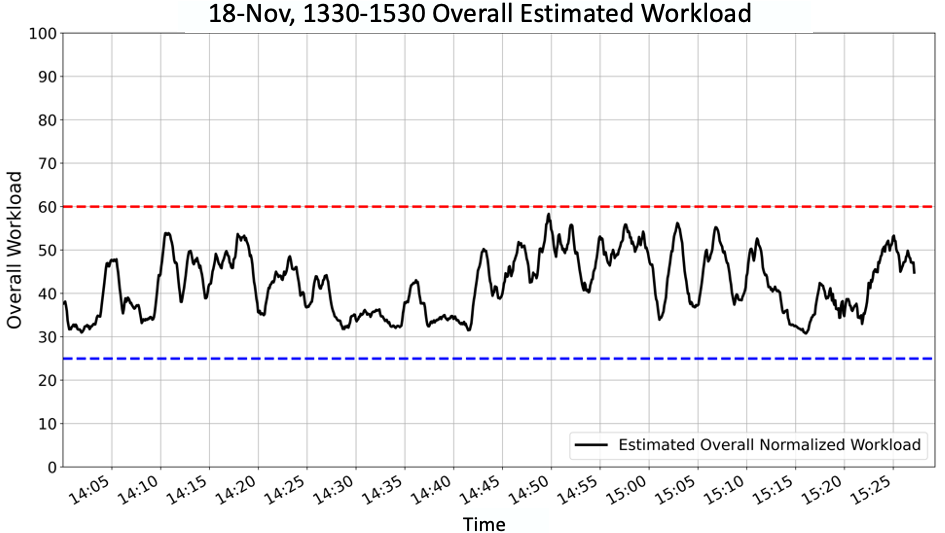}
    \caption{Overall workload estimates for the last FX shift (Nov $18^{th}$, 1330-1530), a joint integrator shift when both SCs simultaneously commanded the swarm.  Results recorded for $SC_{1}$ only. }
    \label{Fig:18-Nov}
\end{figure}

Throughout this initial deployment period, and the entire shift, $SC_{1}$'s estimated overall workload remained in the normal range, as shown in Figure \ref{Fig:18-Nov}. $SC{1}$'s estimated overall workload increased at 1411 after the tactics were issued, but vehicles did not launch. While the estimates oscillated a bit, these higher estimates persist until 1419 as the team attempts to determine why vehicles were not launching. 

$SC_{2}$ fetched the mission plan again at 1420 and launched it a minute later, note these tactics are not shown in Figure~\ref{Fig:18-NovTactics} due to the logging issues. The tasked vehicles in Figure \ref{Fig:18-NovVehicles} show these tactics at 1421. At this time, some of the active vehicles are shown as blocked in Figure \ref{Fig:18-NovBlocks}. The data log files do not indicate that the SCs attempted to mitigate the blockages by issuing explicit tactics. However, it is possible that some tactics were not logged due to communication issues. It is noted that $SC_{1}$'s estimated overall workload increased at this same time. 

Beginning at about 1435, the SCs cannot communicate with the vehicles. It was determined that an LTE sector problem existed, requiring a rest. All vehicles on the launch pad were restarted at 1442. The LTE restarts, followed by vehicle restarts occurred again between 1450 and 1452, respectively. Telemetry data was not recorded during these time periods. 


Virtual vehicles were added at 1450 to which the SCs issued explicit tactics for gathering information. $SC_{1}$'s estimated overall workload peaked just before the switch to virtual vehicles. The SCs' were careful to not task hardware vehicles that were back in communication, as the LTE issues continued. During this time, $SC_{1}$'s overall workload was quite high due to having to select specific vehicles for the tactics. The selection of specific simulated vehicles was not impacted by the display of the other integrator team's vehicle telemetry, because of DARPA's intentional splitting of the CACTF between the two teams. It is hypothesized that if this spatial CACTF split did not exist, and the two team's vehicles were intermixed, the SCs' overall workload associated with this task will increase due to having to differentiate between CCAST's hardware and software vehicles, as well as the other team's vehicles. 

At 1500 $SC_{1}$ indicated that there was a lot of stutter in the I3 display. The videos of the I3 display were recorded on the machine running I3. $SC_{1}$ stopped and restarted the video at 1305-1306, which resolved the issue. $SC_{1}$'s spike in estimated overall workload was due to resolving this issue. During this period, the LTE was again reset at 1500 and the vehicles were powered up at 1505. The SCs were still issuing explicit tactics to simulated vehicles between 1510 and 1514. The SCs began issuing explicit tactics for hardware vehicles around 1517, and $SC_{1}$'s estimated workload increased. $SC_{2}$ issued a series of tactics to launch five UAVs, with the goal of neutralizing a fortified artifact that requires multiple vehicles interact with the artifact simultaneously; however, only one UAV launched. Simultaneously, $SC_{1}$ created a Surveil tactic using a large number of UAVs for a building near the fortified artifact. 

Both SCs issued explicit tactics through the rest of the shift, as seen in Figures \ref{Fig:18-NovVehicles}, \ref{Fig:18-NovTactics}, and b. During this period, especially the last few minutes of the shift, the SCs were verbally coordinating with one another, as they were both issuing tactics to the East side of the CACTF. $SC_{2}$ continued to focus on neutralizing the fortified artifact, while $SC_{1}$ was issued Surveil tactics for ten UAVs to investigate two buildings. Throughout this final push, $SC_{1}$'s estimated overall workload increased. 

This shift resulted in some of $SC_{1}$'s highest reported in situ stress ratings, as shown in Table \ref{Tab:18-NovFatStress}, with seven of ten responses being $> 50$. The SCs both reported that selecting virtual vehicles for explicit tactics, when some hardware vehicles may be in communication, but are not to be tasked, is stressful. $SC_{1}$'s fatigue level was moderate throughout the shift. 

\begin{table}[htb]
\caption{The Nov 18$^{th}$ 1300-1500 shift's subjective in situ fatigue, stress and overall workload as well as the estimated overall workload descriptive statistics recorded throughout the shift.} \label{Tab:18-NovFatStress}
\begin{center}
\begin{tabular}{|c||l|l|l|l|}
\hline
  \multirow{2}{*}{\textbf{Time}} & \textbf{Subj. Overall} & \textbf{Est. Overall} & \multirow{2}{*}{\textbf{Stress}} &  \multirow{2}{*}{\textbf{Fatigue}} \\ 
 & \textbf{Workload} & \textbf{Workload} & & \\ \hline\hline
1400 & 39.78 & 34.43 (2.67) & 50.5 &	34  \\ \hline
1410 & 38.13 & 51.83 (2.77) & 50.5 & 50.5 \\ \hline
1420 & 43.9 & 36.44	(1.78) & 67 &	34  \\ \hline
1430 & 46.38 & 34.96 (0.83) & 50.5	& 34  \\ \hline
1440 & 18.8 & 33.64	(0.60) & 18 &	34  \\ \hline
1450 & 34.83 & 51.56 (1.85) & 18 &	50.5  \\ \hline
1500 & 55.45 & 43.62 (6.48) & 67 &	50.5  \\ \hline
1510 & 37.38 & 48.79 (1.95) & 34 & 50.5 \\ \hline
1520 & 47.2 & 37.30	(1.00) & 67 &	34  \\ \hline
1525 & 48.85 & 49.01 (2.84) & 67	& 50.5  \\ \hline
\end{tabular}
\end{center}
\end{table}

The in situ subjective overall workload and corresponding mean overall workload estimates are presented in Table \ref{Tab:18-NovFatStress}. The estimated workload is generally higher than the in situ ratings, and $SC_{1}$'s results across the shifts show this was a common result. $SC_{1}$ generally reported lower in situ subjective workload as compared to $SC_{2}$ across shifts, as shown in Table \ref{Tab:WLSubj}. While $SC_{1}$ subjectively reported higher overall workload at 1420 then the two earlier points, $SC_{1}$ was not actively doing anything, as the team was waiting for $SC_{2}$ to issue the mission plan, which is reflected in the lower estimated workload value for the same minute.  A similar situation existed at 1430. Both the estimated cognitive and physical workload components for these time periods are very low, as shown in Appendix \ref{Appx:17-Nov-Results} Table \ref{Tab:18-NovWLSubjObj}, and do not appear to be impacted by stress. However, as the SCs created explicit tactics for the vehicles at 1450, $SC_{1}$'s reported subjective overall workload was lower than the estimated workload. At 1500, when $SC_{1}$ was dealing with the visual stuttering issue, the reported workload and stress were high, with the associated estimated workload being lower. The physical workload component estimate is higher than expected, which may be caused by stress.  

Overall, $SC_{1}$'s estimated overall workload remained in the normal range the entire shift, with a mean of 43.02 (SD = 6.74, min = 32.37, and max = 59.37), as shown in Table \ref{Tab:WLMean}. $SC_{1}$'s mean estimated overall workload was the lowest of all this SC's CACTF shifts. While $SC_{1}$ did not take responsibility for executing mission plans, both SCs explicitly generated tactics for vehicles during the shift. Two SCs and the clear allocation of CACTF area responsibility, East vs.\ West appear to have reduced $SC_{1}$'s overall workload, even though this was a joint shift with the CCAST team placing the highest number of hardware vehicles in the launch zone, while also incorporating two CCAST SCs for the very first time.

The overall workload estimates reveal important insights related to the SCs' workload, particularly over very long shifts, that are not attainable otherwise. 
The analysis across the three shifts provides additional evidence, beyond Adams' prior work with Fortune, Harriott and Heard, that the multi-dimensional workload algorithm demonstrates sensitivity to known changes in the SC's workload, even when some workload components are provided by the using a single supervisor-single UAV evaluation's IMPRINT Pro models \cite{Heard2020CogSima}. Overall, the reported results provide the first use of this estimation method to single human-swarm robot deployments in an actual urban operational environment. 

Even though the focus of this section was the overall workload estimates, the individual instances for each workload component can be similarly plotted for additional analysis. While cognitive workload tends to be the primary research focus in the general literature, domains that deploy very complex systems in differing environmental operational conditions impact the workload components' contribution to overall workload differently. Thus, it is critical for safe operation to understand all aspects of workload. 

\subsubsection{Workload Component Contributions}

The cognitive workload component (i.e.,\ channel) has traditionally been the focus of the relevant literature; however, other components can and do contribute to overall workload. The CCAST SC's supervisory interaction is one that is heavily dependent on visual perception, in particular multiple object visual perception, which implies the visual workload component will be a primary contributor to overall workload. However, traditional visual perception multiple object tracking research, for example \cite{WolfeVizSci2020} assumes all visual targets (e.g.,\ vehicles, artifacts) exist on the display simultaneously. Visual tracking of multiple objects via a virtual reality head mounted display, such as that integrated into I3, is nascent and has a different focus \cite{KibleuretalVIP2019} from traditional multiple object visual tracking research. Current efforts in Adams' group are using eye trackers to objectively estimate visual workload, but that technology was not feasible with the existing I3 virtual reality system. 

The speech and auditory workload components are important for the CCAST SC. Often the SC was talking to others to communicate the swarm's current state, verifying that it was safe to issue the mission plan or SC generated tactics (i.e.,\ ensuring that all personnel in the launch area were a safe distance away from the vehicles), or verifying received verbal information from the team member responsible for designating the mission plan and tactics to be issued (i.e.,\ ``Surveil building X in the Northwest corner''). Often associated with these speech acts were auditory components; thus, there is a reasonable demand on both the SC's speech and auditory channels throughout the mission.

Traditional supervisory workstations, in which the human uses a mouse or joystick to interact with a system via a two dimensional interface 
\cite{Heard2020CogSima,CummingsJCEDM2019} do not place a high demand on the physical workload component. However, the CCAST SCs prefer to use I3 while standing, and $SC_{1}$ frequently tended to physically move around the C2 workspace while supervising the swarm.
$SC_{1}$ learned to use the virtual reality tracking devices to determine when the SC's physical placement in the workspace had positioned the SC too far from the trackers, at which point the SC repositioned appropriately. Further, I3 relies on the virtual reality controllers as the SC's inputs to the system; thus, the SC's arms are moving frequently, contributing to physical workload. Thus, the physical workload component is expected to contribute even more to the SCs' overall workload. 

Stress and fatigue are known to impact heart rate, respiration rate, and to some extent heart rate variability. Thus algorithms that estimate cognitive or overall workload using these metrics are impacted by increased stress and fatigue. The multi-dimensional workload components is one means of mitigating the impacts of stress and fatigue on the multi-dimensional workload estimation algorithm's estimates. The cognitive component estimates appear to have limited impact from stress and fatigue; however, physical workload appears to be overestimated. This impact appears to be evident at the start of the Nov $16^{th}$ shift, when $SC_{2}$ reported high subjective stress and continued to report the same high stress level at 1012 and 1020 while generating a large number of tactics. The 1020 physical workload estimation appears to be potentially impacted by stress (see Appendix \ref{Appx:ShiftResults} Table \ref{Tab:16-NovWLSubjObj}, but may also be associated with the generation and issuing of fourteen tactics, a very high number, between 1019 and 1020. 

\subsubsection{Qualitative Swarm Commander Insights}

The post-FX debrief provided some direct insights. The SCs were asked if they felt there were any days or shifts for which they were \textit{unable to sustain their effort or performance due to being overloaded}. $SC_{1}$ indicated feeling ``red lined'' when continual explicit tactic generation was needed, and was unable to trust the CCAST system's to automatically allocate vehicles to tactics. $SC_{1}$ indicated two cases during which this situation arose. During  ``one of the first [live-virtual] shifts when we realized the allocation routines would happily dish out virtual and real platforms to tactic requests. At one point I was instructed to pick through the staging area to only fire off simulated quads, knowing that a mistake there would be potentially dangerous to the safety spotter crew ...'' An example of this situation occurred when CCAST was testing sprinter integration technologies during the Nov $13^{th}$ shift. 
This shift resulted in the estimated overall workload being classified as overload 25 times, or 2.26\% of the estimates, $SC_{1}$'s second highest number of overload instances during a shift. $SC_1$ also noted another situation that was perceived as causing a high overloaded state. LTE communication issues resulted in significant latency and periods of time without telemetry updates. $SC_{1}$ stated: ``When the telemetry started backing up to such an extent we were at least half minute, maybe more, out of sync. We proceeded to task platforms and mission plan elements with full knowledge that I3 and Dispatcher had no accurate picture of the current platform positions.  We were flying blind.''  
$SC_{2}$ did not subjectively perceive reaching a state where performance was impacted by being overloaded; however, 
did indicate that ``during the longer shifts ...(the $> 3$ hour ones), I consistently felt a good amount of physical fatigue near the end, and probably [would not] have lasted much longer standing up, but that [could have] been alleviated by
sitting without really impacting my performance in I3."

The SCs were asked to comment on \textit{what factors [you felt] increase your workload}? Both SCs mentioned needing to communicate with another team member. One SC noted: ``X talking to me, especially when he was rapidly switching between asking for info vs.\ asking for new tasking.'' The other SC also commented that ``Communication/coordination with people outside of I3 does incur a cost. I [would not] rate [the impact to be] large, except for the necessary context switch as we attempt to understand what is being asked of us, then re-submerging into I3.'' 

$SC_{1}$ felt that using mission plans had limited impact on increasing workload. ``Until/unless the mission plans become more malleable through [I3], [they are] fairly low workload. Anxiety is \textit{really high} as the mission kicks off, and a single signal misstep can have very negative consequences, but [the SC is not] really \textit{doing} a lot. There is no real mechanism for [the SC] to consider modification or amendments to the mission plan structure, observing and reacting to new intelligence or threats appearing.'' Generally, the mission plans were used as defined and reaction to new intelligence or threats was handled by the SC generating and issuing new tactics. 

Both SCs noted that generating tactics with an explicit vehicle selection impacted workload.  One noted ``... [single vehicle selection is] not only tedious, but error prone, [needing] to query platforms to cobble together enough to execute a Surveil object. Whenever we [could not] rely on [the system for automatic] allocation, [the SCs] were stuck in the mud, and [could not] split attention to focus on any higher level tasks.'' 

A related tactic generation issue occurred when explicit waypoint designation was required. $SC_{1}$ noted, ``It can be painstaking to … lay down a waypoint within some tight bounds (2' of expected artifact location). There is a fair amount of spatial estimates to gauge inaccuracies in pose estimates, platform locations (GPS), as well as obstacle bounds and buffer space - also with some superficial understanding of how the route/path planner works, and it may discard positions too close together, or it may lock onto the road network under certain conditions - or even that sometimes the best path it finds from A [to] B will take it around the CACTF.'' This comment particularly applies to UGVs that were to use the CACTF’s roads as their primary navigation routes. 
$SC_{2}$ cited another impact on workload related to ``trying to push rovers into just the right position to neutralize [an artifact].'' The vehicles had to be within a specific range of an active artifact in order for the Bluetooth communications to be active and neutralize the artifact, while also ensuring that the vehicle did not become neutralized itself. This situation was particularly challenging for artifacts that require multiple vehicles to simultaneously interact with the artifact. 


The SCs' subjective overall workload decreased after the distinguished visitors day. $SC_{1}$'s objective metrics, both estimated overall workload and frequency of overload classifications, decreased, but $SC_{2}$'s were only slightly lower. The SCs were asked if they generally felt that their \textit{workload during [their] shifts was lower after the distinguished visitor day}. $SC_{1}$ felt that after that date most of the pressure and anxiety had been lifted, swarm deployments at this CACTF had become less stressful and easier to process, and the addition of interaction tools added during the FX that simplified tactic specification and provided better situation awareness. $SC_{2}$ stated:' ``My stress level was definitely lower after [the distinguished visitor] day, because I felt the major goal had been accomplished and we did well, but because we were still  [increasing the number of vehicles] and capabilities, I [do not] think my workload was much lower. It may have
been ``slightly'' lower just because [I had] gained a lot of practice using I3 by that point.''

The SCs were asked \textit{How did the joint shifts (with a single SC) impact workload compared to the prior single team shifts?} $SC_{2}$ indicated ``I [do not] recall the joint shifts having any impact. … Especially since we largely ignored the [other] team.’’ Recall that the CACTF was spatially split between the two teams. $SC_{1}$ felt that the ``spatial deconfliction was relatively straightforward.’’ I3 did display the other team’s vehicles’ telemetry using the standard vehicle glyph that excluded some information, and ``after a short discovery period it was obvious which generic assets types [the other team’s] telemetry mapped into.”  However, the increased communication necessary to incorporate the other team’s vehicle telemetry update did create latency that ``became problematic for I3.’’ This latency ``[increased] application input lag …, making the overall system less responsive.’’

The SCs’ decision to jointly operate the swarm was not an I3  system development or the experimental data collection consideration. The SCs were asked if they \textit{feel that situation increased or lowered your workload?} Recall that workload metrics were not collected for $SC_{2}$, who handled the mission command signals and that the SCs ``split’’ the CCAST team’s designated CACTF area. $SC_{2}$ stated:\ ``Yes, I subjectively felt higher workload during this shift’’ and cited the added chatter in the C2 room, ``needing to tell [$SC_{1}$] about what I was doing when it might conflict’’ and ``[determining] if something happening near the middle of the CACTF [the SCs’ boundary] was due to my actions or [$SC_{1}$’s].
$SC_{1}$, from whom workload metrics were collected, stated workload was ``Lowered.’’ Trust was an important element, as $SC_{1}$ stated ``I trusted the co-commander to see to their area of responsibility, and that they would explicitly coordinate when/if they needed to interact near the boundary we established.’’ $SC_{1}$ also noted ``we were happy to be able to enjoy something new and novel which we had talked about for years, but was never part of the program goal.’’ 

\subsection{Discussion}

The CCAST FX-6 results analysis supports the claim that a single human can supervise a true heterogeneous swarm of robots to complete mission relevant tasks in real world environments. The analysis also demonstrates that the multi-dimensional workload estimation algorithm provides results sensitive to actual SC workload changes. While both CCAST SCs experienced overload conditions, their estimated overall workload was within the normal range for 96\% of the generated estimates across all data collection shifts. 


Adams has led the multi-dimensional workload algorithm development, including the initial investigations into the appropriate physiological metrics since 2008 \cite{HarriottPhD15}. The algorithm is intentionally developed to be sensitive to changes in a human's individual workload components and overall workload, as different complex systems, environments, and application domains impact each workload component differently. The prior laboratory-based human subjects evaluations provided evidence that the algorithm performs well and is sensitive across domains, human-robot teaming relationships (i.e.,\ supervisory, peer-based), and individual differences \cite{Fortune2020HFES,Heard2019jcedm,Heard2019thri}. The algorithm has also been demonstrated to detect shifts in workload in real-time in order to adapt a robot's interaction with the human and autonomously change task responsibilities when the human's workload is over- or underloaded \cite{Heard2020CogSima,HeardFN2022}. However, the prior work depended on knowing a priori the evaluation trials' tasks as well as workload levels and transitions. 

The DARPA OFFSET field exercise presented a unique opportunity to apply this algorithm to a hardware-based human-swarm team completing a complex mission in an actual urban environment. The challenges (i.e.,\ weather conditions, constantly changing situations) generated an exceptionally messy and uncontrollable human subjects evaluation that cannot be replicated by laboratory-based evaluations. The result was a true test for the multi-modal human subjects metric sensors, the multi-dimensional workload algorithm, and the associated analysis. Overall, the physiological sensors generally performed as expected in the extreme FX conditions. The noise meter issues were associated with a factory default setting. 

The nature of the OFFSET field exercise shifts make it very difficult to develop representative underload, normal load and overload IMRPINT Pro models. As as such, previously developed IMRPINT Pro models for a single human supervising a large UAV were used to represent the missing metrics. Previously validated neural network workload component models for the supervisory-based adaptive human-robot teaming architecture \cite{Heard2020CogSima} were used when generating the workload components' and overall workload estimates. While that domain differs quite a bit, especially in the number of vehicles, it was the most representative domain. The choice to use these existing trained models did facilitate an analysis of the shifts that demonstrates the algorithm's sensitivity to changing workload conditions. 

As discussed, stress and fatigue appeared to have a limited impact on cognitive workload, but did impact physical workload. The physical workload component estimation was primarily dependent on heart rate and respiration rate, which coupled with the SC's limited physical movements, led to over estimates of physical workload. A limitation of this overall representation of physical workload is that it does not clearly represent the three types of physical workload: gross motor, fine motor, and tactile. The CCAST SC's physical interactions  were generally fine-grained and tactile, for which the physical workload metrics struggle to assess. Data was collected using Myo devices on the SC's arms intended to capture the fine grained and tactile interactions, but these sensors were not yet integrated into multi-dimensional workload algorithm. Adams' team has recently completed preliminary work to model these physical workload components \cite{BhagatSmithCHMS2022} using the Myo and other sensor results for another domain. The resulting estimates are dependent on metrics that are less susceptible to stress and fatigue, which can improve the reliability of the individual estimates and the overall workload estimates.  

The use of the IMRPINT Pro models to estimate visual workload is a clear limitation. The Valve Index headset does not provide eye tracking, and the headset cannot be worn with an eye tracker, such as a Pupil Lab Pupil Core. The team discussed purchasing and integrating a new headset, but decided it was a low priority. 
Adams' research group 
has only very recently developed the visual workload estimation capability. The incorporation of a metric driven estimate is expected to improve the reliability and accuracy of the overall workload estimates.  

The prior laboratory-based human subjects experiments and associated multi-dimensional workload algorithm validations assume that the human's tasks are known. This task context has been shown to improve the accuracy of the component and overall workload assessments. It is important to note that while the presented analysis demonstrates sensitivity to workload changes in the SCs' task demands, task context was not available. The practical use of the multi-dimensional workload algorithm for actual military deployments, such as the one that the OFFSET program was based, will require the ability to infer the SC's current task. Adams' group is actively developing a multi-dimensional task recognition approach dependent on wearable sensors that accommodates the breadth of tasks in such domains. The initial capabilities are focused on visual task recognition \cite{BaskaranICHMS2022}. Assuming that such a system can reliably infer a SC's tasks, then such context is hypothesized to also improve the component and overall workload estimations.  

The use of the various physiological sensors at FX-6 represents the first time they were used in such harsh conditions. While the sensors generally preformed well, these sensors are not hardened for daily use, let alone routine use in harsh mission conditions. The routine use of the such sensors for disaster response and military domains will require their miniaturization, reduce power consumption, hardening, etc.  

The DARPA OFFSET program assumes the swarm vehicles are highly autonomous, as well as the CCAST team's approach to relying on mission plans or SC specific high level tactics, and I3 design decisions (e.g.,\ an immersive visualization, vehicle and tactic glyphs, artifact icons and associated prioritization filtering) directly enable the SC's ability to deploy and supervise the swarm. Future swarms deployed in similar urban environments that have high volumes of occupied space and require vertically will need similar interaction affordances. It may be desirable to provide very precise individual vehicle and goal point manipulation, and some missions may require such precision; however, achieving such precision may be very difficult for swarms, and if  incorporated, this level of interaction is expected to increase the SC's workload.  

The DARPA OFFSET program uses AprilTags to represent the scenario artifacts, which was done to allow the integrator teams to focus on scaling the hardware swarm's size. The CCAST system uses cameras (e.g.,\ PiCam) and simple image recognition to perceive and differentiate the AprilTags. Assuming an AprilTag is perceived correctly, the tag identifier is mapped to an I3 icon. I3 automatically filters the artifacts so that only the most relevant artifacts are presented to the SC. Note, the SC can display all artifacts if desired. A different system that relies on sensor perception to identify artifacts (e.g.,\ image processing, electronic signal recognition) may have higher perception error that generates false artifact identifications, or requires incorporating a representation of the system's recognition confidence level. This potentially higher error rate, or the increased complexity of incorporating confidence intervals may increase the SC’s workload, but the true impact can only be hypothesized and will be highly dependent on the particular perception system’s error rates, the user interface design, underlying decision support systems, the ability to associate confidence with the perceived artifact, the mission scenario, etc. 

One may want to consider providing live video feed to help recognize an artifact (i.e.,\ not an AprilTag), but the mission complexity, the swarm size and heterogeneity, as well as the broader SC duties will impact the viability of such an approach. It is feasible to believe that permitting a limited number of live video feeds can be reasonably used by a SC, but even adding a small number of feeds is expected to increase the workload. The impact on workload from such a feature will be highly dependent on how many live feeds are permitted, the steps required to enable/disable a feed, the feed's presentation within I3 and its association relative to the associated vehicle, the sensor’s field of view, reliability and accuracy, as well as the purpose of the live feed. Assuming those issues are solved, available communication bandwidth, and even whether or not a vehicle is in communication will impact the usability of such information, which will impact the SC's workload. Live feeds were investigated in earlier DARPA OFFSET field exercises, and it was determined that given the low quality images, limited available bandwidth, and the likelihood of vehicles being out of communication, the video feeds were not very useful for the outdoor mission elements. When the vehicles are within a building, they are out of communication and no live sensor feed is feasible. 

\section{Conclusions}
The DARPA OFFSET program demonstrated that a single human can deploy and supervise a swarm of 100 heterogeneous robots. The CCAST team's earlier DARPA OFFSET program field exercise observations demonstrated a trained SC's ability to deploy swarms over shifts that were up to three hours in duration.  The FX-6 outcomes provided further validation of these observations. The CCAST team collected various metrics across twelve shifts (eight CACTF shifts) that were used to estimate the SC's workload components (i.e.,\ cognitive, speech, auditory and physical) and overall workload via the multi-dimensional workload algorithm.  The estimated overall workload was manageable and generally remained within a reasonable normal range. SCs' perceived stress was manageable, but spiked during critical shifts, such as distinguished visitors day, and perceived fatigue was manageable, but varied for many reasons, including shift duration. Generally, the resulting estimates demonstrate that the overall workload estimates increased as the SC's tasks increased, even though the physical workload estimates demonstrated some susceptibility to stress and fatigue. This human subjects data set represents the first known data set for a single human deploying a hardware swarm in an actual urban environment to complete a complex mission. 
The results have broader implications that indicate the viability of future civilian single SC-swarm applications, such as disaster response (e.g.,\ infrastructure safety inspections, wildland fire identification and tracking) and commercial applications (e.g.,\ general logistics, deliveries). 



\appendix
\section{FX-6 Contextual Information}
\label{Appx:Conditions}
This appendix provides information pertaining to multiple aspects of the FX-6 shifts and the associated conditions that impact the shift deployments and the SC's associated effort and workload. 

The weather conditions for each day are provided in Table \ref{Tab:FX6Info}.  The weather was highly variable in temperature ranges, as well as wind speeds. 

\begin{table}[thb]
\caption{FX-6 Climate Conditions (Temperature: F, Pressure: inches, Wind: MPH). } \label{Tab:FX6Info}
\begin{center}
\begin{tabular}{|C{0.5in}|C{0.4in}|C{0.4in}|C{0.4in}|C{0.8in}|C{0.5in}|C{1.05in}|C{0.75in}|}
  \hline
\multirow{2}{*}{\textbf{ Date}}  & \textbf{Low Temp} & \textbf{High Temp} & \textbf{Dew Point} & \textbf{Barometric Pressure} & \textbf{Mean Wind} & \textbf{Max Sustained Wind} & \textbf{Max Wind Gust}\\
  \hline\hline
12-Nov & 39\textdegree & 60\textdegree & 41\textdegree & 30.00  &  10 & 22 & 29  \\ \hline
13-Nov & 27\textdegree & 48\textdegree & 31\textdegree & 30.10  & 6  & 18  & 25  \\ \hline
14-Nov & 33\textdegree & 54\textdegree & 34\textdegree & 30.03  & 8  & 20  & 26  \\ \hline
15-Nov & 28\textdegree & 57\textdegree & 32\textdegree & 30.16 & 4  & 9  & 9 \\  \hline
16-Nov & 40\textdegree & 67\textdegree & 47\textdegree & 30.03 & 6  & 14  & 14\\ \hline
17-Nov & 57\textdegree & 72\textdegree & 58\textdegree & 30.04 & 12  & 21  & 28  \\ \hline
18-Nov & 41\textdegree & 69\textdegree & 40\textdegree & 30.24 & 10  & 23  & 32 \\ \hline  \hline
\end{tabular}
\end{center}
\end{table}

\clearpage
Each shift had a different composition of vehicles, as shown in Table \ref{Tab:FX6Vehicles}. Most shifts included hardware vehicles, while some incorporated virtual vehicles. 

\begin{table}[!htb]
\caption{FX-6 shift allocations by SC and pre-mission brief vehicle counts. Gray rows represent $SC_{2}$'s shifts, no data was collected for red rows, and NR means no data recorded.} \label{Tab:FX6Vehicles}
\begin{center}
\begin{tabular}{|c|c||l|l|l|l||l|l|}
  \hline

\multicolumn{2}{|c||}{ \textbf{Shift}} & \multicolumn{2}{|c|}{ \textbf{Ground Vehicles}} & \multicolumn{2}{|c|}{ \textbf{Aerial Vehicles}} & \multicolumn{2}{|c||}{ \textbf{Total Vehicles}} \\ \hline
 \textbf{Date} & \textbf{Time} &\textbf{Hardware} & \textbf{Virtual} &\textbf{Hardware} & \textbf{Virtual} &\textbf{Hardware} & \textbf{Virtual}\\
  \hline\hline
\multirow{4}{*}{ 11-Nov }&
\cellcolor{tearose}1100-1200 & \cellcolor{tearose}0 & \cellcolor{tearose}20 & \cellcolor{tearose}0 & \cellcolor{tearose}80& \cellcolor{tearose}0 & \cellcolor{tearose}100\\ \cline{2-8}
& 1300-1400 & 0 & 15 & 0 & 65 & 0 & 80 \\ \cline{2-8}
  & 1500-1600 & 0 & 6 & 0 & 20 & 0 & 26\\ \cline{2-8}
  & \cellcolor{lightgray}1630-1730 & \cellcolor{lightgray}0 & \cellcolor{lightgray}6 & \cellcolor{lightgray}0 & \cellcolor{lightgray}20 & \cellcolor{lightgray}0 & \cellcolor{lightgray}26\\ \cline{2-8}
  & 1800-1900 & 0 & 2/23 & 0 & 15/70 & 0 & 17/93\\ \hline
12-Nov & \cellcolor{lightgray}0830-1130 & \cellcolor{lightgray}10 & \cellcolor{lightgray}5 & \cellcolor{lightgray}44 & \cellcolor{lightgray}20 & \cellcolor{lightgray}55 &\cellcolor{lightgray}25\\ \hline
13-Nov & 1430-1630 & 8 & 10 & 66 & 10 & 74 & 20\\ \hline
14-Nov & 0800-1130 & 8 & NR & 78 & NR & 93/0 & 118/80\\ \hline
15-Nov & \cellcolor{tearose}1300-1630 & \cellcolor{tearose}10 & \cellcolor{tearose}10 & \cellcolor{tearose}78 & \cellcolor{tearose}20 & \cellcolor{tearose}88 & \cellcolor{tearose}30\\ \hline
16-Nov & \cellcolor{lightgray}1000-1200 & \cellcolor{lightgray}10 & \cellcolor{lightgray}10 & \cellcolor{lightgray}81 &\cellcolor{lightgray}20 & \cellcolor{lightgray}91 & \cellcolor{lightgray}30\\ \hline

\multirow{3}{*}{17-Nov} & {\cellcolor{lightgray}}1200-1400 & \cellcolor{lightgray}10/0/ & \cellcolor{lightgray}0/10/ & \cellcolor{lightgray}108/0/ &\cellcolor{lightgray}0/20/ & \cellcolor{lightgray}118/0/ & \cellcolor{lightgray}0/30/\\ \cline{3-8}
& \multirow{-2}{*}{\cellcolor{lightgray}} & \cellcolor{lightgray}0/10 & \cellcolor{lightgray}20/10 & \cellcolor{lightgray}0/71 &\cellcolor{lightgray}105/20 & \cellcolor{lightgray}0/81 & \cellcolor{lightgray}125/30\\\cline{2-8}
 & 1400-1630 & 8 & 0 & 23 & 0 &  31 & 0 \\ \hline
\multirow{2}{*}{ 18-Nov} & \cellcolor{lightgray}1000-1130 & \cellcolor{lightgray}10 & \cellcolor{lightgray}0 & \cellcolor{lightgray}17 & \cellcolor{lightgray}0 & \cellcolor{lightgray}27 & \cellcolor{lightgray}0\\ \cline{2-8}
 & 1330-1530 & 30 & 10 & 110 & 40 & 140 & 50 \\ \hline 
\end{tabular}
\end{center}
\end{table}

\clearpage 
\section{Results}
This appendix provides additional results.

\subsection{Subjective Results}
\label{Appx:SubResults}

The descriptive statistics for the in situ subjective workload component responses are provided in Table~\ref{Tab:WLCompSubj}. These results are provided by shift and SC. 

\begin{table}[!htb]
\caption{The subjective in situ workload component responses descriptive statistics, mean (SD), by shift and SC. Gray cells represent $SC_{2}$'s results.} \label{Tab:WLCompSubj}
\begin{center}
\begin{tabular}{|c|c||l|l|l|l|l|}
  \hline
\multicolumn{2}{|c||}{ \textbf{Shift}} & \multirow{2}{*}{\textbf{Cognitive}} & \multirow{2}{*}{\textbf{Speech}} & \multirow{2}{*}{\textbf{Auditory}} & \multirow{2}{*}{\textbf{Visual}} & \multirow{2}{*}{\textbf{Physical}}  \\ \cline{1-2} 
 \textbf{Date} & \textbf{Time} &  &   &  & & \\
  \hline\hline
\multirow{4}{*}{ 11-Nov }&  1300 & 27.6 (8.76) & 21.2 (7.16)& 18 (0)  & 37.4 (13.27) &24.2 (14.7)   \\ \cline{2-7}
& 1500 & 17.67 (16.5) & 17.67 (16.5)  & 34 (0) & 23.17 (25.15) & 18 (0) \\ \cline{2-5}
& \cellcolor{lightgray}1630 &  \cellcolor{lightgray}28.67 (9.24) & \cellcolor{lightgray}28.67 (9.24) & \cellcolor{lightgray}45 (9.53)  & \cellcolor{lightgray}17.67 (16.5) & \cellcolor{lightgray}28.67 (9.24) \\ \cline{2-7}
& 1800  & 34.13 (13.27) & 22 (8) & 26 (9.39) & 38.13 (8.25) & 34 (0) \\ \hline
12-Nov & \cellcolor{lightgray}0830 & \cellcolor{lightgray}37.28 (21.32) & \cellcolor{lightgray}40.96 (20.36) & \cellcolor{lightgray}35.5 (19.09) & \cellcolor{lightgray}31.92 (16.96)  &
\cellcolor{lightgray}38.21 (12.29)  \\ \hline
13-Nov & 1430 & 40.65 (11.43) & 27.65 (11.3) & 21.1 (10.33) & 42.3 (11.55) & 30.85 (10.23) \\ \hline
14-Nov & 0800 & 41.39 (16.14) & 30.53 (11.86) & 22.42 (10.9) & 43.22 (12.82) & 45.94 (10.97)  \\ \hline
 16-Nov & \cellcolor{lightgray}1000 & \cellcolor{lightgray}49.32 (7.83) & \cellcolor{lightgray}37.61 (11.41) & \cellcolor{lightgray}44.64 (13.82) & \cellcolor{lightgray}42.29 (12.46) & \cellcolor{lightgray}37.61 (11.41)  \\ \hline
\multirow{2}{*}{17-Nov} & \cellcolor{lightgray}1200 & \cellcolor{lightgray}37.09 (12.23) & \cellcolor{lightgray}26.77 (11.11) & \cellcolor{lightgray}32.64 (11.37) & \cellcolor{lightgray}32.64 (17.1) & \cellcolor{lightgray}26.77 (8.36)  \\ \cline{2-7}
 & 1400 & 26.93 (18.71) & 24.79 (16) & 22.36 (18.32) & 31.71 (17.57) & 24.71 (12.89)  \\ \hline
\multirow{2}{*}{ 18-Nov} & \cellcolor{lightgray}1000 & \cellcolor{lightgray}32.5 (11.97) & \cellcolor{lightgray}29.25 (10.91) & \cellcolor{lightgray}26.05 (11.43) & \cellcolor{lightgray}22.6 (13.53) & \cellcolor{lightgray}27.6 (8.26) \\ \cline{2-7}
 & 1330 & 37.89 (17.92) & 36.58 (9.07) & 35.35 (10.43) & 39.19 (15.46) & 34.077 (9.38)   \\ \hline 
\end{tabular}

\end{center}
\end{table}

\clearpage
\subsection{Individual Shift Estimated Workload Analysis}
\label{Appx:ShiftResults}

\paragraph{Nov $16^{th}$ Shift:}
\label{Appx:16-Nov-Results}

The comparison of the in situ workload component responses compared to the individual workload component estimates are provided in Table~\ref{Tab:16-NovWLSubjObj}. The overall workload results also provided for completeness. 

\begin{table}[!htb]
\caption{The Nov $16^{th}$ shift's subjective in situ workload component responses and overall workload descriptive statistics along with the corresponding mean (SD) by in situ subjective data collection time point.} \label{Tab:16-NovWLSubjObj}
\begin{center}
\begin{tabular}{|c|c||l|l|l|l|l|l|}
  \hline
\textbf{Time} & \textbf{Metric} & \textbf{Cognitive} & \textbf{Speech} & \textbf{Auditory} & \textbf{Visual} & \textbf{Physical} & \textbf{Overall}  \\ \hline\hline
\multirow{2}{*}{1000}& Subj. & 34 &	34	& 34	& 34	& 18	& 31.60  \\ \cline{2-8}
 & \cellcolor{vlgray}{Est.} & \cellcolor{vlgray}{43.52	(3.75)} & \cellcolor{vlgray}{9.73 (0.00)} & \cellcolor{vlgray}{59.20 (4.87)} & \cellcolor{vlgray}{--} & \cellcolor{vlgray}{61.34 (4.06)} &	\cellcolor{vlgray}{53.73 (1.80)}  \\ \hline 
 \multirow{2}{*}{1012}& Subj. & 50.5 & 50.5 &	50.5 & 34 & 34	& 44.73\\ \cline{2-8}
 & \cellcolor{vlgray}{Est.} & \cellcolor{vlgray}{58.48	(4.94)} & \cellcolor{vlgray}{12.54 (4.50)} & \cellcolor{vlgray}{57.50 (3.39)} & \cellcolor{vlgray}{--} & \cellcolor{vlgray}{18.30 (0.95)} &	\cellcolor{vlgray}{50.51 (1.96)} \\ \hline 
  \multirow{2}{*}{1020}& Subj. & 50.5 &	50.5 &	50.5 &	34	& 34 & 44.73\\ \cline{2-8}
 & \cellcolor{vlgray}{Est.} & \cellcolor{vlgray}{56.30 (1.34)} & \cellcolor{vlgray}{41.79 (5.94)} & \cellcolor{vlgray}{62.26 (3.44)} & \cellcolor{vlgray}{--} & \cellcolor{vlgray}{51.73 (2.93)} & \cellcolor{vlgray}{58.55 (1.3)}\\ \hline 
   \multirow{2}{*}{1030}& Subj. & 50.5 & 50.5 &	67 & 34 & 34 & 46.38\\ \cline{2-8}
 & \cellcolor{vlgray}{Est.} &  \cellcolor{vlgray}{55.38 (5.26)} &	\cellcolor{vlgray}{47.09 (8.27)} & \cellcolor{vlgray}{50.68 (2.98)} & \cellcolor{vlgray}{--} & \cellcolor{vlgray}{24.53 (1.65)} & \cellcolor{vlgray}{51.95 (1.12)} \\ \hline 
    \multirow{2}{*}{1043}& Subj. & 50.5	& 34 & 18 & 34 & 34	& 39.00\\ \cline{2-8}
 & \cellcolor{vlgray}{Est.} &  \cellcolor{vlgray}{56.45 (2.95)} &  \cellcolor{vlgray}{57.58 (2.57)} & \cellcolor{vlgray}{51.16 (2.89)} & \cellcolor{vlgray}{--} & \cellcolor{vlgray}{62.80 (5.02)} & \cellcolor{vlgray}{60.59 (2.15)} \\ \hline 
    \multirow{2}{*}{1051}& Subj. & 50.5	& 34 & 50.5	& 50.5	& 50.5	& 48.03 \\ \cline{2-8}
 & \cellcolor{vlgray}{Est.} &  \cellcolor{vlgray}{40.88 (3.17)} &	\cellcolor{vlgray}{27.73 (5.92)} & \cellcolor{vlgray}{54.38 (1.67)} & \cellcolor{vlgray}{--} & \cellcolor{vlgray}{22.59 (2.47)} & \cellcolor{vlgray}{45.64	(0.73)} \\ \hline 
    \multirow{2}{*}{1101}& Subj. & 50.5 &	34 &	50.5 &	34 &	34 &	42.25 \\ \cline{2-8}
 & \cellcolor{vlgray}{Est.} & \cellcolor{vlgray}{58.07	(3.36)} &	\cellcolor{vlgray}{56.41 (5.79)} & \cellcolor{vlgray}{48.66 (3.15)} & \cellcolor{vlgray}{--} &  \cellcolor{vlgray}{11.89 (0.99)} & 	\cellcolor{vlgray}{50.73 (1.11)}\\ \hline 
    \multirow{2}{*}{1111}& Subj. & 50.5	& 50.5	& 67	& 50.5	& 50.5 &	52.15\\ \cline{2-8}
 & \cellcolor{vlgray}{Est.} &  \cellcolor{vlgray}{52.10 (1.52)} &	 \cellcolor{vlgray}{28.49 (3.51)} & \cellcolor{vlgray}{69.39 (3.44)} & \cellcolor{vlgray}{--} &  \cellcolor{vlgray}{14.66 (1.83)} &	\cellcolor{vlgray}{49.63 (0.68)}\\ \hline
     \multirow{2}{*}{1120}& Subj. & 50.5 &	50.5 &	34 &	50.5 &	34 &	46.38\\ \cline{2-8}
 & \cellcolor{vlgray}{Est.} & \cellcolor{vlgray}{56.30 (1.34)} &	\cellcolor{vlgray}{41.79 (5.94)} & \cellcolor{vlgray}{50.41 (3.98)} & \cellcolor{vlgray}{--} & \cellcolor{vlgray}{51.73 (2.93)} &	\cellcolor{vlgray}{49.91 (2.37)} \\ \hline 
     \multirow{2}{*}{1130}& Subj. & 50.5 &	34 &	50.5 &	50.5 &	34 &	45.55\\ \cline{2-8}
 & \cellcolor{vlgray}{Est.} &  \cellcolor{vlgray}{37.97 (4.83)} &	\cellcolor{vlgray}{54.77 (7.04)} & \cellcolor{vlgray}{52.82 (1.80)} & \cellcolor{vlgray}{--} &  \cellcolor{vlgray}{11.24 (2.41)} &	\cellcolor{vlgray}{43.74 (1.74)}\\ \hline 
     \multirow{2}{*}{1140}& Subj. & 50.5 &	34 &	50.5 &	67	& 50.5 &	51.33\\ \cline{2-8}
 & \cellcolor{vlgray}{Est.} &  \cellcolor{vlgray}{46.74 (7.08)} &	\cellcolor{vlgray}{72.78 (3.31)} & \cellcolor{vlgray}{42.52 (2.73)} & \cellcolor{vlgray}{--} &  \cellcolor{vlgray}{18.14 (4.18)} & 	\cellcolor{vlgray}{48.25 (2.18)}\\ \hline 
     \multirow{2}{*}{1150}& Subj. & 50.5 &	18	& 34	& 50.5	& 50.5	& 43.98\\ \cline{2-8}
 & \cellcolor{vlgray}{Est.} & \cellcolor{vlgray}{56.59 (4.52)} &	\cellcolor{vlgray}{57.61 (3.09)} & \cellcolor{vlgray}{54.71 (3.07)} & \cellcolor{vlgray}{--} & \cellcolor{vlgray}{9.80 (2.46)} &	\cellcolor{vlgray}{50.46 (1.39)} \\ \hline
     \multirow{2}{*}{1155}& Subj. & 67	& 34	& 34	& 50.5	& 50.5	& 52.98\\ \cline{2-8}
 & \cellcolor{vlgray}{Est.} &  \cellcolor{vlgray}{39.13 (7.44)} &	\cellcolor{vlgray}{44.28 (4.82)} & \cellcolor{vlgray}{55.57 (3.49)} & \cellcolor{vlgray}{--} &  \cellcolor{vlgray}{17.35 (2.66)} &	\cellcolor{vlgray}{45.04 (2.86)}\\ \hline 
\end{tabular}
\end{center}
\end{table}

\clearpage
\paragraph{Nov $17^{th}$ 1200-1400 Shift:}
\label{Appx:17-Nov-Results}

The comparison of the in situ workload component responses compared to the individual workload component estimates are provided in Table~\ref{Tab:17-NovWLSubjObj}. The overall workload results also provided for completeness. 

\begin{table}[htb]
\caption{The Nov $17^{th}$ 1200-1400 shift's subjective in situ workload component responses and overall workload descriptive statistics along with the corresponding mean (SD) by in situ subjective data collection time point.} \label{Tab:17-NovWLSubjObj}
\begin{center}
\begin{tabular}{|c|c||l|l|l|l|l|l|}
  \hline
\textbf{Time} & \textbf{Metric} & \textbf{Cognitive} & \textbf{Speech} & \textbf{Auditory} & \textbf{Visual} & \textbf{Physical} & \textbf{Overall}  \\ \hline\hline
 \multirow{2}{*}{1220}& Subj. & 34 & 18& 18& 18 & 18 & 24.4\\ \cline{2-8}
 & \cellcolor{vlgray}{Obj.} & \cellcolor{vlgray}{32.04	(3.84)} & \cellcolor{vlgray}{18.47 (5.63)} &	\cellcolor{vlgray}{61.73 (3.63)} & \cellcolor{vlgray}{--} & \cellcolor{vlgray}{62.33	(9.86)} &	\cellcolor{vlgray}{50.58 (2.85)} \\ \hline 
  \multirow{2}{*}{1234}& Subj. & 18 & 18 & 18 & 18 & 18 & 18\\ \cline{2-8}
 & \cellcolor{vlgray}{Obj.} & \cellcolor{vlgray}{38.53	(7.56)} & \cellcolor{vlgray}{28.62 (2.10)} & \cellcolor{vlgray}{57.56	(2.37)} & \cellcolor{vlgray}{--} & \cellcolor{vlgray}{22.71	(2.57)} & \cellcolor{vlgray}{45.19 (2.51)} \\ \hline 
   \multirow{2}{*}{1250}& Subj. & 34 & 34 & 50.5 & 18 & 18 & 30.05\\ \cline{2-8}
 & \cellcolor{vlgray}{Obj.} &  \cellcolor{vlgray}{36.06 (13.29)} &	\cellcolor{vlgray}{51.72 (9.02)} &	\cellcolor{vlgray}{53.94 (4.63)} & \cellcolor{vlgray}{--} & \cellcolor{vlgray}{17.22	(1.67)} & \cellcolor{vlgray}{44.18 (4.57)} \\ \hline 
    \multirow{2}{*}{1300}& Subj. & 34 & 18 & 34 & 34 & 34 & 31.6\\ \cline{2-8}
 & \cellcolor{vlgray}{Obj.} &  \cellcolor{vlgray}{37.45 (8.45)} &	\cellcolor{vlgray}{42.88 (4.13)} &	\cellcolor{vlgray}{64.39 (3.23)} & \cellcolor{vlgray}{--} & \cellcolor{vlgray}{35.00	(1.65)} & \cellcolor{vlgray}{48.76 (3.42)} \\ \hline 
    \multirow{2}{*}{1310}& Subj. & 34 &18 & 18 & 50.5 & 34 & 33.3 \\ \cline{2-8}
 & \cellcolor{vlgray}{Obj.} &  \cellcolor{vlgray}{53.98 (6.77)} &	\cellcolor{vlgray}{19.31 (6.04)} &	\cellcolor{vlgray}{58.60 (1.90)} & \cellcolor{vlgray}{--} & \cellcolor{vlgray}{25.96	(4.72)} & \cellcolor{vlgray}{50.93 (1.87)} \\ \hline 
    \multirow{2}{*}{1320}& Subj. &  50.5 & 50.5 & 34 & 50.5 & 34 & 46.38\\ \cline{2-8}
 & \cellcolor{vlgray}{Obj.} & \cellcolor{vlgray}{43.45	(3.34)}	& \cellcolor{vlgray}{52.58 (4.98)} & \cellcolor{vlgray}{54.78	(2.58)} & \cellcolor{vlgray}{--} &  \cellcolor{vlgray}{34.37	(4.36)} & \cellcolor{vlgray}{50.37	(1.96)}\\ \hline 
    \multirow{2}{*}{1330}& Subj. & 18 & 18 & 34 & 50.5 & 50.5 & 30.98\\ \cline{2-8}
 & \cellcolor{vlgray}{Obj.} &  \cellcolor{vlgray}{46.73 (3.05)} &	\cellcolor{vlgray}{9.32 (1.40)} & \cellcolor{vlgray}{51.73 (4.40)} & \cellcolor{vlgray}{--} & \cellcolor{vlgray}{43.83 (4.18)} &  \cellcolor{vlgray}{50.62 (1.62)} \\ \hline 
    \multirow{2}{*}{1340}& Subj. & 50.5 & 34 & 34 & 50.5 & 18 & 41.5 \\ \cline{2-8}
 & \cellcolor{vlgray}{Obj.} &  \cellcolor{vlgray}{58.31 (5.02)} &	\cellcolor{vlgray}{49.14 (5.71)} &	\cellcolor{vlgray}{44.93 (4.14)} & \cellcolor{vlgray}{--} & \cellcolor{vlgray}{36.45	(1.86)} & \cellcolor{vlgray}{54.90 (1.81)} \\ \hline 
    \multirow{2}{*}{1350}& Subj. & 50.5 & 34 & 34 & 34 & 34 &40.6 \\ \cline{2-8}
 & \cellcolor{vlgray}{Obj.} &  \cellcolor{vlgray}{42.51 (7.34)} &	\cellcolor{vlgray}{29.94 (4.05)} &	\cellcolor{vlgray}{52.64 (3.45)} & \cellcolor{vlgray}{--} & \cellcolor{vlgray}{34.40	(6.69)} &	\cellcolor{vlgray}{48.52 (1.62)} \\ \hline 
    \multirow{2}{*}{1355}& Subj. & 50.5 & 34 & 50.5 & 34 & 18 & 39.85\\ \cline{2-8}
 & \cellcolor{vlgray}{Obj.} & \cellcolor{vlgray}{31.22	(10.68)} &	\cellcolor{vlgray}{53.80 (9.56)} & 	\cellcolor{vlgray}{60.30 (2.27)}  & \cellcolor{vlgray}{--} &  \cellcolor{vlgray}{43.19 (3.10)} &	\cellcolor{vlgray}{48.38 (3.80)}\\ \hline 
\end{tabular}
\end{center}
\end{table}

\clearpage
\paragraph{Joint Integrator and SCs Shift, Nov $18^{th}$ 1330-1530 Shift:}
\label{Appx:18-Nov-Results}

The comparison of the in situ workload component responses compared to the individual workload component estimates are provided in Table~\ref{Tab:18-NovWLSubjObj}. The overall workload results also provided for completeness. 

\begin{table}[htb]
\caption{The Nov $18^{th}$ 1330-1530 shift's subjective in situ workload component responses and overall workload descriptive statistics along with the corresponding mean (SD) by in situ subjective data collection time point.} \label{Tab:18-NovWLSubjObj}
\begin{center}
\begin{tabular}{|c|c||l|l|l|l|l|l|}
  \hline
\textbf{Time} & \textbf{Metric} & \textbf{Cognitive} & \textbf{Speech} & \textbf{Auditory} & \textbf{Visual} & \textbf{Physical} & \textbf{Overall}  \\ \hline\hline
   \multirow{2}{*}{1400}& Subj. & 34 & 34 & 34 & 50.5 & 34 & 39.78\\ \cline{2-8}
 & \cellcolor{vlgray}{Obj.} &  \cellcolor{vlgray}{10.29 6.44)} & \cellcolor{vlgray}{--} & \cellcolor{vlgray}{56.35	(2.81)} & \cellcolor{vlgray}{--} & \cellcolor{vlgray}{13.76	(0.88)} &	
 \cellcolor{vlgray}{34.43 (2.67)} \\ \hline 
    \multirow{2}{*}{1410}& Subj. & 50.5 & 34 & 34 & 34 & 34 & 38.13\\ \cline{2-8}
 & \cellcolor{vlgray}{Obj.} &  \cellcolor{vlgray}{56.52 (7.22)} & \cellcolor{vlgray}{--} & \cellcolor{vlgray}{55.21 (4.9)} & \cellcolor{vlgray}{--} & \cellcolor{vlgray}{18.76	(1.66)} &	\cellcolor{vlgray}{51.83 (2.77)} \\ \hline 
    \multirow{2}{*}{1420}& Subj. & 50.5 & 50.5 & 50.5 & 34 & 34 & 43.9 \\ \cline{2-8}
 & \cellcolor{vlgray}{Obj.} &  \cellcolor{vlgray}{12.32 (7.43)} & \cellcolor{vlgray}{--} & \cellcolor{vlgray}{59.22 (1.63)} & \cellcolor{vlgray}{--} & \cellcolor{vlgray}{18.75 (5.48)} &	\cellcolor{vlgray}{36.44 (1.78)} \\ \hline 
    \multirow{2}{*}{1430}& Subj. & 50.5 & 34 & 50.5 & 50.5 & 34 & 46.38 \\ \cline{2-8}
 & \cellcolor{vlgray}{Obj.} & \cellcolor{vlgray}{12.62	(1.19)} & \cellcolor{vlgray}{--} & \cellcolor{vlgray}{52.88	(2.44)} & \cellcolor{vlgray}{--} &  \cellcolor{vlgray}{13.97 (2.65)} &	
 \cellcolor{vlgray}{34.96 (0.83)}\\ \hline 
  \multirow{2}{*}{1440}& Subj. & 18 & 18 & 18 & 18 & 34 & 18.8\\ \cline{2-8}
 & \cellcolor{vlgray}{Obj.} & \cellcolor{vlgray}{12.59	(0.55)} & \cellcolor{vlgray}{--} & \cellcolor{vlgray}{50.68 (7.43)} & \cellcolor{vlgray}{--} & \cellcolor{vlgray}{8.52 (1.19)} &	
 \cellcolor{vlgray}{33.64	(0.6)} \\ \hline 
  \multirow{2}{*}{1450}& Subj. & 34 & 34 & 34 & 34 & 50.5 & 34.83\\ \cline{2-8}
 & \cellcolor{vlgray}{Obj.} & \cellcolor{vlgray}{52.59	(5.59)} & \cellcolor{vlgray}{--} & \cellcolor{vlgray}{49.89	(3.34)} & \cellcolor{vlgray}{--} & \cellcolor{vlgray}{27.17	(7.99)} &	
 \cellcolor{vlgray}{51.56 (1.85)} \\ \hline 
   \multirow{2}{*}{1500}& Subj. & 50.5 & 50.5 & 50.5 & 67 & 34 & 55.45\\ \cline{2-8}
 & \cellcolor{vlgray}{Obj.} & \cellcolor{vlgray}{28.65 (15.48)} & \cellcolor{vlgray}{--} & \cellcolor{vlgray}{56.1 (3.64)} & \cellcolor{vlgray}{--} & \cellcolor{vlgray}{27.12 (14.33)} &	
 \cellcolor{vlgray}{43.62 (6.48)} \\ \hline 
    \multirow{2}{*}{1510}& Subj. &34 & 34 & 18 & 50.5 & 34 & 37.38\\ \cline{2-8}
 & \cellcolor{vlgray}{Obj.} & \cellcolor{vlgray}{40.17	(5.15)}  & \cellcolor{vlgray}{--} & \cellcolor{vlgray}{58.37 (1.66)} & \cellcolor{vlgray}{--} & \cellcolor{vlgray}{31.24 (0.38)} &	
 \cellcolor{vlgray}{48.79 (1.95)} \\ \hline 
    \multirow{2}{*}{1520}& Subj. & 50.5 & 50.5 & 34 & 50.5 & 34 & 47.2 \\ \cline{2-8}
 & \cellcolor{vlgray}{Obj.} &  \cellcolor{vlgray}{19.34	(2.71)} & \cellcolor{vlgray}{--} & \cellcolor{vlgray}{55.77 (2.98)} & \cellcolor{vlgray}{--} & \cellcolor{vlgray}{12.24 (1.48)} &	
 \cellcolor{vlgray}{37.3 (1)} \\ \hline 
    \multirow{2}{*}{1525}& Subj. & 67 & 34 & 34 & 50.5 & 50.5 & 48.85 \\ \cline{2-8}
 & \cellcolor{vlgray}{Obj.} & \cellcolor{vlgray}{46.39	(4.28)} & \cellcolor{vlgray}{--} & \cellcolor{vlgray}{55.45 (3.08)} & \cellcolor{vlgray}{--} &  \cellcolor{vlgray}{22.63 (6.41)} &	
 \cellcolor{vlgray}{49.01 (2.84)}\\ \hline 
\end{tabular}
\end{center}
\end{table}

\subsubsection*{Acknowledgments}
The authors thank the entire CCAST team, including Drs.\ Shane Clark and David Diller. Adams thanks Prakash Baskaran, Robert Brown and Dr.\ Jamison Heard for contributing to the data collection tools and assisting with the data analysis. Importantly, the authors thank the CCAST SCs for their willingness to be subjected to this human subjects evaluation.

This research was developed with funding from the Defense Advanced Research
Projects Agency (DARPA). The views, opinions, and findings expressed are
those of the authors and are not to be interpreted as representing the official
views or policies of the Department of Defense or the U.S. Government.

DISTRIBUTION STATEMENT A: Approved for public release: distribution
unlimited.

\bibliographystyle{apalike}
\bibliography{Heard,Adams}

\end{document}